\documentclass[lettersize,journal]{IEEEtran}

\usepackage{cite}
\usepackage{graphicx}
\usepackage{url}
\usepackage{amsmath}
\usepackage{amssymb}
\usepackage[caption = false, font=footnotesize]{subfig}
\usepackage{float}
\usepackage{booktabs} 
\usepackage{multirow}
\usepackage[colorlinks=true,linkcolor=blue]{hyperref}%
\usepackage{array}
\usepackage{color}
\usepackage{tabularx}
\usepackage{longtable}
\usepackage{tabu}
\usepackage{pifont}
\usepackage{amssymb}
\usepackage{ragged2e}
\usepackage{xcolor}
\usepackage{colortbl, booktabs}
\usepackage{algorithm}  
\usepackage{algpseudocode}

\makeatletter
\newcommand{\removelatexerror}{\let\@latex@error\@gobble}

\begin{document}

\title{Pleno-Generation: A Scalable Generative Face Video Compression Framework with Bandwidth Intelligence}



\author{Bolin Chen,~\IEEEmembership{Member,~IEEE},
        Hanwei Zhu,~\IEEEmembership{Member,~IEEE},
        Shanzhi Yin,
        Lingyu Zhu,~\IEEEmembership{Member,~IEEE},
        Jie Chen,
        Ru-Ling Liao,
        Shiqi Wang,~\IEEEmembership{Senior Member,~IEEE} and 
        Yan Ye,~\IEEEmembership{Senior Member, IEEE}
\IEEEcompsocitemizethanks{
\IEEEcompsocthanksitem Bolin Chen, Shanzhi Yin, Lingyu Zhu and Shiqi Wang are with the Department of Computer Science, City University of Hong Kong, Hong Kong (e-mail: bolinchen3-c@my.cityu.edu.hk, shanzhyin3-c@my.cityu.edu.hk, lingyzhu-c@my.cityu.edu.hk and shiqwang@cityu.edu.hk).\\
\IEEEcompsocthanksitem Hanwei Zhu is with the Alibaba-NTU Global e-Sustainability CorpLab, Nanyang Technological University, SG (e-mail: hanwei.zhu@ntu.edu.sg).\\
\IEEEcompsocthanksitem  Jie Chen is with DAMO Academy, Alibaba Group, Beijing, China (e-mail: jiechen.cj@alibaba-inc.com).\\
\IEEEcompsocthanksitem  Ru-Ling Liao and Yan Ye are with DAMO Academy, Alibaba Group, Sunnyvale, CA 94085 USA (e-mail: ruling.lrl@alibaba-inc.com and yan.ye@alibaba-inc.com).}
}

\maketitle

\begin{abstract}
\justifying
Generative model based compact video compression is typically operated within a relative narrow range of bitrates, and often with an emphasis on ultra-low rate applications. There has been an increasing consensus in the video communication industry that full bitrate coverage should be enabled by generative coding. However, this is an extremely difficult task, largely because generation and compression, although related, have distinct goals and trade-offs. The proposed Pleno-Generation (PGen) framework distinguishes itself through its exceptional capabilities in ensuring the robustness of video coding by utilizing a wider range of bandwidth for generation via bandwidth intelligence. In particular, we initiate our research of PGen with face video coding, and PGen offers a paradigm shift that prioritizes high-fidelity reconstruction over pursuing compact bitstream. The novel PGen framework leverages scalable representation and layered reconstruction for Generative Face Video Compression (GFVC), in an attempt to imbue the bitstream with intelligence in different granularity. Experimental results illustrate that the proposed PGen framework can facilitate existing GFVC algorithms to better deliver high-fidelity and faithful face videos. In addition, the proposed framework can allow a greater space of flexibility for coding applications and show superior RD performance with a much wider bitrate range in terms of various quality evaluations. Moreover, in comparison with the latest Versatile Video Coding (VVC) codec, the proposed scheme achieves competitive Bjøntegaard-delta-rate savings for perceptual-level evaluations.
\end{abstract}


\section{Introduction}

\begin{figure*}[tb]
\centering
\vspace{-2em}
\subfloat[General Flowchart]{\includegraphics[width=0.58\textwidth,height=4cm]{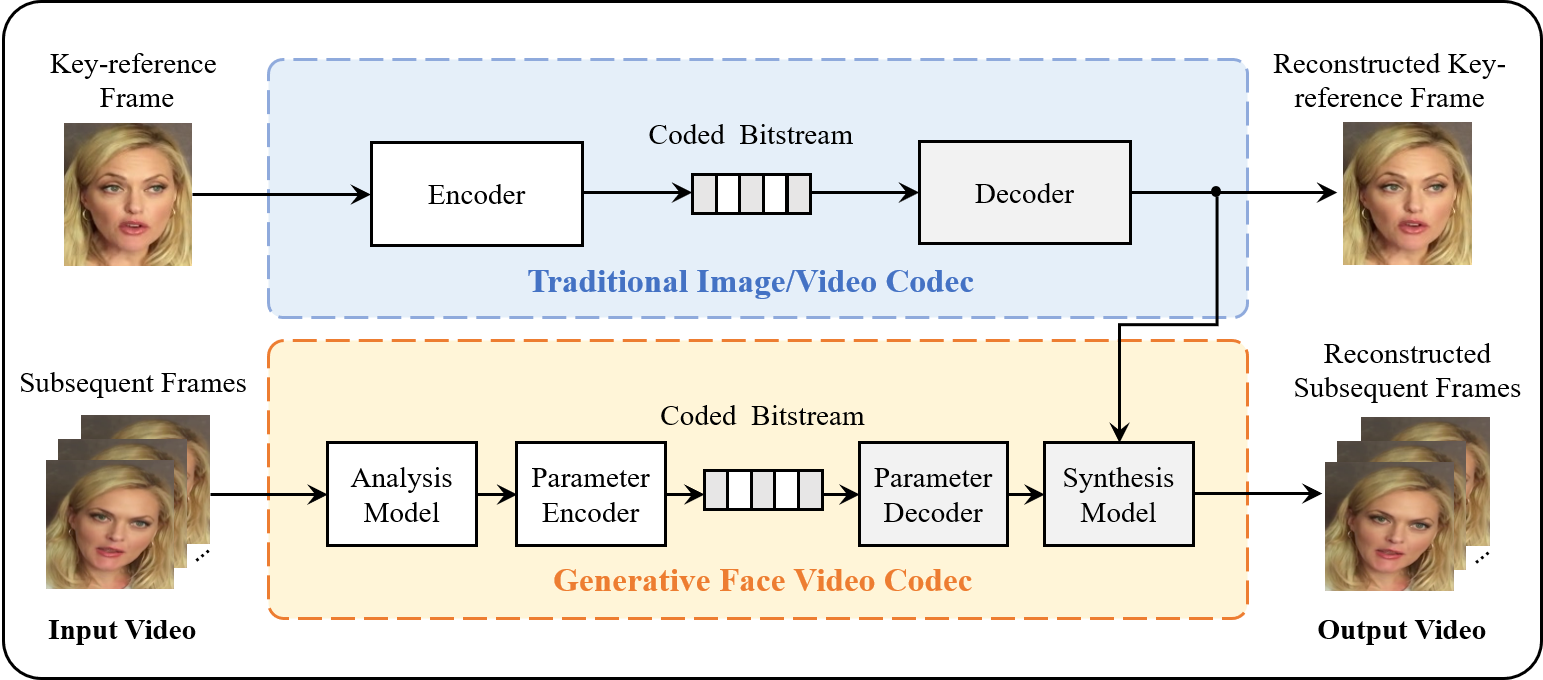}}
\hspace{1.2em}
\subfloat[Compact Facial Representations]{\includegraphics[width=0.38\textwidth,height=4cm]{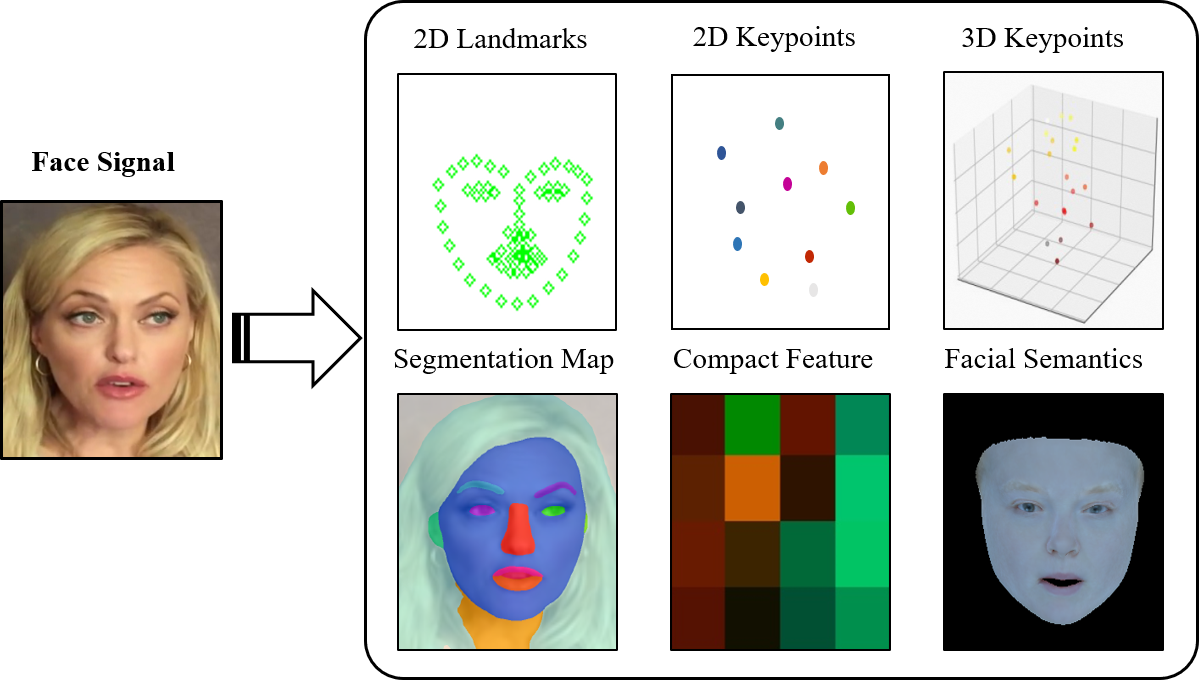}}
\caption{The general flowchart and different facial representations for GFVC~\cite{chen2023generative}.} 
\label{fig1} 
\end{figure*}

\IEEEPARstart{R}{ecent} years have witnessed a paradigm shift in the compression of video data, from signal-level coding to generative representation powered by Artificial Intelligence Generated Content (AIGC) models. In particular, face video coding is important in various meta-verse and real-time video applications. For example, Model-Based Coding (MBC)~\cite{7268565,1457470,1989Object,lopez1995head,150969,364463}, which could be dated back to the 1950s and rapidly developed in the 1990s, was devoted to improving face video compression efficiency by exploiting strong face priors. However, performance was greatly hindered by the not-so-powerful synthesis and analysis abilities of the faces at the time. Motivated by the rapid development of deep learning technologies in various face-related tasks ~\cite{9764682,9349088,9868051,10345691,kong2025pixel, wang2018recurrent,10381809,10547422,6920084,9099607}, especially for strong face generation~\cite{VAE,goodfellow2014generative,NEURIPS2021_49ad23d1}, the MBC framework can now be optimized for realistic reconstruction with vivid texture. 
More specifically, learning-based face reenactment/animation models~\cite{FOMM,8954170,siarohin2021motion,10345691,10547422} have provided reasonable solutions for Generative Face Video Compression (GFVC)~\cite{facebook2021,ultralow,9859867,CHEN2022DCC,icip2022zhao,chen2023generative,10811831,10743340}. As illustrated in Fig. \ref{fig1}, the encoder side employs the analysis model to effectively characterize complex facial motions, whilst the decoder side utilizes the synthesis model to reconstruct high-quality face video. More specifically, the pixel-level facial signal can be economically represented into compact representations, such as 2D landmarks, 2D keypoints, 3D keypoints, temporal trajectory feature, segmentation map and facial semantics, in alignment with the mild assumption that the representation of visual information should follow certain curvatures for human visual system~\cite{henaff2019perceptual}. Moreover, since the year of 2023, the Joint Video Experts Team (JVET) of ISO/IEC JTC 1/SC 29 and ITU-T SG16 has made efforts to standardize GFVC~\cite{chen2024standardizing} and developed the corresponding test conditions~\cite{JVET-AJ2035}/software tools~\cite{JVET-AG0042,JVET-AH0114}/high-level syntax working draft~\cite{JVET-AJ2006}. As a consequence, it is possible to accomplish face video communications in the ultra-low bitrate scenarios.

Although promising compression performance can be achieved along this vein, these novel GFVC approaches still face limitations when following a generative pipeline. Firstly, it is widely acknowledged that GFVC schemes cannot effectively utilize arbitrarily allocated coding rate, such that the superior RD performance is typically constrained within a limited bitrate range and unable to reach a wider bitrate coverage like a traditional video codec. This is largely due to the fact that current GFVC schemes rely on the straightforward application of generation models without specific design.  
Secondly, due to the internal mechanism of GFVC algorithms that rely on the key-reference frame for texture guidance, the overall reconstruction quality could be greatly limited by the key-reference frame. This could cause the reconstructed face videos to suffer from low fidelity, poor local motion representation, temporal inconsistency, and occlusion artifacts. 
Thirdly, generation and compression, although related, have distinct goals and trade-offs. Generative models focus on generating visually rich textures with given features, while compression aims to reconstruct a given video with the allocated bitrate. Therefore, in the context of learning-based compression, the inference process inherently incorporates the ground-truth video content in encoding. This provides us substantial room to design innovative and tailored generative techniques specifically for compression. 

In this paper, motivated by the daunting challenges and perceived opportunities, we propose a novel framework – Pleno-Generation (PGen) that fully and intelligently utilizes the bandwidth for compression. In particular, the proposed PGen attempts to bridge generative models and effective compression based on Scalable Representation and Layered Reconstruction (SRLR)~\cite{jia2019layered, 9547677}. The proposed PGen paradigm is essentially based upon the marriage of computational representation in Marr’s Theory~\cite{man1982computational} and the philosophy of Scalable Video Coding (SVC)~\cite{schwarz2007overview}, where different granularity-level visual representations can well describe the content richness and motion information in a hierarchical manner. This ultimately achieves intelligent bandwidth utilization such that the generation can be achieved with all available resources. To conclude, the proposed PGen enjoys several advantages, including high coding flexibility, perceptual quality improvement and universally plug-and-play property. Firstly, it can well entail conceptually-explicit information to ensure the interpretability of bitstream and allow different layered reconstruction with great coding flexibility. Secondly, it can provide auxiliary visual feature to optimize the existing GFVC algorithms and improve the perceptual quality of reconstructed face video, aligning it better with the perceptual characteristics of Human Visual System (HVS). Finally, it is compatible with all existing GFVC algorithms based on compact representation, which is a universally plug-and-play component to support high-fidelity reconstruction when bandwidth permits.

The main contributions of this paper can be summarized as follows,
\begin{itemize}
\item{We comprehensively summarize and discuss the promises and challenges of GFVC schemes. Inspired by great potentials of GFVC approaches, we further advance GFVC with the PGen for promising RD performance and practical deployments. To our best knowledge, the proposed PGen is the first universal generative coding framework which can hierarchically accommodate for variable bitrate video communication scenarios and actualize high-fidelity face reconstruction towards broad bandwidth coverage.}
\item{We develop a multi-granularity facial description strategy to regularize long-term dependencies between video frames and compensate for motion estimation errors caused by compact representations of motion information. By exploiting the visual information in different granularity, the enhanced coding framework (PGen) could achieve satisfactory human visual perception and bandwidth intelligence in a progressive fashion.}
\item{We design a unique generative decoder that could facilitate reliable and robust face video reconstruction from multi-granularity visual representations. In particular, this decoded auxiliary visual information is employed to recalibrate base-layer-reconstructed face signal with attention, whilst the coarse-to-fine generation strategy is developed to avoid error accumulation. The conducted experiments demonstrate the effectiveness of the proposed PGen for the existing GFVC algorithms, achieve promising RD performance in a broad bitrate coverage. }
\end{itemize}

\section{Literature Overview}

In this section, we provide a literature overview of compact video representation from the traditional hybrid video coding to the generative coding.  

The past decades have witnessed the standardization and deployment of H.264/Advanced Video Coding (AVC)~\cite{wiegand2003overview}, H.265/High Efficiency Video Coding (HEVC)~\cite{sullivan2012overview} and H.266/Versatile Video Coding (VVC)~\cite{bross2021overview}, greatly enabling a variety of broadcasting, streaming and conferencing applications based on the hybrid video coding framework. In addition, a series of scalable video coding extensions~\cite{schwarz2007overview,ohm2005advances,amon2007file,boyce2015overview,ye2014scalable} have been developed to improve coding flexibility via a set of scalability features. In recent years, inspired by the development of deep learning techniques, some tools in hybrid coding framework, such as intra/inter prediction~\cite{zhu2019generative,zhao2018enhanced}, entropy coding~\cite{ma2019convolutional}, interpolation filtering~\cite{liu2019one}, and in-loop filtering~\cite{jia2019content,MengJZWM22a}, have been further optimized and replaced by neural networks for better RD performance.
Afterwards, learning-based image/video coding paradigms~\cite{balle2018variational,lu2019dvc,yang2020learningRLVC,liu9578733,lin2023deepsvc,zhang2023elfic,li2023neural,wang2023EVC,li2024neural} are further developed by optimizing the entire compression framework in an end-to-end manner, showing great potentials for the RD performance improvement. These video coding schemes are typically designed for universal scenarios instead of specific applications (\textit{e.g.,} face video communication). 
For the latter, there is still significant room for improvement in coding efficiency, and further advancements can be achieved by leveraging strong prior information.

In terms of the visual compression for specific scenarios, early MBC techniques ~\cite{7268565,1457470,1989Object,lopez1995head,150969,364463}, which could be dated back to the 1950s and to the 1990s, provided pioneering solutions to fully exploit these statistical priors in image/video coding. In particular, the analysis model can economically represent the structural information (\textit{e.g.,} semantic parameters or facial edges) from face images as transmitted symbols and then the synthesis model uses these decoded symbols to reconstruct face images. These MBC techniques were also proposed to be incorporated in the hybrid coding framework, but the overall reconstruction quality was not satisfactory due to the hand-crafted analysis-synthesis models. 

Recently, deep generative models, such as Variational Auto-Encoding (VAE)~\cite{VAE}, Generative Adversarial Network (GAN)~\cite{goodfellow2014generative}, and Diffusion Model (DM)~\cite{NEURIPS2021_49ad23d1}, have shown remarkable visual synthesis abilities, which can further advance the progress of MBC technology. In particular, face animation/reenactment techniques ~\cite{wiles2018x2face, zakharov2019few, wang2019few, 8954170, siarohin2021motion,hong2022depth} are proposed to economically represent the pixel-level signal with compact facial representations as shown in Fig. \ref{fig1} (b). The high-quality face videos can be further generated by animating a given face image with these compact representations. Herein, these generative models rely on their powerful inference capabilities to facilitate the modeling of compact information, the estimation of complex facial movements, and the synthesis of high-quality facial signal. As such, generative face video compression can be applied towards ultra-low bitrate communication as illustrated in Fig. \ref{fig1} (a). 

Specifically, the First Order Motion Model (FOMM)~\cite{FOMM} proposed to characterize face frames with a series of learned 2D keypoints along with their local affine transformations, achieving the face video animation via the deep generative model. Such an end-to-end face animation framework greatly facilitates the development of face video compression towards ultra-low bit-rate communication~\cite{facebook2021}~\cite{ultralow}~\cite{9859867}. Other facial representations have also been explored for GFVC. Oquab \textit{et al.}~\cite{facebook2021} adopted the SPADE architecture and segmentation maps to develop the first real-time GFVC system on the mobile platform.  Feng \textit{et al.}~\cite{9455985} utilized 2D face landmarks and Chen \textit{et al.}~\cite{CHEN2022DCC} adopted compact feature representation as the transmitted animation symbols to reconstruct talking face videos. Chen \textit{et al.}~\cite{chen2023interactive} utilized the 3DMM model to disentangle the face into semantic representations for interactive bit-stream editing. Moreover, a series of novel techniques, such as frame interpolation~\cite{compressing2022bmvc}, residual enhanced coding~\cite{konuko2023predictive}, multi-view aggregation~\cite{volokitin2022neural}, multi-reference dynamic prediction~\cite{icip2022zhao}, spatial-temporal adversarial training~\cite{chen2023csvt}, progressive feature representation~\cite{CHEN2025DCC}, multi-resolution processing~\cite{10743918} and feature transcoding~\cite{yin2024parametertranslator} are utilized in the GFVC framework to further improve RD performance and expand its application scenarios.

Moreover, JVET has initiated efforts to standardize GFVC since January of 2023, aiming to design a standardized high-level syntax to encapsulate various GFVC features and insert them into the VVC coded bitstream. Notably, a unified GFV Supplementary Enhancement Information (SEI) message syntax~\cite{JVET-AD0051,JVET-AI0191} was introduced to facilitate the carriage of different GFVC feature formats within VVC encoded bitstreams. Additionally, specifications were outlined for the interfaces to GFVC translator, generator, and enhancer neural networks~\cite{JVET-AE0280, JVET-AI0189}. Efforts were also made to ensure interoperability among various GFVC feature formats~\cite{JVET-AG0048}, develop fusion/enhancement modules for quality enhancement~\cite{JVET-AI0194, JVET-AH0127,JVET-AH0110}, and reduce model complexity~\cite{JVET-AG0139} to improve GFVC performance and facilitate practical deployment. More importantly, JVET has established GFVC test conditions~\cite{JVET-AJ2035} and software tools~\cite{JVET-AG0042,JVET-AH0114}, and the related GFV SEI and its enhancement extensions have been finally standardized in the official working draft~\cite{JVET-AJ2006} of Versatile Supplementary Enhancement Information (VSEI) standard.

Our proposed PGen scheme is designed to bridge traditional hybrid coding, model-based coding, scalable coding, end-to-end learning-based coding and generative compression, such that the coding flexibility and competitive RD performance can be well guaranteed for talking face video compression. Firstly, the proposed scheme relies on the traditional hybrid coding framework to compress the key-reference frame, providing strong texture reference for the generation of subsequent inter frames. Secondly, the proposed scheme is made up of multi-granularity facial descriptor and attention-guided generative module, which is a typical analysis-synthesis model-based coding framework. Thirdly, the proposed scheme follows the philosophy of scalable coding, which is able to encode the visual face signals into feature representation at different granularities to support the higher-quality reconstruction towards bandwidth intelligence. Fourthly, the proposed scheme employs the learning-based entropy model to learn the distribution of feature representation and compress these auxiliary information in an end-to-end manner, which leverages the philosophy of end-to-end coding. Fifthly, the proposed scheme is a universal and plug-and-play enhanced coding tool for generative compression, which can remedy the drawbacks in low-fidelity reconstruction quality and limited bitrate coverage. Finally, the proposed scheme was successfully adopted as a VSEI message and included in the JVET Ad hoc Group 16 reference software~\cite{JVET-AH0110}.

\begin{figure}[tb]
\centering
\vspace{-1.2em}
\subfloat[Rate-DISTS]{\includegraphics[width=0.25\textwidth]{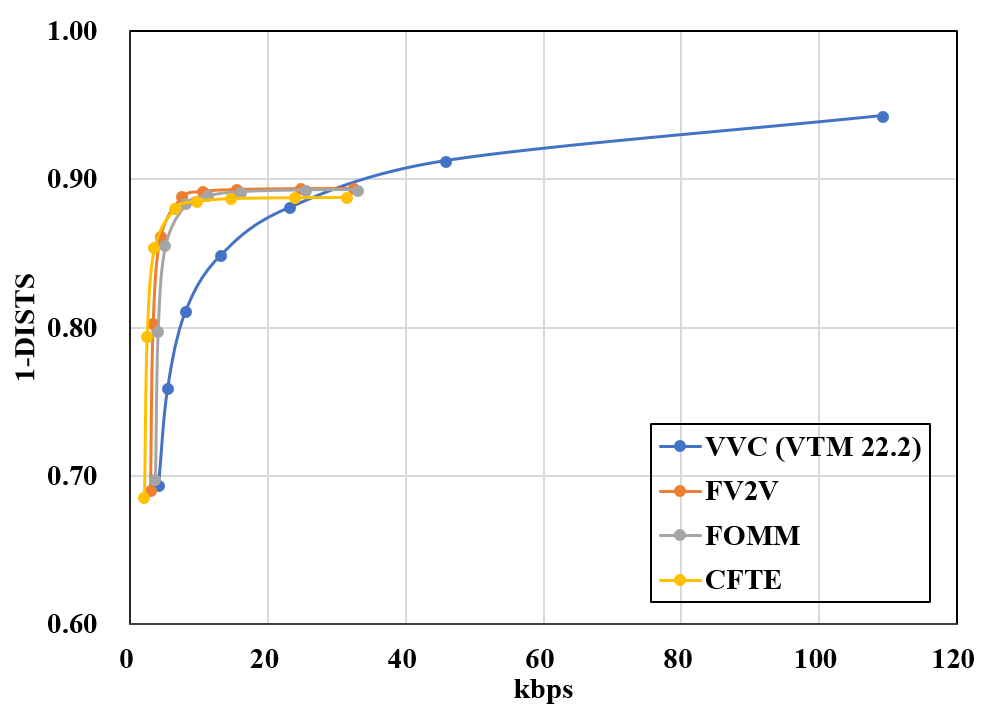}}
\subfloat[Rate-LPIPS]{\includegraphics[width=0.25\textwidth]{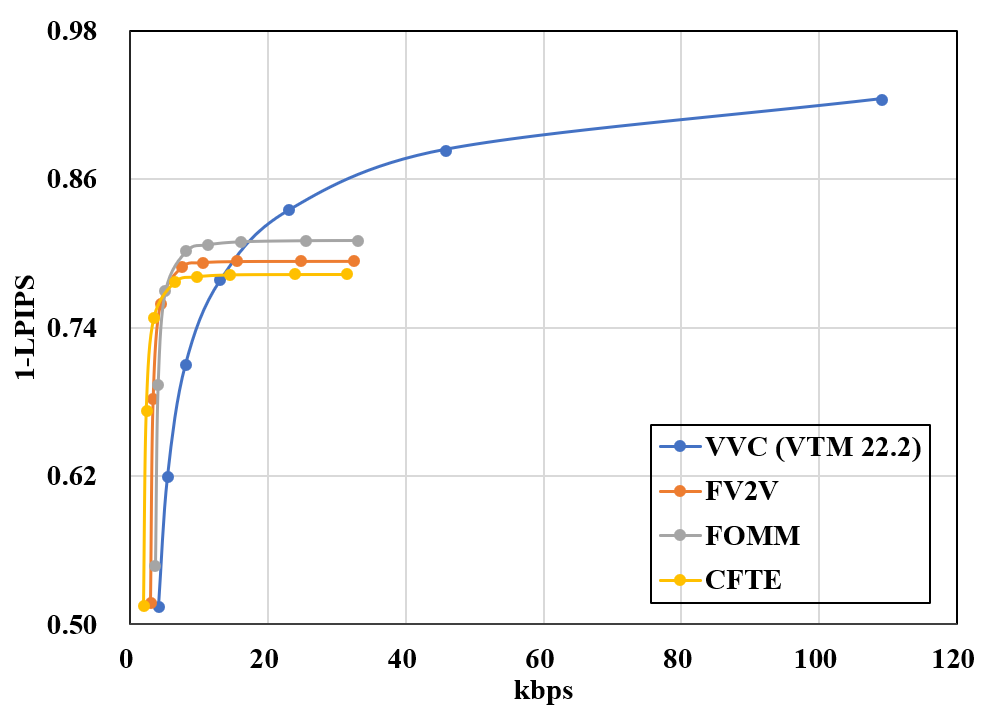}}\\
\subfloat[Rate-PSNR]
{\includegraphics[width=0.25\textwidth]{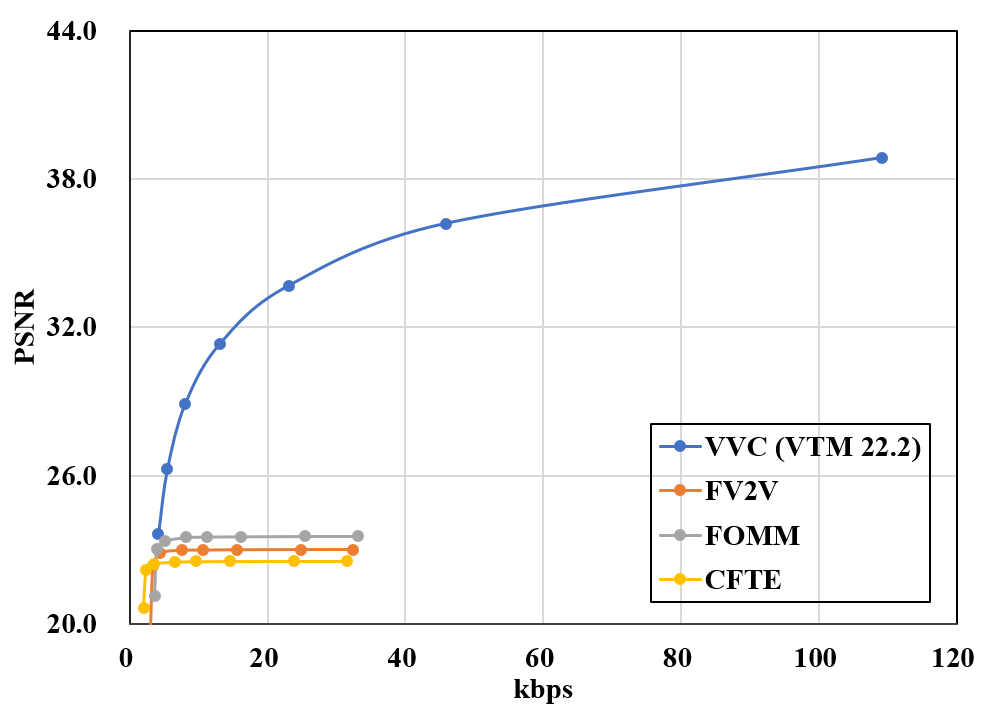}}
\subfloat[Rate-SSIM]
{\includegraphics[width=0.254\textwidth]{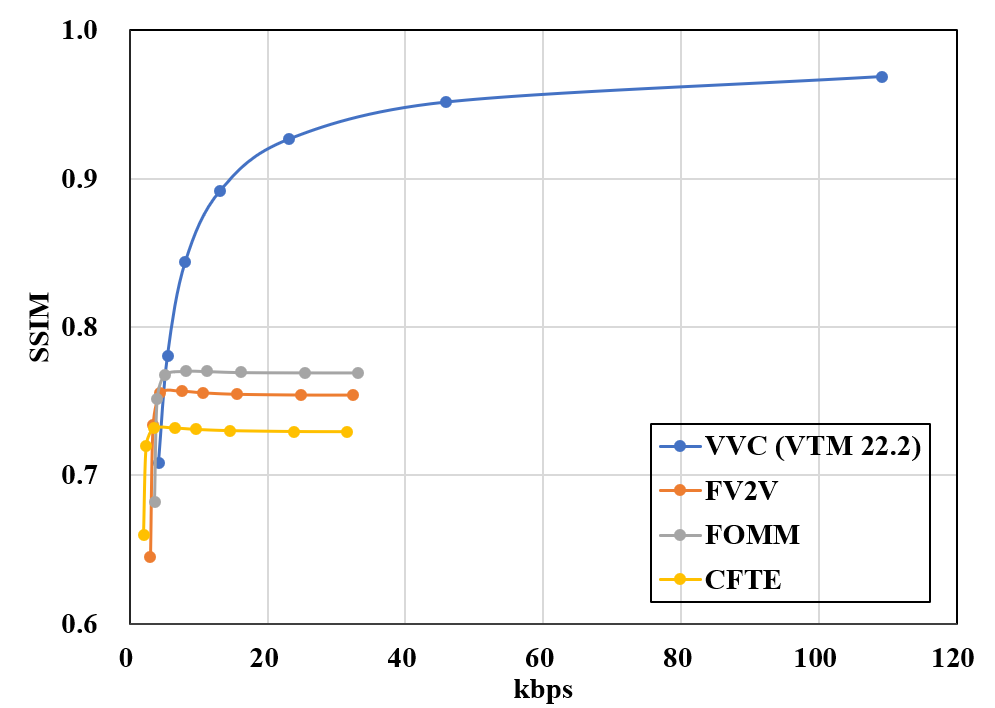}}
\hspace{0.1em}
\caption{Comparisons and analyses of rate-distortion limitations for VVC codec and latest GFVC systems in terms of perceptual-level and pixel-level measures.}
\label{fig3} 
\end{figure}

\begin{figure}[tb]
\centering
\vspace{-1.2em}
\subfloat[Reference]{\includegraphics[width=0.11\textwidth,height=4.3cm]{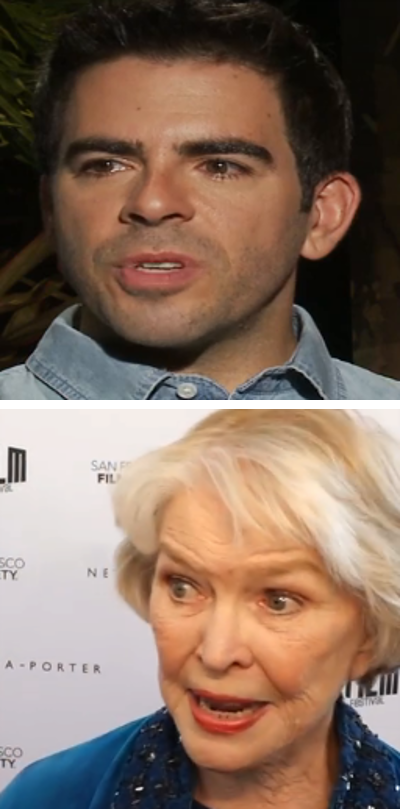}}
\hspace{0.1em}
\subfloat[Generated Occlusions]{\includegraphics[width=0.11\textwidth,height=4.3cm]{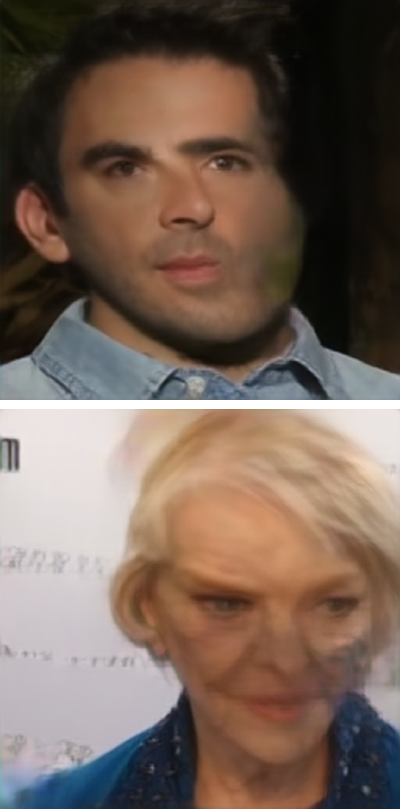}}
\hspace{0.1em}
\subfloat[Low Fidelity]{\includegraphics[width=0.11\textwidth,height=4.3cm]{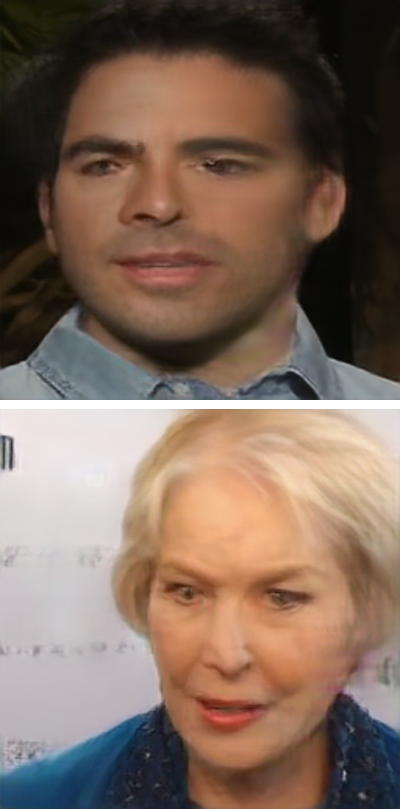}}
\hspace{0.1em}
\subfloat[Poor Local Motion]{\includegraphics[width=0.11\textwidth,height=4.3cm]{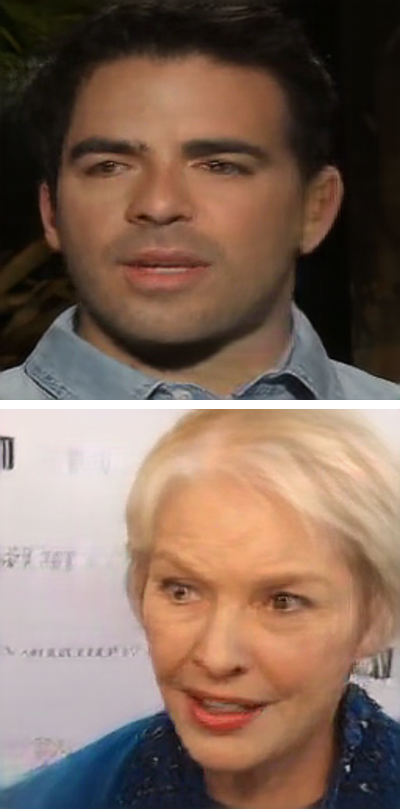}}
\caption{Limitations of visual reconstruction quality for GFVC systems, where these generated faces are sourced from FOMM, CFTE and FV2V.}
\label{fig2} 
\end{figure}

\begin{figure*}[tb]
\centering
\vspace{-2.0em}
\centerline{\includegraphics[width=0.82 \textwidth]{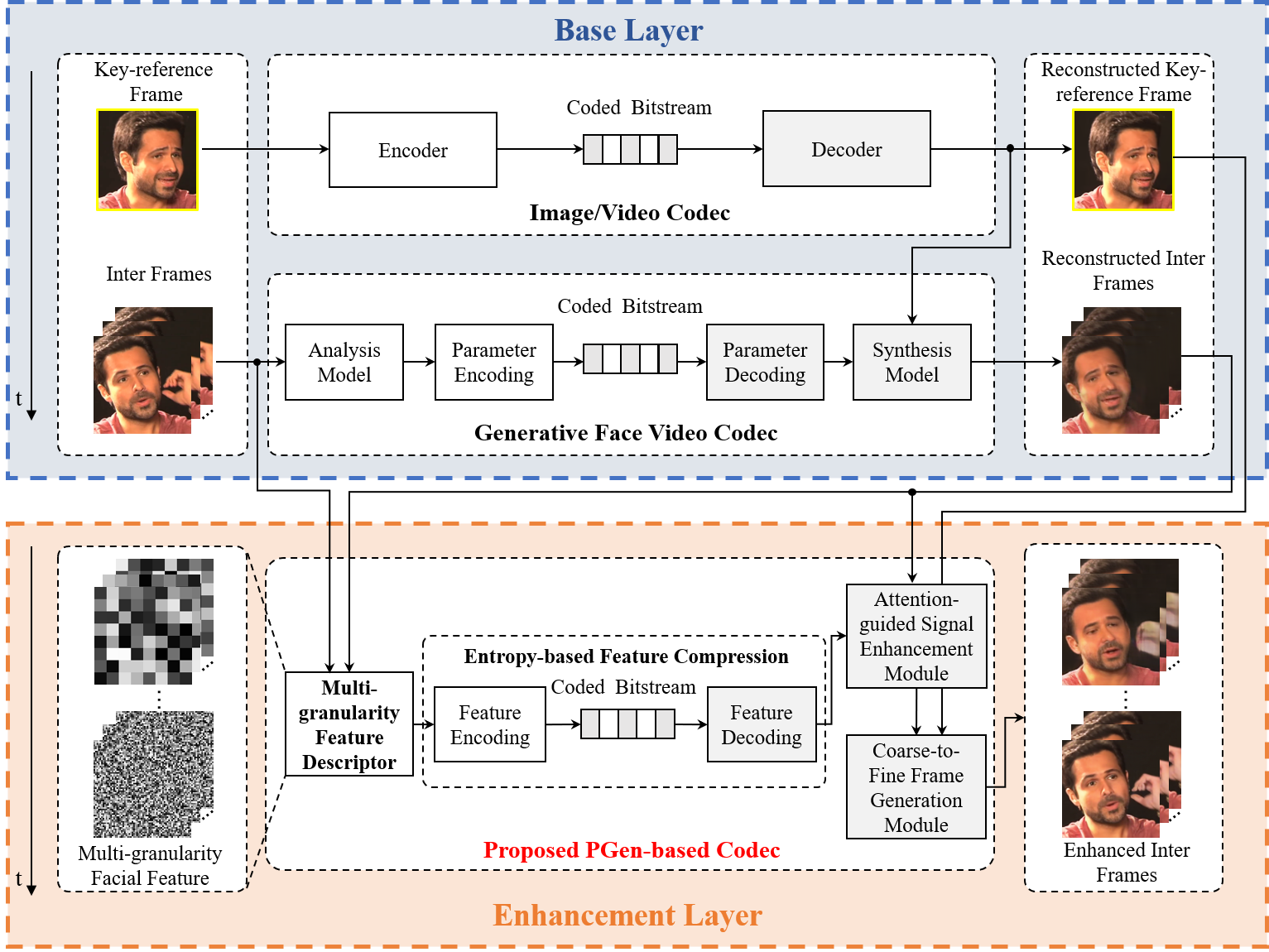}}  
\caption{Overview of the proposed Pleno-Generation (PGen) framework based on scalable representation and layered reconstruction for bandwidth-intelligent face video communication.}
\label{figframework}
\vspace{-1em}
\end{figure*}

\section{Motivations}
\label{section: 1}
Lossy video compression aims at achieving the lowest possible perceived distortion (\textit{i.e.,} \(D\)) given the bitrate budget (\textit{i.e.,} \(R\)). Alternatively, it can be relaxed as a rate-distortion optimization problem aimed at minimizing the overall cost $\jmath _{cost}$ with a trade-off coefficient between \(R\) and \(D\) as follows,
\begin{equation}
\label{eq1}
\begin{array}{c}
{
\jmath _{cost}=  D +\lambda R 
},
\end{array}
\end{equation} where $\lambda$ is the Lagrange multiplier representing the \(R-D\) relationship for a particular quality level. 

Though the RD relationship that distortion decreases along with the increase of bitrate has been regarded as a golden principle in video coding, it is argued that the current generative coding does not follow this rule, as shown in Fig. \ref{fig3}. The specific phenomena and analyses are listed below,

\begin{itemize}
\item{\textbf{Limited bitrate coverage.} The coding overhead of generative compression algorithms generally includes the cost of key-reference frames with vivid texture/color and the compact representations of complex temporal motion. In particular, the bitrate range is mainly determined by the number of the key-reference frames and their degree of compression, instead of by the compact representations for motion. In other words, these generative compression algorithms, which are based on compact representation, have cleverly bridged the gap between generation and compression tasks with distinct goals and trade-offs, such that the current mechanism typically forces these algorithms to operate at one particular bitrate range, \textit{i.e.,} ultra-low bitrate range.\\
\textbf{Evidence:} Fig. \ref{fig3} provides the RD performance for the VVC codec and latest GFVC algorithms in terms of perceptual-level measures (\textit{i.e.,} Deep Image Structure and Texture Similarity (DISTS)~\cite{dists} and Learned Perceptual Image Patch Similarity (LPIPS)~\cite{lpips}) and pixel-level measures (\textit{i.e.,} Peak Signal-to-Noise Ratio (PSNR)~\cite{2009Mean} and Structural Similarity Index Measure (SSIM) ~\cite{wang2004image}), revealing that allocating more coding bits to key-reference frames cannot make optimal RD trade-offs within the whole bitrate coverage; in other words, the available bandwidth cannot be intelligently utilized. More specifically, once the reference-frame quality reaches its peak saturation point, additionally allocated bits cannot facilitate the corresponding quality improvement, violating the rate-distortion relationship. As a result, the current generative compression algorithms are constrained within a limited bitrate range, and a wider bitrate coverage cannot be reached.}

\item{\textbf{Unstable reconstruction quality.} 
Although generative models possess powerful inference capabilities, their robustness in complex motion and long-term dependencies scenes cannot be well guaranteed. In real applications, they could be prone to failure cases with annoying artifacts, missing details and temporal inconsistency. 
Therefore, if the constraints of the generative models are completely ignored for the sake of efficient representation costs, the generative capability will be greatly restricted.
This also explains the reason why it is sometimes difficult to achieve stable visual reconstruction with precise motion and vivid texture from compact feature representations.\\
\textbf{Evidence:} Fig. \ref{fig2} provides annoying  reconstruction examples with the existing GFVC approaches, illustrating that these GFVC algorithms may also suffer from low face fidelity, inaccurate local motion the eye and/or mouth regions, temporal inconsistency, and occlusion artifacts. These unacceptable visual reconstruction results can be easily perceived by human visual system and seriously affect the final RD trade-offs. However, it is worth mentioning that effectively allocating more bits cannot compensate for such distortions, in particular when the generalization ability of generative models is not sufficient to handle complex scenes including large head movements.}
\end{itemize}
Motivated by these daunting challenges, we also perceive promising   opportunities in generative compression, including the optimization on RD, bitrate coverage, and model robustness. In this work, we aim to bridge generative content and effective compression with an innovative and tailored SRLR-based PGen framework. As such, the GFVC can achieve bandwidth intelligence and offer sufficiently high visual quality given enough bitrate.

\section{The Proposed PGen Framework}
In this section, we first give an overall introduction of the proposed PGen framework from the perspective of base and enhancement layers. In addition, we provide the details of GFVC algorithms with compact representations and introduce how to regard the GFVC as the base layer to achieve face video compression. Moreover, a universal enhanced coding PGen framework is proposed as the enhancement layer, which can deliver different-granularity facial signals and improve the reconstruction quality of face videos with bandwidth intelligence. Finally, we provide optimization details for the proposed PGen framework to extend the bitrate ranges and improve the reconstruction quality.

\subsection{Overview} 
The proposed PGen scheme is well rooted in the widely accepted view that face frames own prior statistics and can be automatically characterized into meaningful representations, such as latent code (\textit{i.e.,} keypoints, facial semantics and compact feature) and enriched signal (\textit{i.e.,} residual signal, segmentation map and dense flow). These facial representations also align with the recent neuroscience hypothesis that HVS can achieve straight transformation from overall visual signals to key motion trajectories~\cite{henaff2019perceptual}. The proposed SRLR-based PGen framework for intelligent-bandwidth face video communication is shown in Fig. \ref{figframework}, consisting of base and enhancement layers. 

More specifically, the base layer can be compatible with all existing GFVC algorithms and achieve ultra-low bitrate face video communication with compact representations. The face data is characterized with latent code (\textit{i.e.,} keypoints, facial semantics and compact feature) at the encoder side, and the extracted information is further compressed and conveyed to reconstruct faces via the deep generative model at the decoder side. Afterwards, the enhancement layer is capable of providing enriched facial signal and support high-quality face video communication when bandwidth permits. The encoder side can further characterize visual face data with different-granularity facial signals and compress them into the bitstream, whilst the decoder receives the bitstream and exploits these decoded auxiliary facial signals to improve the reconstruction quality of the base-layer output frames. The details of base and enhancement layers will be introduced in subsequent subsections.

The proposed PGen framework enjoys several potential benefits and enables promising face communication scenarios, where the corresponding analyses are listed as follows,  
\begin{itemize}
\item{\textbf{Coding flexibility.} The face data is able to be characterized with compact representation and enriched signal, which can be naturally incorporated into a unified PGen framework. In particular, these conceptually explicit visual information can be packed into the segmentable and interpretable bitstream, such that they can be partially transmitted and decoded to actualize multi-layer visual reconstructions IN the specific bandwidth environment. As such, the coding flexibility can be well guaranteed. }
\item{\textbf{Perceptual quality improvement.} The proposed PGen framework can well overcome the reconstruction quality limitations of the existing GFVC algorithms such as occlusion artifacts, low face fidelity and poor local motion. In particular, with the guidance of auxiliary visual signal, the base-layer motion estimation errors can be perceptually compensated and the long-term dependencies among face frames can be accurately regularized. As a consequence, the enhancement-layer output can greatly improve the reconstruction quality at the pixel-level and with faithful representation of texture and motion.}
\item{\textbf{Universal plug-and-play component.}  The proposed PGen framework strictly follows the scalable philosophy and includes the base and enhancement layers. In particular, the enhancement layer is designed as a universally plug-and-play component to warrant the service of the base layer that is compatible for all existing GFVC algorithms. In addition to the flexibility of external components, the internal mechanism of the enhancement layer is very flexible, which can realize multi-granularity signal representation and support multi-quality face video communication according to the bandwidth environment. }
\end{itemize}

\subsection{Base Layer} 
The base layer in the proposed SRLR framework can achieve ultra-low bitrate face video compression based upon compact representation (\textit{i.e.,} keypoints, facial semantics and compact feature), where it refers to the existing GFVC algorithm (\textit{i.e.,} FOMM~\cite{FOMM} FV2V\cite{wang2021Nvidia}, CFTE~\cite{CHEN2022DCC} and etc.). As shown in Fig. \ref{fig1}, it includes the encoding and decoding processes. More specifically, the encoder in the base layer mainly consists of three modules: an intra-coding tool for compressing the key-reference frame, an analysis model for representing key motion trajectories of inter frames, and a parameter encoding module for high-efficiency parameter compression. Regarding the decoder of the base layer, it is also composed of three main parts, including a decoding scheme for reconstructing the key-reference frame, a parameter decoding module for the reconstruction of compact representation, and a synthesis model for the generation of the final video. 

For a face video clip, it can be divided into a key-reference frame $K$ and subsequent inter frames $I_{l} \left (1\le l \le n , l\in Z   \right ) $. First, the key-reference frame $K$ is compressed with the VVC codec to establish the fundamental texture representation for the reconstruction of successive frames. The reconstruction of key-reference frame $\hat{K}$ can be denoted as follows,
\begin{equation}
\hat{K}=Dec \left ( Enc\left ( K \right )  \right )  ,
\end{equation}  where $Enc\left ( \cdot  \right )$ and  $Dec\left ( \cdot  \right )$ represent the encoding and decoding processes of the VVC codec. The VVC codec is chosen as the intra-coding tool lies in that it is the latest video coding standard with the most promising coding performance. 

Afterwards, the subsequent inter frames $I_{l}$ are fed into an analysis model in order to characterize the dynamic trajectory variations in a compact and latent manner, where the extraction of $\mathbb{F} _{comp}^{I_{l}}$ can be formulated by,
\begin{equation}
\mathbb{F} _{comp}^{I_{l}} = \varphi\left ( I_{l} \right ) ,
\end{equation} where $\varphi\left (\cdot  \right )$ denotes the analysis process that evolves the prior statistics of face data into the compact representations (\textit{i.e.,} keypoints, facial semantics and compact feature) via the analysis model. Such a compact representation learning mechanism can lead to reduced coding bits and facilitate face video compression at very low bitrate. Subsequently, the extracted compact representations $\mathbb{F} _{comp}^{I_{l}}$ are inter predicted, quantized and entropy coded into bitstream via the parameter encoding module. When the decoder receives the bitstream of compact representations, the related parameter decoding process (\textit{i.e.,} inverse quantization and compensation) is executed to reconstruct compact representation $\hat{\mathbb{F}} _{comp}^{I_{l}}$.

Finally, the face video can be reconstructed based upon the reconstructed compact representation $\hat{\mathbb{F}} _{comp}^{I_{l}}$ and decoded key-reference frame $\hat{K}$ with the frame synthesis model. More specifically, $\hat{K}$ is further projected into its compact representation $\mathbb{F} _{comp}^{\hat{K}}$ via the analysis model, such that the generation of motion filed can be yielded between $\mathbb{F} _{comp}^{\hat{K}}$ and $\hat{\mathbb{F}} _{comp}^{I_{l}}$. As such, the frame synthesis model can well exploit the strong inference ability of deep neural networks to synthesize high-quality video frames $\hat{I_{l}}$ with the guidance of $\hat{K}$. The processes can be given by,
\begin{equation}
\hat{I_{l}} =\zeta \left ( \hat{K} ,\chi \left ( \hat{\mathbb{F}} _{comp}^{I_{l}} , \varphi\left ( \hat{K} \right ) \right )  \right ) ,
\end{equation} where $\chi\left ( \cdot  \right )$ and  $\zeta\left ( \cdot  \right )$ represents the motion field calculation and frame generation processes, respectively.

\begin{figure*}[tb]
\vspace{-1.8em}
\centering
\includegraphics[width=1.0\textwidth]{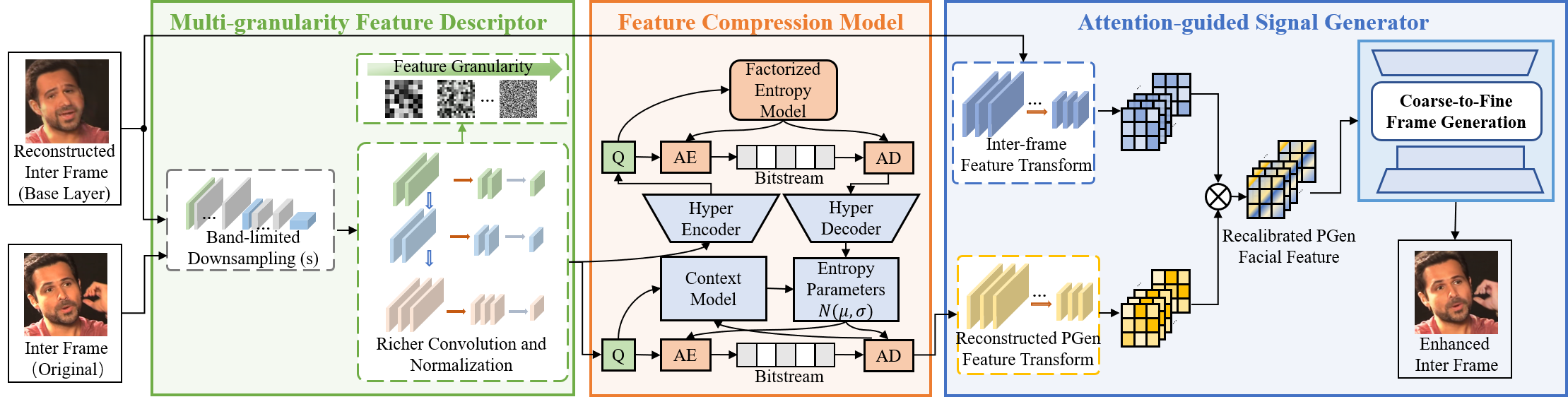} %
\caption{The detailed enhancement-layer architecture to achieve multi-granularity facial feature description, entropy-based feature compression and attention-guided signal generation (\textit{i.e.,} attention-guided face signal enhancement and coarse-to-fine frame generation). In particular, Q, AE, AD and Conv denote the quantization, arithmetic encoding, arithmetic decoding and convolution processes, respectively.}
\vspace{-1em}
\label{enhancement}
\end{figure*}

Through a series of operations in the base layer, face video can be compressed at ultra-low bitrate due to the compact representation. However, the quality of reconstructed face video could contain unappealing texture and motion, such as occlusion artifacts, low face fidelity and poor local motion representation, which c could result in low perceptual quality.

\subsection{Enhancement Layer}

Fig. \ref{enhancement} provides the details of the proposed universal and unified enhanced coding framework that can improve the reconstruction quality and signal fidelity of base-layer face frames via enriched facial signal description. More specifically, a multi-granularity feature representation mechanism is first introduced to hierarchically compensate for motion estimation errors according to different bandwidth environments. Then, a learning-based entropy model is employed to improve the compression efficiency of auxiliary facial feature via hyper priors of base layer. Finally, the attention-guided generative adversarial network with the coarse-to-fine generation strategy is proposed to reconstruct more reliable and robust face video based on the base-layer face frames and the reconstructed facial signals. 

\subsubsection{Multi-granularity Feature Representation} 
In order to solve the problems of inaccurate motion estimation and unstable reconstruction quality caused by the compact representations of GFVC algorithms, auxiliary facial feature mechanism is introduced in a multi-granularity and bandwidth-intelligent manner. 

More specifically, the subsequent inter frames $I_{l}$ are fed into a band-limited down-sampling module~\cite{FOMM} using the anti-aliasing mechanism and padding/convolution operations, which can better preserve the input signal when down-sampling. Afterwards, a UNet-like network~\cite{RFB15a}~\cite{Hourglass2016} is employed to realize the transformation from the input face image to high-dimensional face feature map. In addition, the richer convolutional architecture and Generalized Divisive Normalization (GDN)~\cite{J2015Density} operations are utilized to fully exploit multi-level information and parametric nonlinear transformation from high-dimensional feature map, such that these extracted features can be further combined in a holistic manner and better compensate the information loss of compact representations. The process can be given by,
\begin{equation}
S_{I_{l}} = g_{\left (conv,GDN  \right )} \left (f_{UNet}\left ( \nu \left ( I_{l},s \right )  \right )\right ),
\end{equation} where $\nu \left ( \cdot \right )$, $f_{UNet}\left ( \cdot \right )$ and $g_{\left (conv,GDN  \right )}\left ( \cdot \right )$ denote signal down-sampling, high-dimensional feature learning and multi-level feature extraction processes, respectively. Besides, $s$ is the scale factor to determine the spatial size of face image. Herein, the granularities of auxiliary facial feature $S_{I_{l}}$ are $32\times32\times1$,  $16\times16\times1$ and $8\times8\times1$.

Our proposed multi-granularity feature representation mechanism fully takes into account the shortcomings of the existing GFVC algorithm, thereby delivering auxiliary feature to describe the important motion information (\textit{e.g.,} expression and headpose) and complex background changes in a more scalable and flexible manner. As such, it can well characterize signals at different granularity levels according to different bandwidth environments.

\subsubsection{Entropy-based Feature Compression} 
The orange part in Fig. \ref{enhancement} provides an overview of our optimized entropy model for the compression of multi-granularity auxiliary feature. Inspired by the success of autoregressive priors and hierarchical entropy models, Minnen \textit{et al.}~\cite{NEURIPS2018_53edebc5} developed a joint auto-regressive and hierarchical priors for learned image compression, which can greatly improve compression performance and optimize the end-to-end training. As such, we train the entropy model with joint auto-regressive and hierarchical priors to achieve the high-efficiency compression of auxiliary facial feature $S_{I_{l}}$.
In details, the core structure of the feature compression model includes a context model, an auto-regressive model over latents, and a hyper-network (Hyper Encoder and Hyper Decoder blocks) to learn a probabilistic model for entropy coding of auxiliary feature $S_{I_{l}}$, which can correct the context-based predictions and reduce coding bits. 
$S_{I_{l}}$ is passed through quantizer (\textit{i.e.,} Q) and arithmetic encoder (\textit{i.e.,} AE) to produce the coded bitstream, and then the bitstream is decoded with the arithmetic decoder (\textit{i.e.,} AD) to reconstruct the feature. To improve the compression performance, side information is additionally introduced to assist decoding. More specifically, the variance $\sigma $ of the Gaussian distribution can be predicted from the base-layer auxiliary feature $S_{\hat{I}_{l}}$ without any entropy coding process via the hyper-network. On the other hand, the context model can facilitate the reconstruction of the mean value $\mu$, where it can be further combined with the predicted variance $\sigma $ to simulate the Gaussian distribution and finally assist the decoder to reconstruct the auxiliary facial feature $\hat{S_{I_{l}}}$. 

\subsubsection{Attention-guided Signal Generator} 
Herein, we propose an attention-guided generative adversarial network with the coarse-to-fine generation strategy to facilitate  more reliable and robust face signal reconstruction. 

\textbf{Attention-guided Signal Enhancement:} To effectively exploit the reconstructed auxiliary facial feature $\hat{S_{I_{l}}}$ to boost the generation quality of the base-layer-reconstructed face $\hat{I_{l}}$, we propose an attention-guided signal enhancement module that can better compensate for the motion estimation error and preserve facial structure and background information. More specifically, the reconstructed face frame $\hat{I_{l}}$ from the base layer is first transformed into facial features $F_{if}$, whilst the signal feature $F_{fs}$ with the same feature dimension as $F_{if}$ is also obtained by transforming the auxiliary facial signal $\hat{S_{I_{l}}}$. Afterwards, $F_{if}$ will further execute a Hadamard product operation with $F_{fs}$, aiming to recalibrate the features of the base-layer reconstruction with more accurate motion features. The specific processes can be described as follows,
\begin{equation}
F_{rec}=F_{if}\odot F_{fs},
\end{equation} where $\odot$ is a Hadamard product function.

\begin{figure}[tb]
\centering
\centerline{\includegraphics[width=0.49 \textwidth,height=5.2cm]{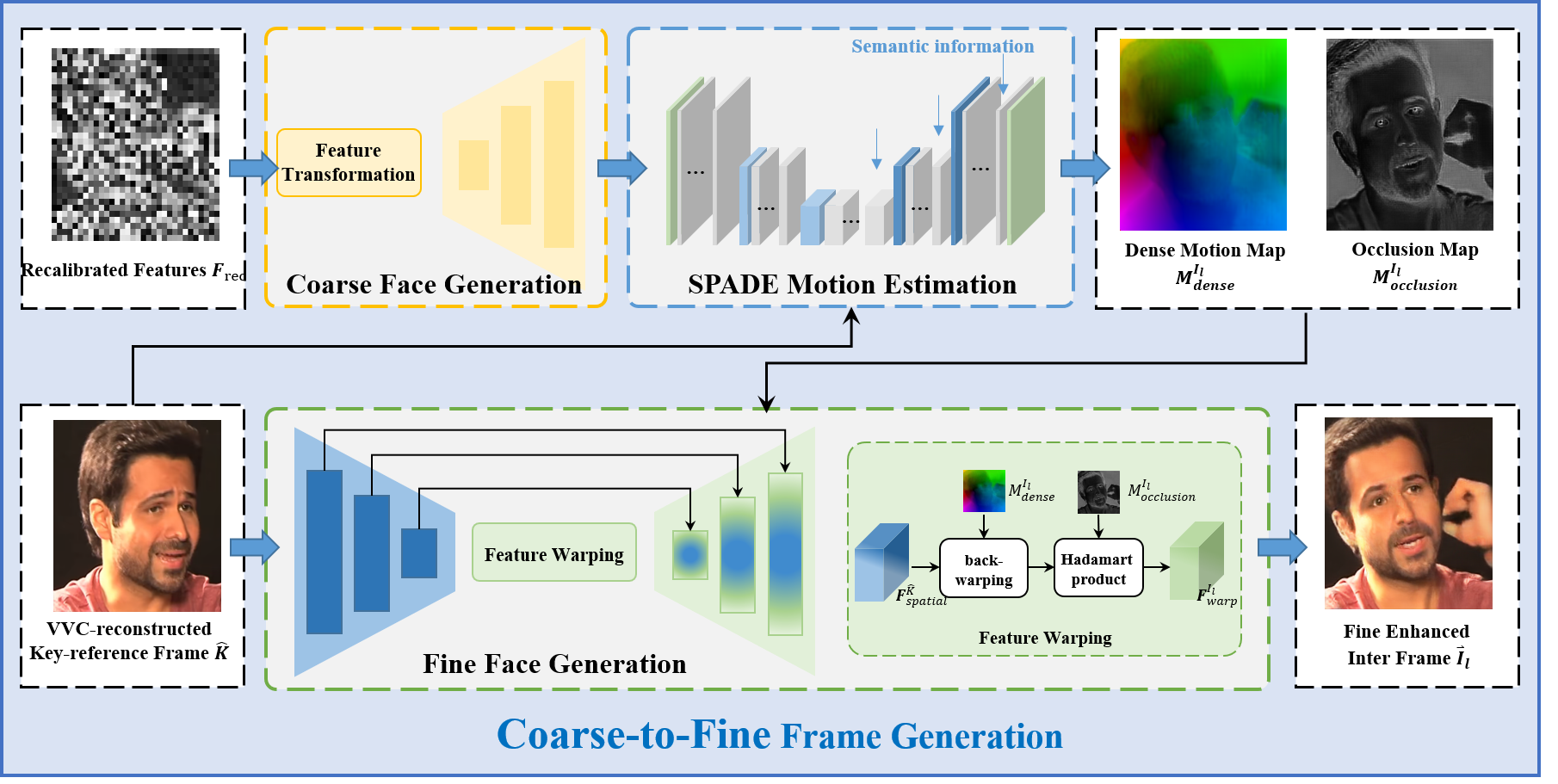}}
\caption{Overview of our proposed coarse-to-fine frame generation module for accurate texture appearance, head posture, and facial expression.}
\label{generative}
\vspace{-0.5em}
\end{figure} 
\textbf{Coarse-to-Fine Frame Generation:} As shown in Fig. \ref{generative}, we adopt the generative adversarial network that possesses strong inference capability to reconstruct high-fidelity face results. The employed GAN includes the generator and discriminator. Analogous to ~\cite{FOMM}~\cite{wang2021Nvidia}, the recalibrated feature $F_{rec}$ from the attention-guided face enhancement module is fed into the decoder of U-Net architecture~\cite{RFB15a}~\cite{Hourglass2016} to produce the coarsely enhanced inter frames $\tilde{I_{l}}$ that have accurate motion information but lossy texture reference. This process is denoted as follows,
\begin{equation}
{
\tilde{I_{l}}= G_{frame} \left ( F_{rec} \right )
},
\end{equation} where  $G_{frame}\left (\cdot \right )$ is the subsequent network layers of U-Net. 

Considering the recalibrated feature $F_{rec}$ could be mixed with  inaccurate base-layer texture representation, we propose a coarse-to-fine generative strategy to ensure that the final generation result has high-quality texture reference without any error accumulation. More specifically, the reconstructed key-reference frame $\hat{K}$ and the coarse-generated inter frame $\tilde{I_{l}}$ are further jointly fed into a Spatially-Adaptive Normalization (SPADE) network~\cite{park2019SPADE} (\textit{i.e.,} $\mathbb{S}_{\mathbb{PADE}}\left ( \cdot \right )$) to better preserve semantic information and constantly utilize this semantic information to learn accurate motion estimation field (\textit{i.e.,} dense motion map $\mathcal{M}^{I_{l}}_{dense} $ and facial occlusion map $\mathcal{M}^{I_{l}}_{occlusion} $) as follows,
\begin{equation}
\mathcal{M}^{I_{l}}_{dense}=P_{1}\left ( { \mathbb{S}_{\mathbb{PADE}}}  \left ( { \mathrm{concat} }  \left (\hat{K},\tilde{I_{l}}  \right ) \right )  \right ),
\end{equation}
\begin{equation}
\mathcal{M}^{I_{l}}_{occlusion}=P_{2}\left ( { \mathbb{S}_{\mathbb{PADE}}}  \left ( { \mathrm{concat} }  \left (\hat{K},\tilde{I_{l}}  \right ) \right )  \right ),
\end{equation}  where $P_{1}\left ( \cdot \right )$ and $P_{2}\left ( \cdot \right )$ represent two different predicted outputs, and  ${ \mathrm{concat} }  \left (\cdot \right ) $ denotes the concatenation operation.

After obtaining the explicit motion information, a multi-scale GAN module is further utilized for realistic talking face reconstruction. First, the U-Net encoder is able to transform the VVC-reconstructed key-reference frame $\hat{K}$ into multi-scale spatial features $F_{spatial}^{\hat{K}} $. Then, the dense motion field $\mathcal{M}^{I_{l}}_{dense}$ and facial occlusion map $\mathcal{M}^{I_{l}}_{occlusion}$ execute the feature warping operation with these multi-scale spatial features $F_{spatial}^{\hat{K}} $ to obtain multi-scale transformed feature $F_{warp} $.
\begin{equation}
{
F^{I_{l}}_{warp} =  \mathcal{M}^{I_{l}}_{occlusion} \odot  f_{w}\left ( F_{spatial}^{\hat{K}} , \mathcal{M}^{I_{l}}_{dense} \right ) 
},
\end{equation} where $f_{w}\left (\cdot \right )$ represents the feature warping process.

Finally, these multi-scale warped features $F_{warp}$ are fed into the U-Net decoder to reconstruct high-quality and high-fidelity face as follows,
\begin{equation}
{
\vec{{I_{l}}} = G_{frame} \left ( F^{I_{l}}_{warp} \right )
},
\end{equation} where the discriminator is further used to guarantee that $\vec{{I_{l}}}$ can be reconstructed towards a realistic image. 

\subsection{Optimization}
During our model training, the self-supervised training strategy is adopted to optimize multi-granularity signal descriptor, learning-based entropy model, attention-guided signal enhancement and coarse-to-fine frame generation module. The corresponding loss objectives during model training include perceptual loss $\mathcal L_{per-coarse}$/$\mathcal L_{per-fine}$, adversarial loss $\mathcal L_{adv}$, identity loss $\mathcal L_{id}$, optical flow loss $\mathcal L_{flow}$ and rate-distortion loss $\mathcal L_{RD}$. Herein, the overall training loss can be summarized as follows,
\begin{equation}
\begin{aligned}
\begin{array}{c}
\mathcal L_{total}=\lambda_{per-coarse}\mathcal L_{per-coarse}+\lambda_{per-fine}\mathcal L_{per-fine}\\+ \lambda_{RD}\mathcal L_{RD}+ \lambda_{adv}\mathcal L_{adv}+\lambda_{id}\mathcal L_{id} + \lambda_{flow}\mathcal L_{flow},
\end{array}
\end{aligned}
\end{equation} where the hyper-parameters are set as follows: $\lambda _{per-coarse}=50$, $\lambda _{per-fine}=50$, $\lambda _{adv}=10$, $\lambda _{id}=40$ and $\lambda _{flow}=20$. As for $\lambda _{RD}$, it can be set according to the size of multi-granularity signal, \textit{i.e.,} when the feature sizes are $32\times32\times1$,  $16\times16\times1$ and $8\times8\times1$, $\lambda _{RD}$ are empirically set at 1500, 268 and 64, respectively. It should be mentioned that the perceptual loss term is used twice between the coarse and fine prediction results $\tilde{I_{l}}$/$\vec{I_{l}}$ and the original inter frame $I_{l}$.

\begin{figure*}[!t]
\centering
\subfloat[VoxCeleb]{\includegraphics[width=0.32 \textwidth]{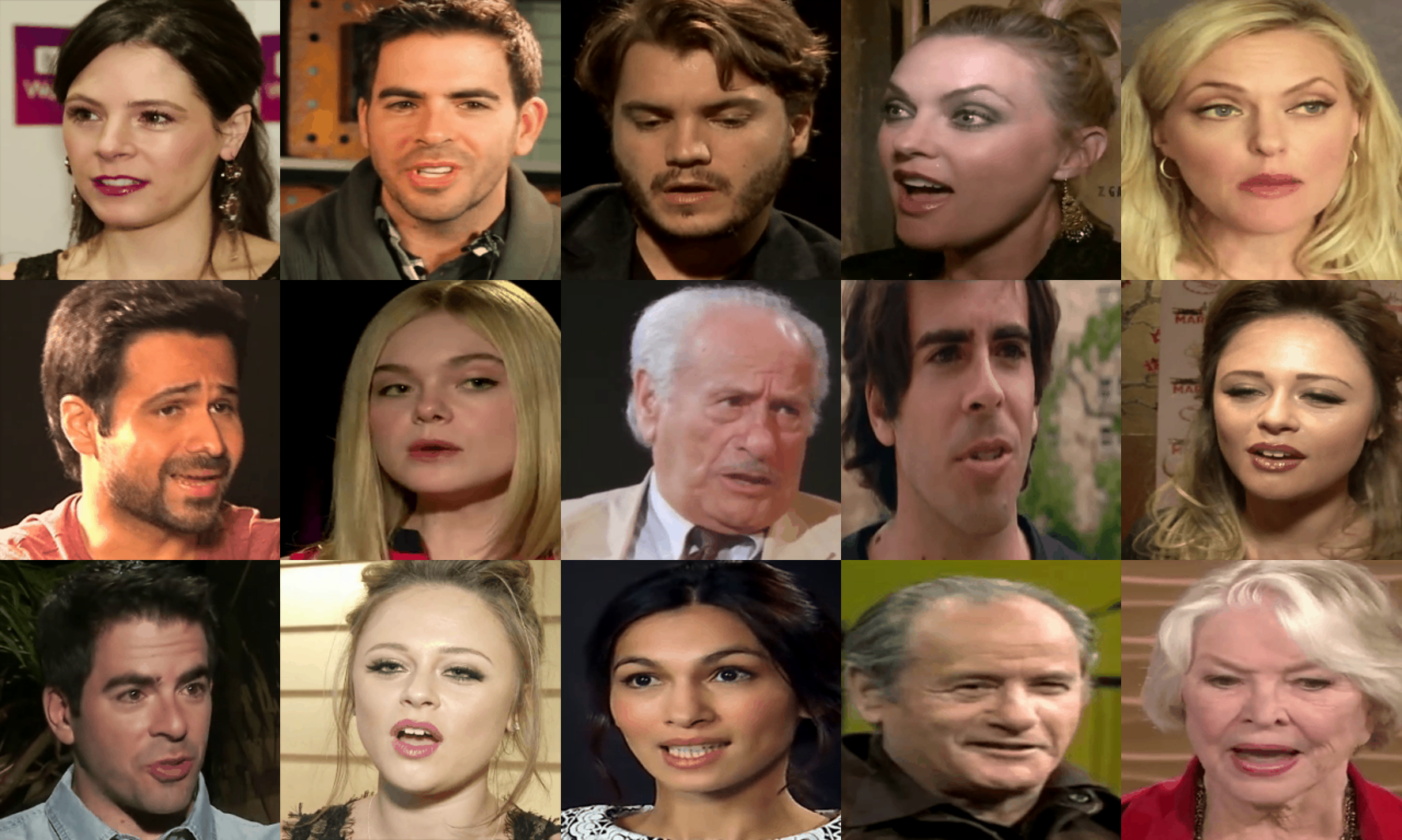}}
\hfil
\subfloat[300-VW]{\includegraphics[width=0.32 \textwidth]{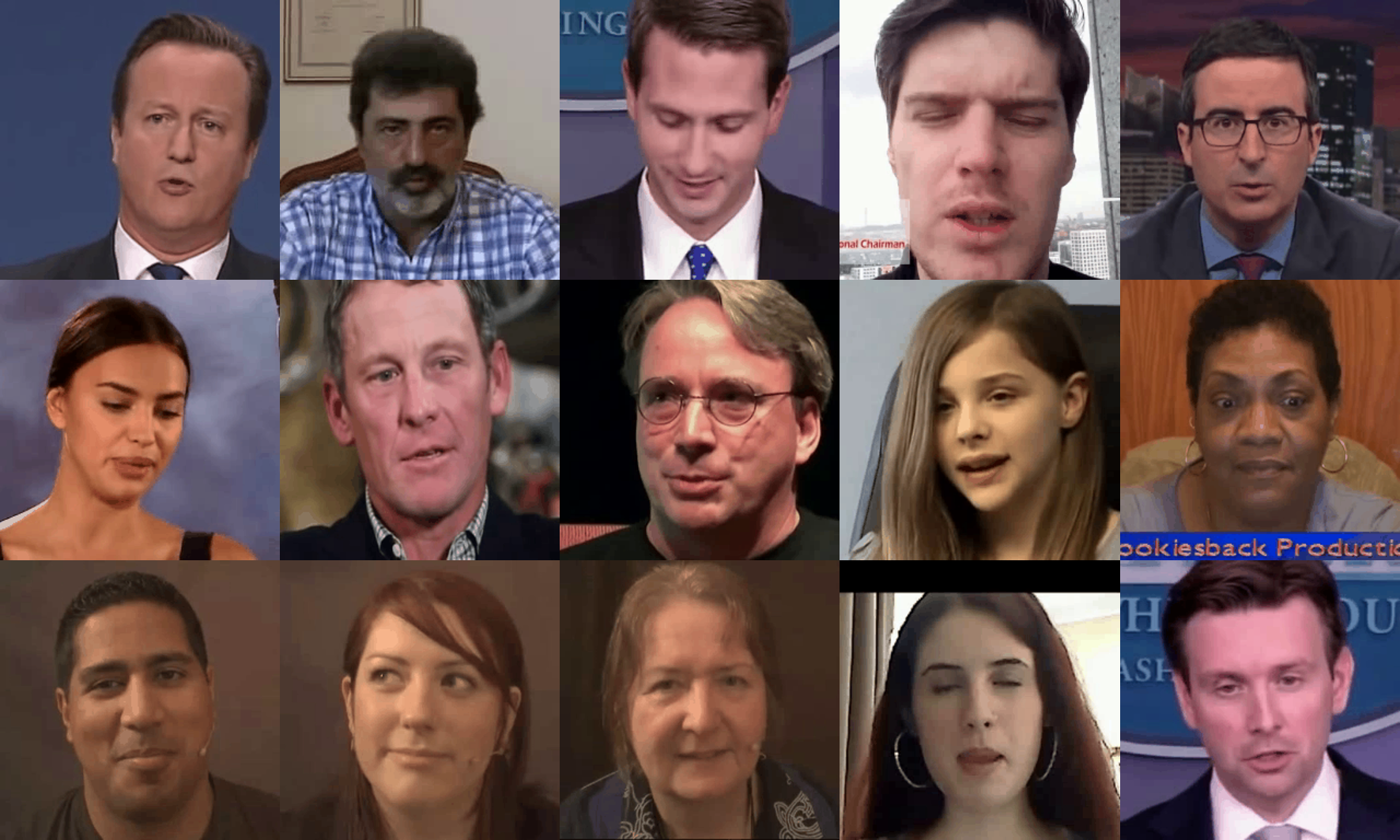}}
\hfil
\subfloat[CFVQA]{\includegraphics[width=0.32 \textwidth]{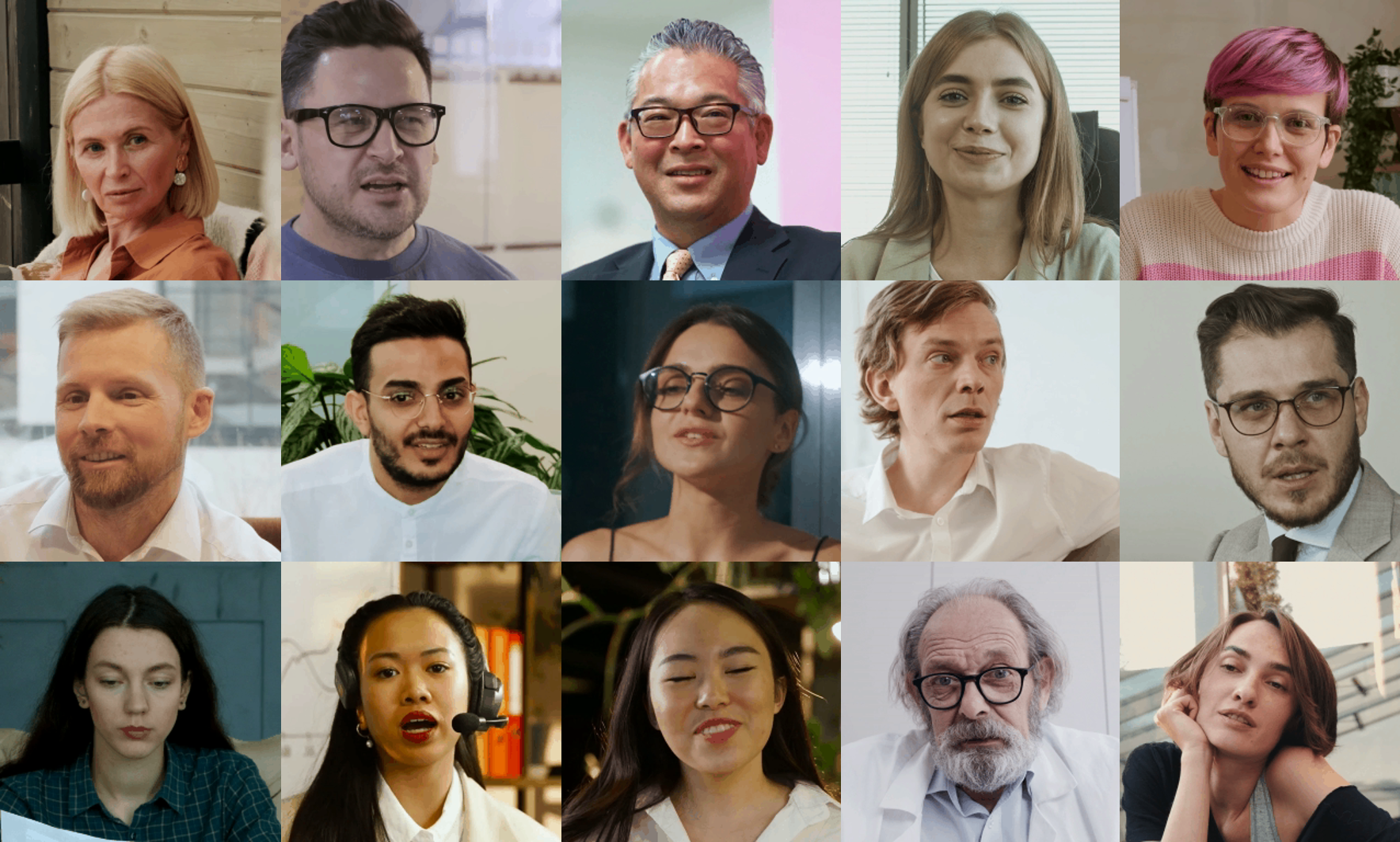}}
\caption{Visual samples of testing sequences selected and pre-processed from VoxCeleb testing dataset~\cite{Nagrani17}, 300-VW~\cite{300VW} and CFVQA~\cite{li2023perceptual}.}
\label{dataset}
\end{figure*}

\section{Experiments}
In this section, we first give a detailed introduction of experimental settings. Besides, the performance comparisons and analysis are provided to verify the effectiveness of the proposed PGen scheme.

\subsection{Experimental Settings}

\subsubsection{Implementation Details}
We implement our proposed PGen scheme on NVIDIA TESLA A100 GPUs with on Pytorch libraries. The Adam optimizer is used within the parameters $\beta _{1}$ = 0.5 \& $\beta _{2}$= 0.999, and the learning rate is set at 0.0002 for model convergence. In addition, the epochs of model training are 200 via the synchronized BatchNorm initialization and the data repeating strategy. 

\subsubsection{Experimental Datasets}
The proposed PGen model is trained with mixed training data from the pre-processed VoxCeleb training dataset~\cite{Nagrani17} and CelebV-HQ training dataset~\cite{zhu2022celebvhq}, which respectively focus on face-centric and head-and-shoulder content at the resolution of 256$\times$256.
For the testing dataset, we use 45 face video sequences selected and pre-processed from the VoxCeleb testing dataset~\cite{Nagrani17}, 300-VW~\cite{300VW} and CFVQA~\cite{li2023perceptual} as shown in Fig. \ref{dataset}. These selected sequences focus on face-centric as well as head-and-shoulder content, and span a considerable diversity in terms of age, gender, ethnicity, appearance, expression, head position, head motion, camera motion, background, etc. Each of these testing sequences has the resolution of 256×256 with the color format of RGB 444. The VoxCeleb and 300-VW sequences are 10-second long (\textit{i.e.,} 250 frames at 25 fps), and the CFVQA sequences are 5-second long (\textit{i.e.,} 125 frames at 25 fps). It should be mentioned that the VoxCeleb and CFVQA testing datasets have been included as test material by JVET for the generative face video compression experiments of Ad hoc Group 16~\cite{JVET-AJ2035}.

\subsubsection{Evaluation Measures}
To provide comprehensive objective quality evaluations from the perspective of pixel-level, perceptual-level, temporal consistency and no-reference GAN-based, the following objective quality measures are used,
\begin{itemize}
\item{\textbf{Pixel-level measures:} PSNR~\cite{2009Mean} and SSIM~\cite{wang2004image}) are traditional full-reference quality metrics used in compression tasks for pixel-level distortion calculation. It is reported in~\cite{compressionmeasure} that PSNR and SSIM might not be suitable for assessing GAN-based compression results with feature-level alignment.}
\item{\textbf{Perceptual-level measures:} DISTS~\cite{dists} and LPIPS~\cite{lpips} are learning-based full-reference quality metrics that could learn the characteristics of human visual perception via a deep neural network and map correlations of the feature maps to evaluate the perceptual-level quality. As such, they may be more appropriate for the GAN-based image evaluation. For these measures, lower scores are associated with better perceived picture quality. }
\item{\textbf{Temporal consistency measure:} Fréchet Video Distance (FVD)~\cite{Unterthiner2019FVDAN} is temporal-domain metric that considers a distribution over videos on the basis of Fréchet inception distance (FID)~\cite{heusel2017gans}, which can correlate well with qualitative human judgment of generated videos. For FVD, the perceived image quality is higher when the score is smaller.}
\item{\textbf{No-reference GAN-based measure:} Multi-dimension Attention Network for No-reference Image Quality Assessment (MANIQA)~\cite{yang2022maniqa} improves the performance on GAN-based distortion in the no-reference domain. It has achieved the best performance in the NTIRE 2022 Perceptual Image Quality Assessment Challenge. For this measure, the perceived image quality is higher when the score is larger.}
\end{itemize}

In addition to these quality evaluation measures, we also calculate the consumed bitrate (\textit{i.e.,} kbps) in video compression task, such that the rate-distortion curve (\textit{i.e.,} RD-curve) and Bjøntegaard-delta-rate (\textit{i.e.,} BD-rate)~\cite{Bjntegaard2001CalculationOA} can be summarized for the evaluation of compression performance.

\subsubsection{Testing Conditions}
To validate the compression performance, the latest video coding standard VVC~\cite{bross2021overview} is adopted as the compression anchor. In addition, since our PGen algorithm is designed to extend the bitrate range and improve the reconstruction quality of all existing GFVC algorithms, we select three representative generative compression schemes with different facial feature representations as the base-layer algorithm, including FOMM~\cite{FOMM}, FV2V~\cite{wang2021Nvidia} and CFTE~\cite{CHEN2022DCC}. The testing settings are provided as follows, 
\begin{itemize}
\item{\textbf{VVC anchor:} is configured with the Low-Delay-Bidirectional (LDB) mode in VTM reference software 22.2, where the Quantization Parameters (QPs) are set at 27, 32, 37, 42, 47 and 52 to take into account different bitrate ranges.}
\item{\textbf{GFVC anchors:} strictly follow experimental testing specified in JVET AhG 16~\cite{JVET-AJ2035,JVET-AG0042}. In particular, the key-reference frame is compressed via the VTM reference software 22.2 with the QPs of 2, 12, 22, 32, 42 and 52, and other subsequent frames are evolved into compact parameters and compressed via a context-adaptive arithmetic coder.}
\item{\textbf{Proposed PGen algorithm:} includes two layers, where the base-layer setting is the same as the GFVC anchor but the QP of key-reference frame is set at 22, and the enhancement layer is consisted of 3 different feature granularities with the feature size of 8$\times$8, 16$\times$16 and 32$\times$32. These different-granularities feature maps are compressed by the learning-based entropy model.}
\end{itemize}

\begin{figure*}[tb]
\centering
\subfloat[FOMM: Rate-DISTS]{\includegraphics[width=0.25\textwidth]{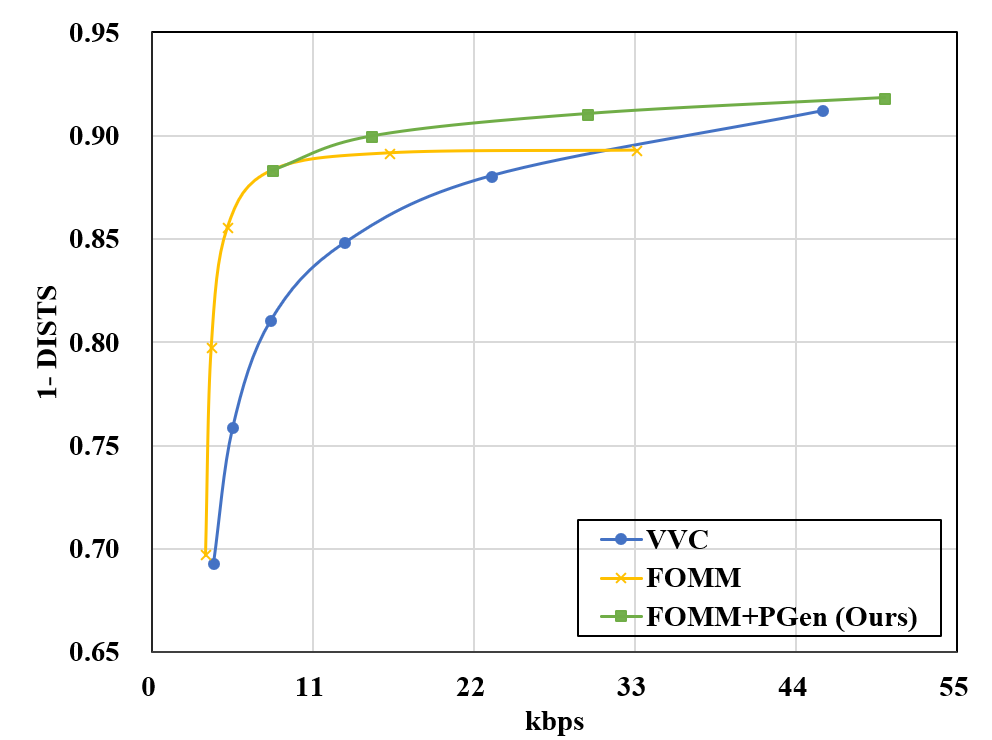}}
\subfloat[FOMM: Rate-LPIPS]{\includegraphics[width=0.25\textwidth,]{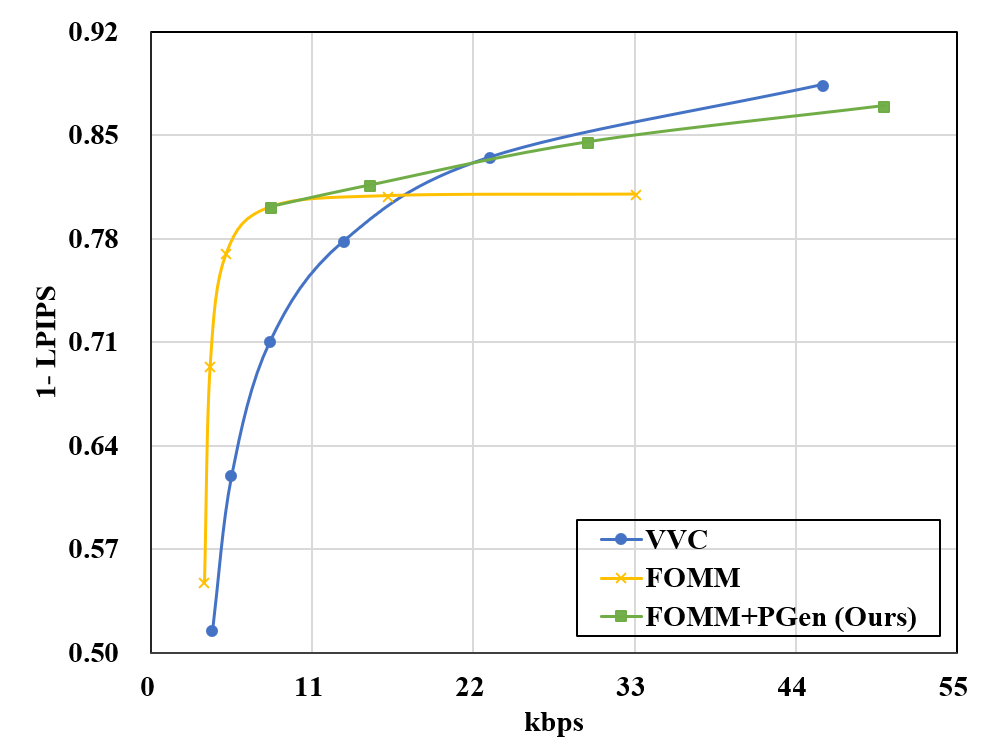}}
\subfloat[FOMM: Rate-FVD]
{\includegraphics[width=0.25\textwidth]{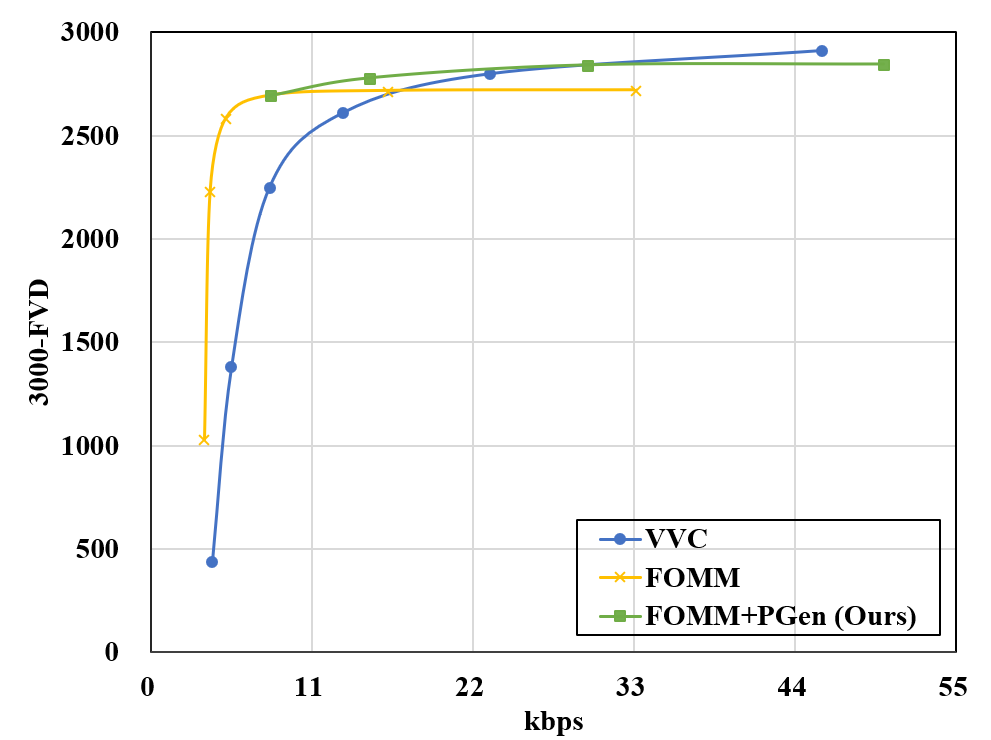}}
\subfloat[FOMM: Rate-MANIQA]{\includegraphics[width=0.25\textwidth]{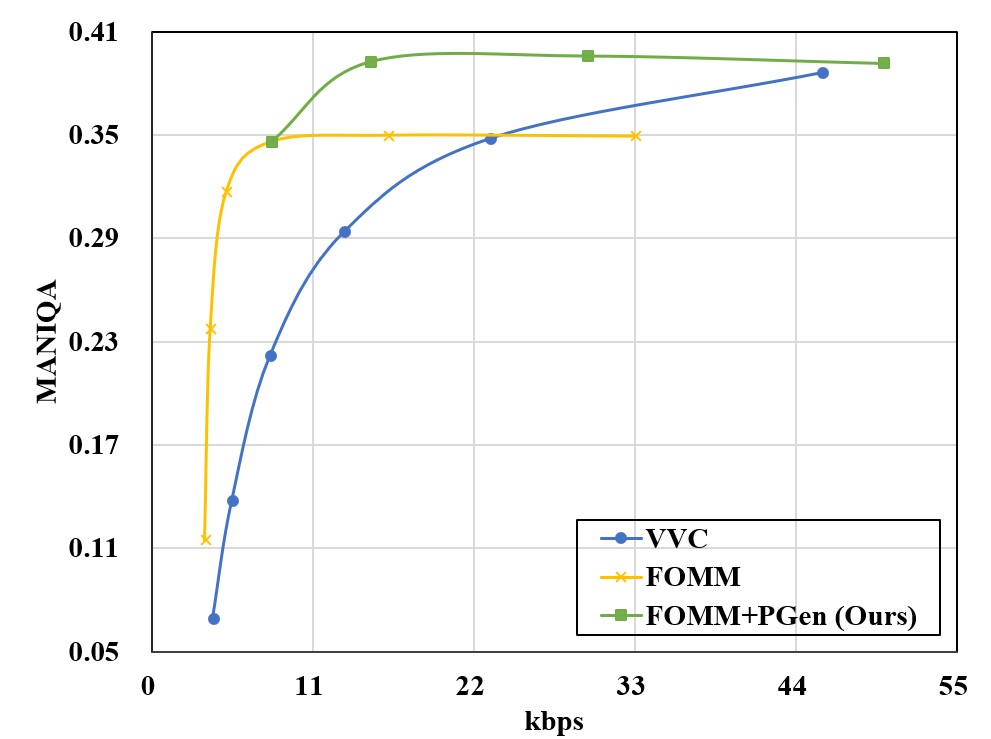}}\\
\subfloat[FV2V: Rate-DISTS]{\includegraphics[width=0.25\textwidth]{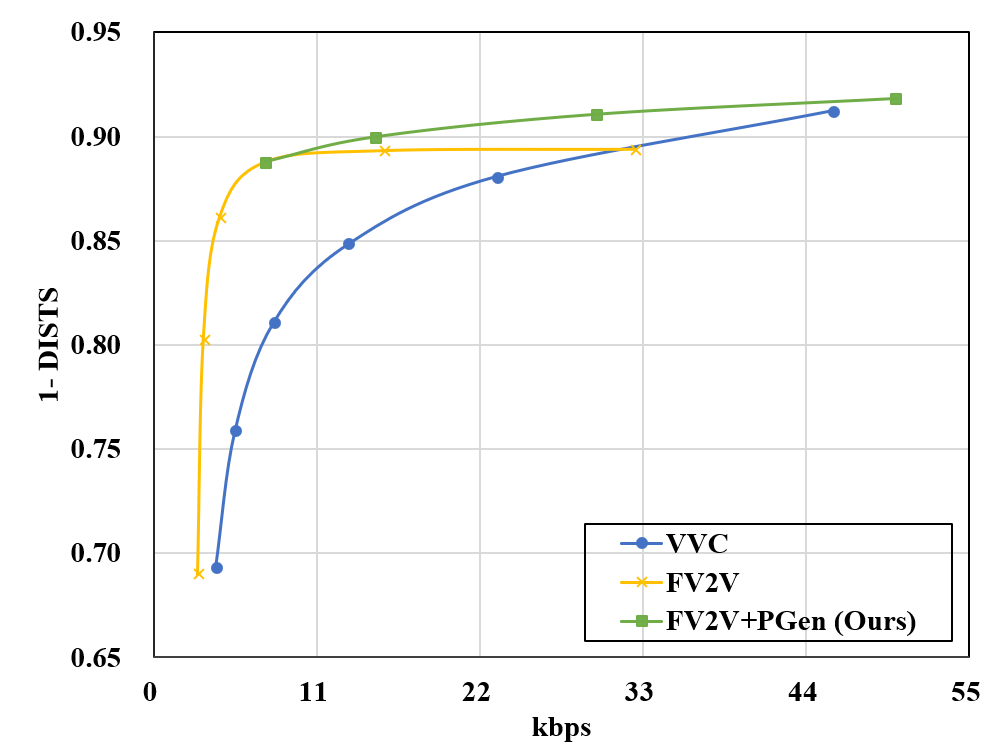}}
\subfloat[FV2V: Rate-LPIPS]{\includegraphics[width=0.25\textwidth,]{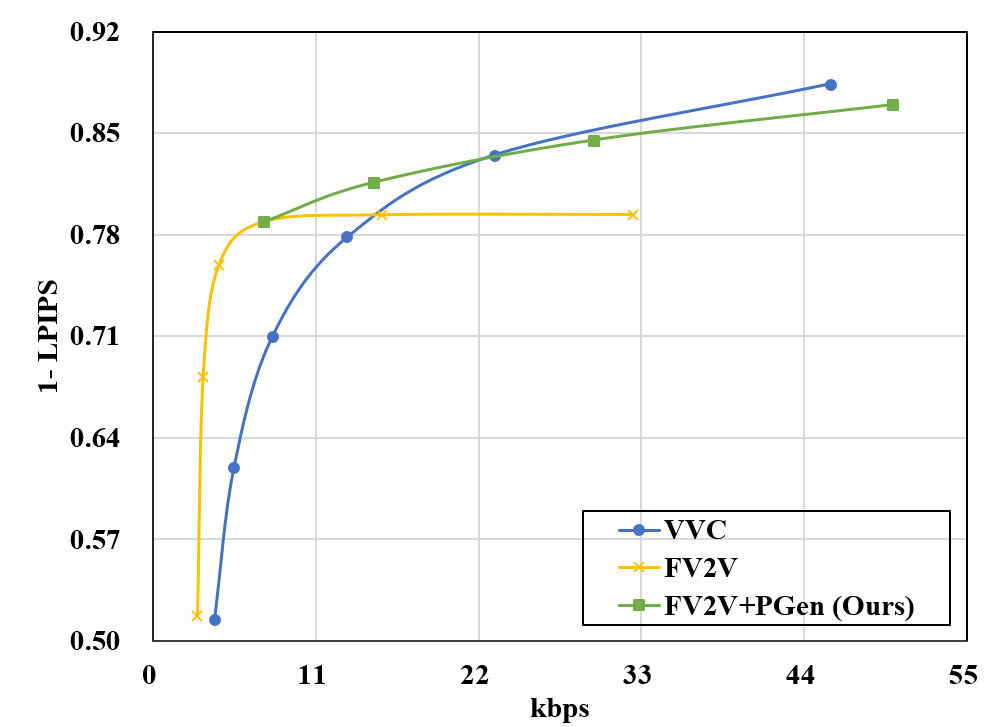}}
\subfloat[FV2V: Rate-FVD]
{\includegraphics[width=0.25\textwidth]{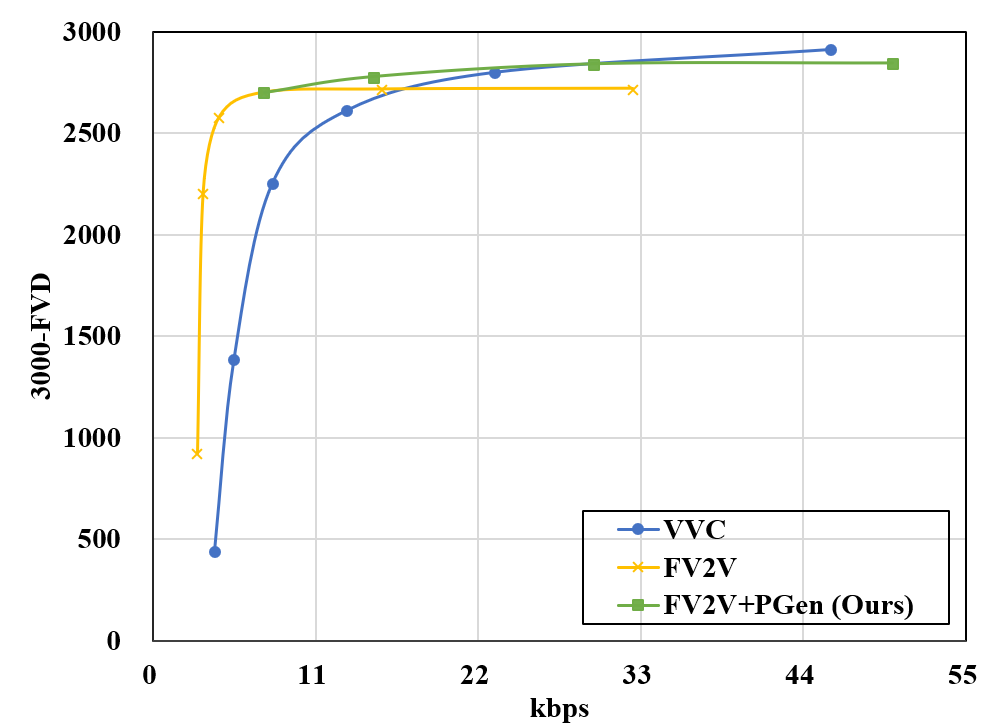}}
\subfloat[FV2V: Rate-MANIQA]{\includegraphics[width=0.25\textwidth]{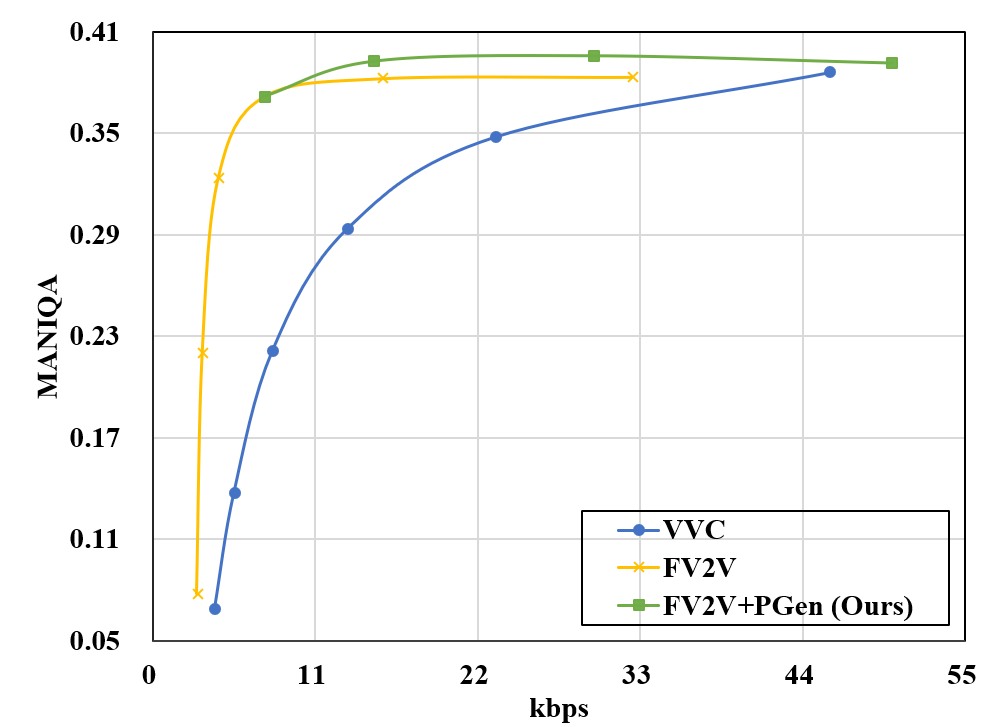}}\\
\subfloat[CFTE: Rate-DISTS]{\includegraphics[width=0.25\textwidth]{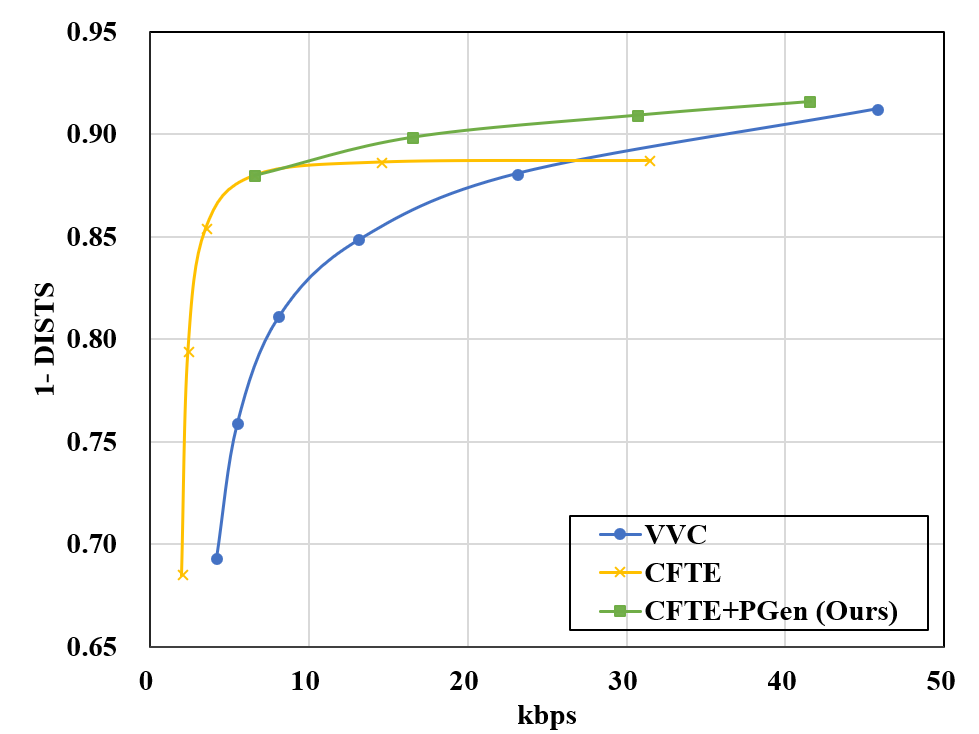}}
\subfloat[CFTE: Rate-LPIPS]{\includegraphics[width=0.25\textwidth,]{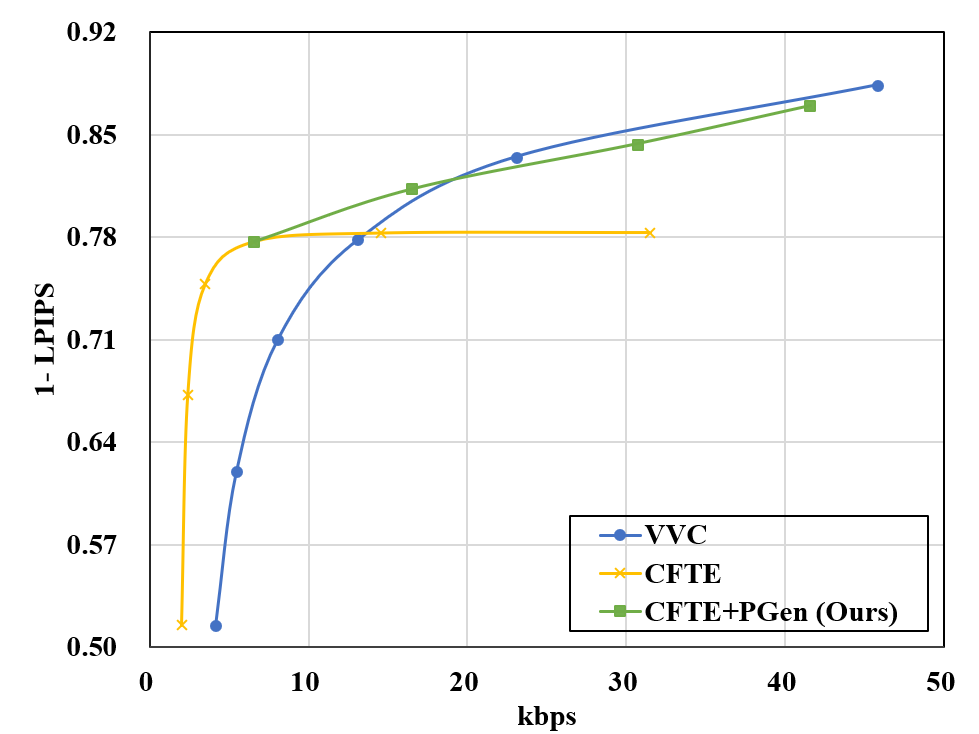}}
\subfloat[CFTE: Rate-FVD]
{\includegraphics[width=0.25\textwidth]{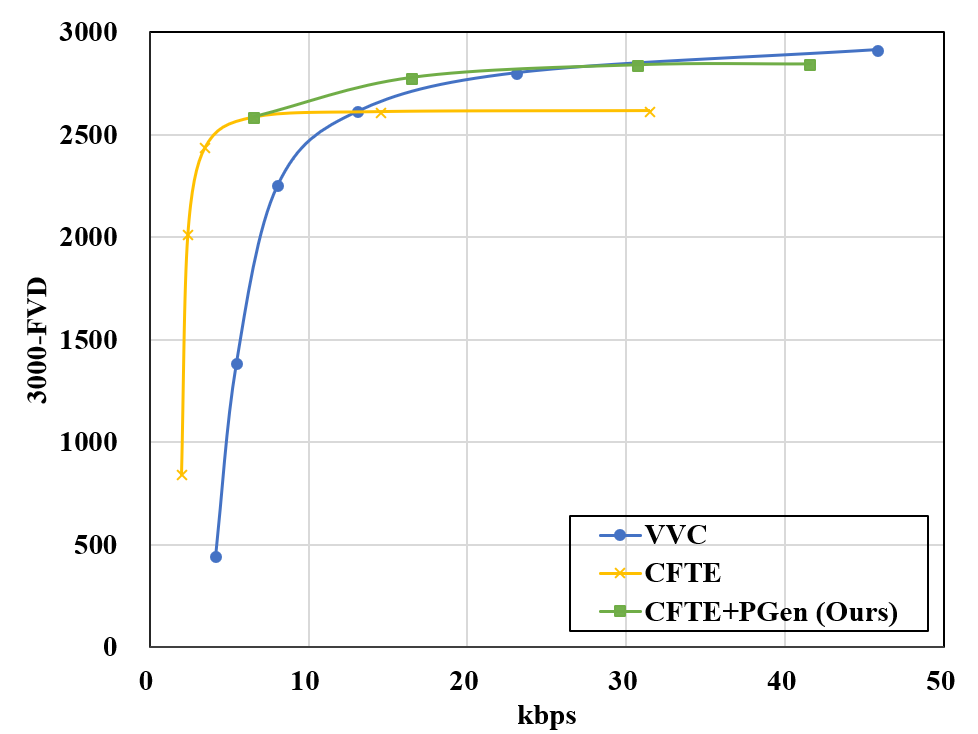}}
\subfloat[CFTE: Rate-MANIQA]{\includegraphics[width=0.25\textwidth]{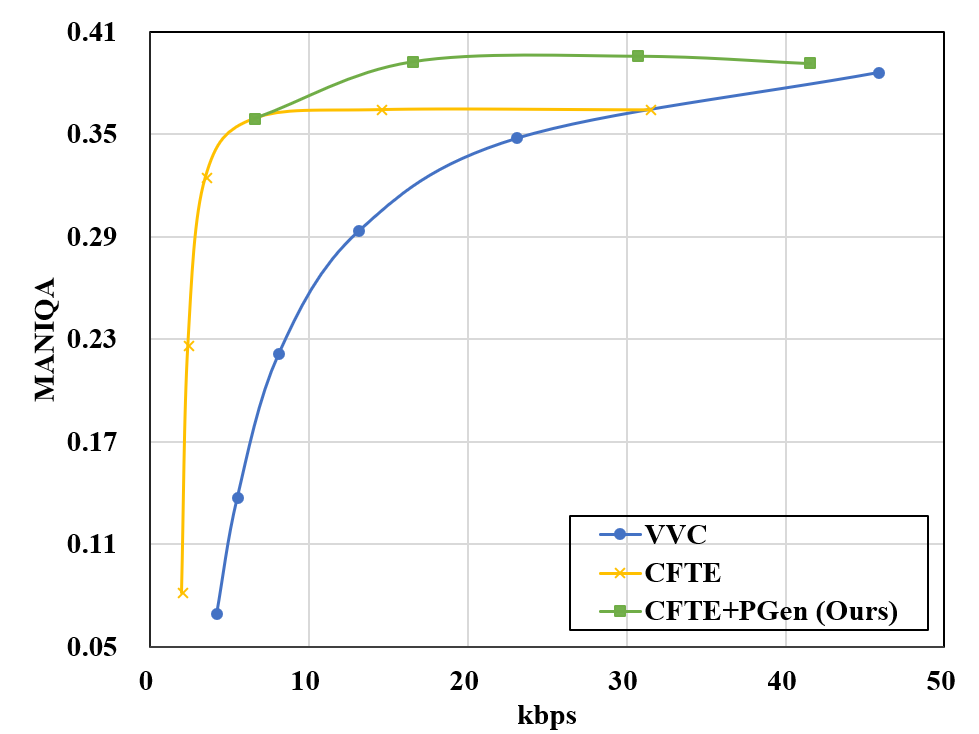}}
\caption{RD performance comparisons with VVC~\cite{bross2021overview}, FOMM~\cite{FOMM}, FV2V~\cite{wang2021Nvidia}, CFTE~\cite{CHEN2022DCC} and our proposed PGen scheme in terms of DISTS, LPIPS, FVD and MANIQA. } %
\label{fig8} 
\end{figure*}

\begin{table*}[t]
\caption{Average Bit-rate savings of 45 testing sequences from 3 different testing datasets in terms of DISTS, LPIPS, FVD and MANIQA}  
\renewcommand\arraystretch{1.30}
\resizebox{\linewidth}{!}{
\label{table_bd256}
\begin{tabular}{c|cccc|cccccccc}
\hline
\multirow{2}{*}{Dataset} & \multicolumn{4}{c|}{CFTE+PGen (Ours) v.s. VVC (VTM22.2)}                    & \multicolumn{4}{c}{FOMM+PGen (Ours) v.s. VVC (VTM22.2)}                                           & \multicolumn{4}{c}{FV2V+PGen (Ours) v.s. VVC (VTM22.2)}                     \\ \cline{2-13} 
                         & DISTS             & LPIPS           & FVD               & MANIQA            & DISTS             & LPIPS            & FVD               & \multicolumn{1}{c|}{MANIQA}            & DISTS             & LPIPS           & FVD               & MANIQA            \\ \hline
VoxCeleb                 & -40.29\%          & 0.43\%          & -42.11\%          & 33.78\%           & -38.12\%          & 16.60\%          & -43.22\%          & \multicolumn{1}{c|}{-70.03\%}          & -36.78\%          & 1.55\%          & -50.08\%          & -58.60\%          \\
300VW                    & -31.16\%          & 17.47\%         & -19.38\%          & -34.95\%          & -24.38\%          & 38.75\%          & -5.57\%           & \multicolumn{1}{c|}{-39.68\%}          & -26.87\%          & 21.73\%         & -14.80\%          & -41.97\%          \\
CFVQA                    & -34.84\%          & 1.11\%          & 22.39\%           & -54.99\%          & -34.07\%          & 0.48\%           & 14.63\%           & \multicolumn{1}{c|}{-46.12\%}          & -29.65\%          & -3.46\%         & 2.76\%            & -25.78\%          \\ \hline
\textbf{Average}         & \textbf{-35.43\%} & \textbf{6.33\%} & \textbf{-13.03\%} & \textbf{-18.72\%} & \textbf{-32.19\%} & \textbf{18.61\%} & \textbf{-11.39\%} & \multicolumn{1}{c|}{\textbf{-51.95\%}} & \textbf{-31.10\%} & \textbf{6.61\%} & \textbf{-20.71\%} & \textbf{-42.12\%} \\ \hline
\end{tabular}}
\end{table*}

\begin{table}[!t]
\renewcommand\arraystretch{1.3}
\caption{Coding speed of the VVC codec and proposed scalable compression scheme in terms of inference time (Sec) }
\label{table_time}
\centering
\begin{tabular}{cccc}
\hline
\multicolumn{2}{c}{Codec}                                                                                            & Encoder & Decoder \\ \hline
\multicolumn{2}{c}{VVC (VTM 22.2 LDB)}                                                                               & 780.13 & 0.45   \\
\multirow{3}{*}{\begin{tabular}[c]{@{}c@{}}Scalable\\ Compression \\ Scheme\end{tabular}} & Base layer (GFVC)        & 27.95  & 12.21   \\
                                                                                          & Enhancement layer (PGen) & 7.79  & 18.37  \\
                                                                                          & Total                    & 35.74 & 30.88  \\ \hline
\end{tabular}
\end{table}

\begin{table}[!t]
\renewcommand\arraystretch{1.3}
\caption{Model Complexity of the proposed PGen framework in terms of Params (M) and MAdd (G)}
\label{table_complexity}
\centering
\begin{tabular}{cccc}
\hline
Side                     & Component          & Params (M)                                           & MAdd (G)                                            \\ \hline
\multirow{3}{*}{Encoder} & Feature Descriptor & 14.13                                               & 3.96                                               \\
                         & Entropy Model    & 3.90$\times 10^{-4}$ & 8.45$\times 10^{-6}$ \\ 
                         & \textbf{Total}     & \textbf{14.13}                                      & \textbf{3.96}                                      \\ \hline
\multirow{5}{*}{Decoder} & Feature Descriptor & 14.13                                               & 3.96                                               \\
                         & Entropy Model   & 3.90$\times 10^{-4}$ & 8.45$\times 10^{-6}$ \\ 
                         & Frame Generator    & 56.66                                               & 174.41                                             \\
                         & \textbf{Total}     & \textbf{70.79}                                      & \textbf{178.37}                                    \\ \hline
\end{tabular}
\end{table}

\begin{figure*}[tb]
\centering
\subfloat[CFTE+PGen: VoxCeleb Seq 002]{\includegraphics[width=0.48\textwidth]{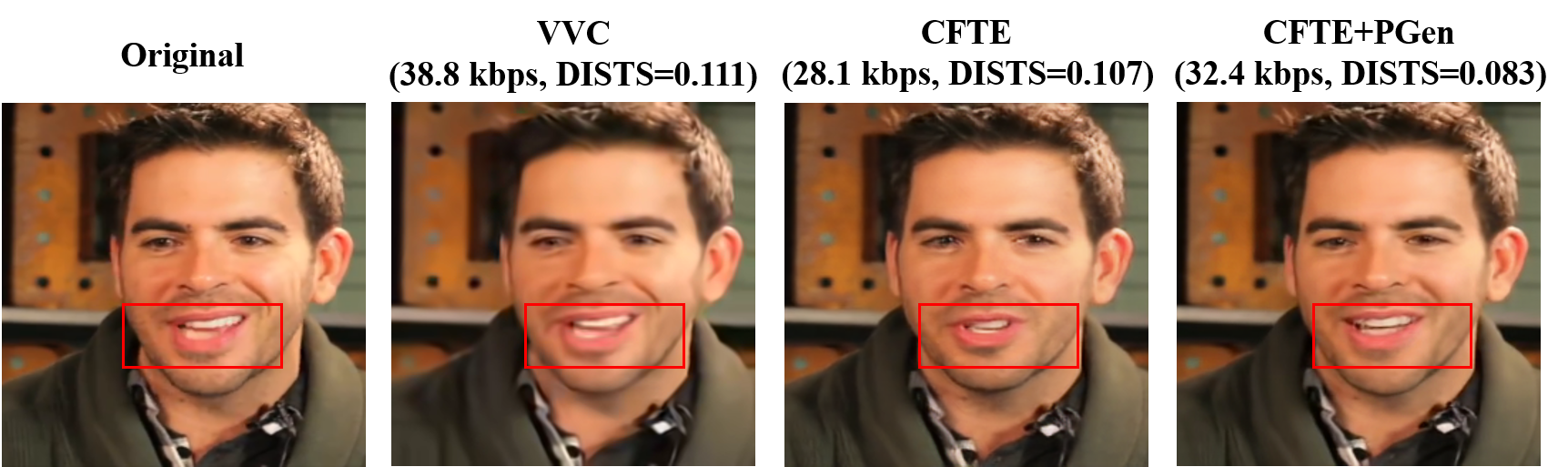}}
\hspace{1em}
\subfloat[CFTE+PGen: VoxCeleb Seq 011]{\includegraphics[width=0.48\textwidth]{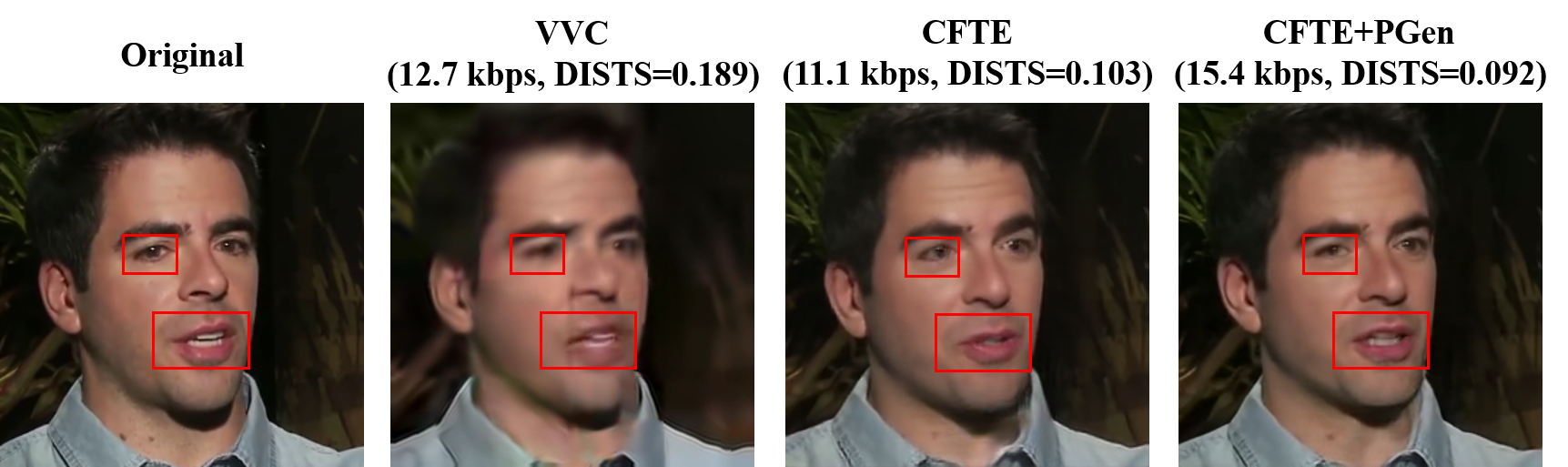}} \\
\subfloat[FOMM+PGen: 300-VW Seq 001]{\includegraphics[width=0.48\textwidth]{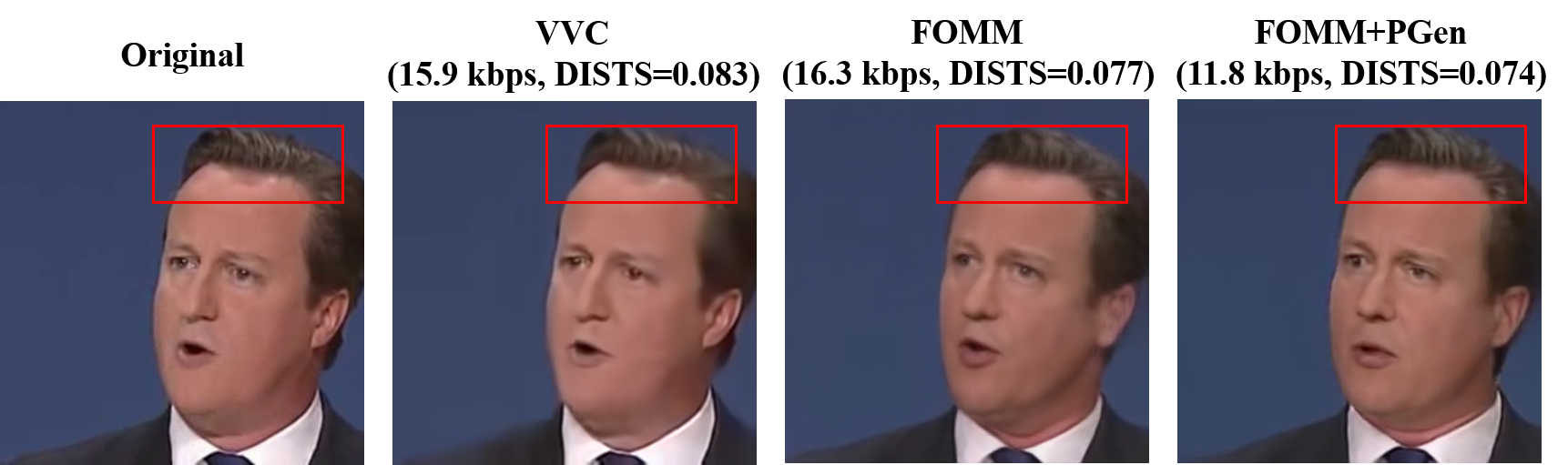}}
\hspace{1em}
\subfloat[FOMM+PGen: 300-VW Seq 006]{\includegraphics[width=0.48\textwidth]{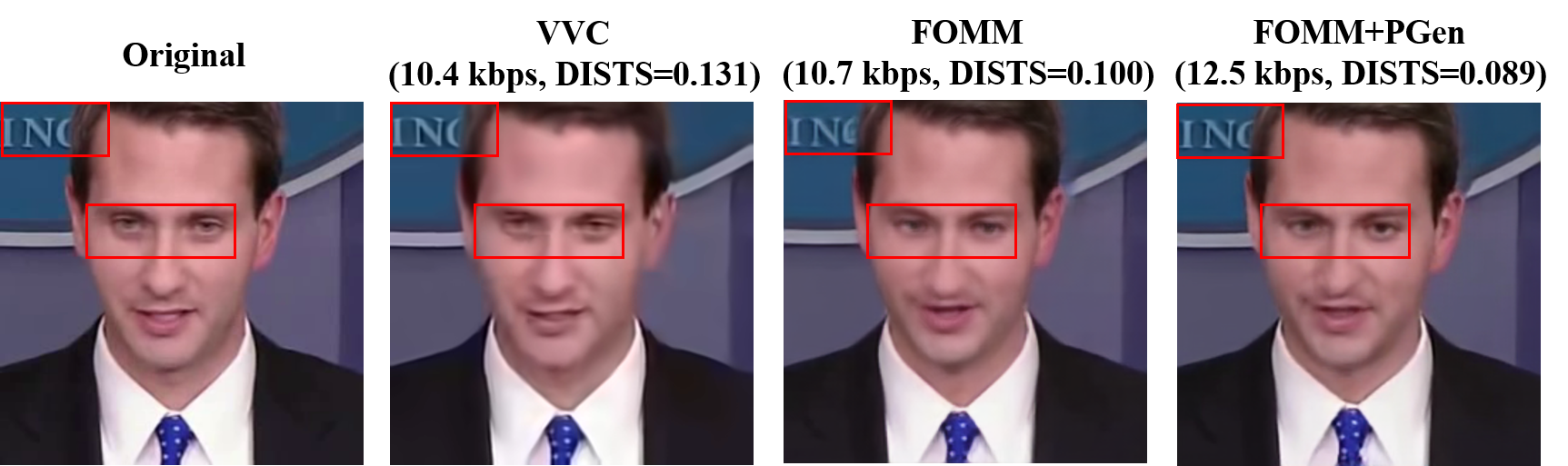}} \\
\subfloat[FV2V+PGen: CFVQA Seq 006]{\includegraphics[width=0.48\textwidth]{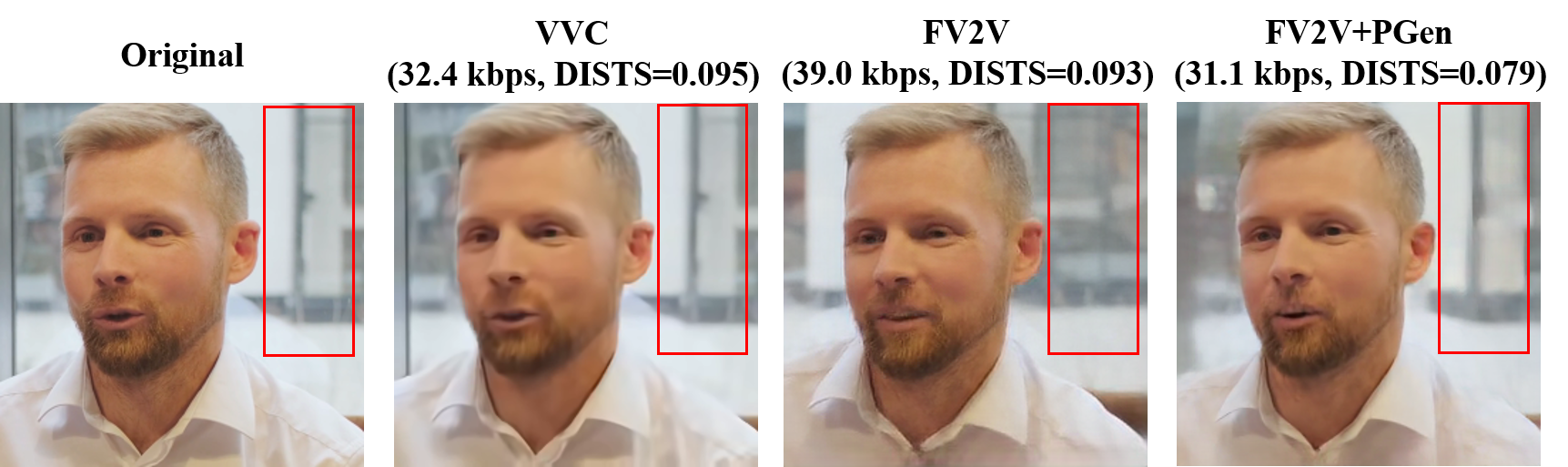}}
\hspace{1em}
\subfloat[FV2V+PGen: CFVQA Seq 010]{\includegraphics[width=0.48\textwidth]{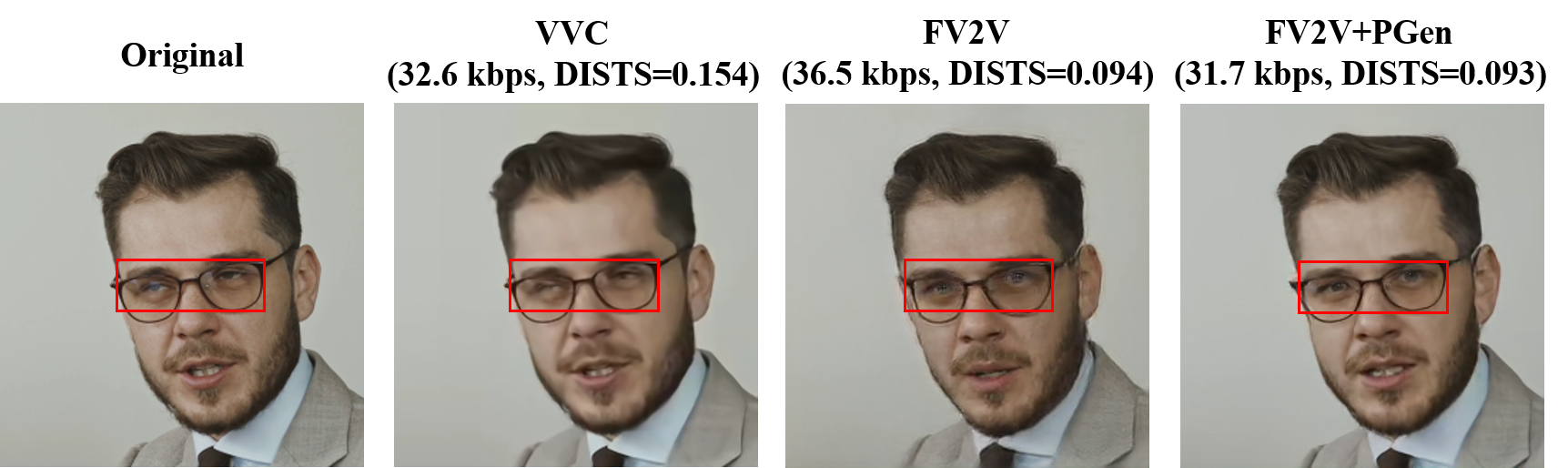}} 
\caption{Visual quality comparisons among VVC~\cite{bross2021overview}, 3 representative GFVC algorithms (CFTE~\cite{CHEN2022DCC}, FOMM~\cite{FOMM} and FV2V~\cite{wang2021Nvidia}) and their corresponding PGen-based scheme at similar bitrate.}
\label{fig:demo_subjective} 
\end{figure*}

\begin{figure}[tb]
\centering
\subfloat[CFTE+PGen Example]{\includegraphics[width=0.48\textwidth]{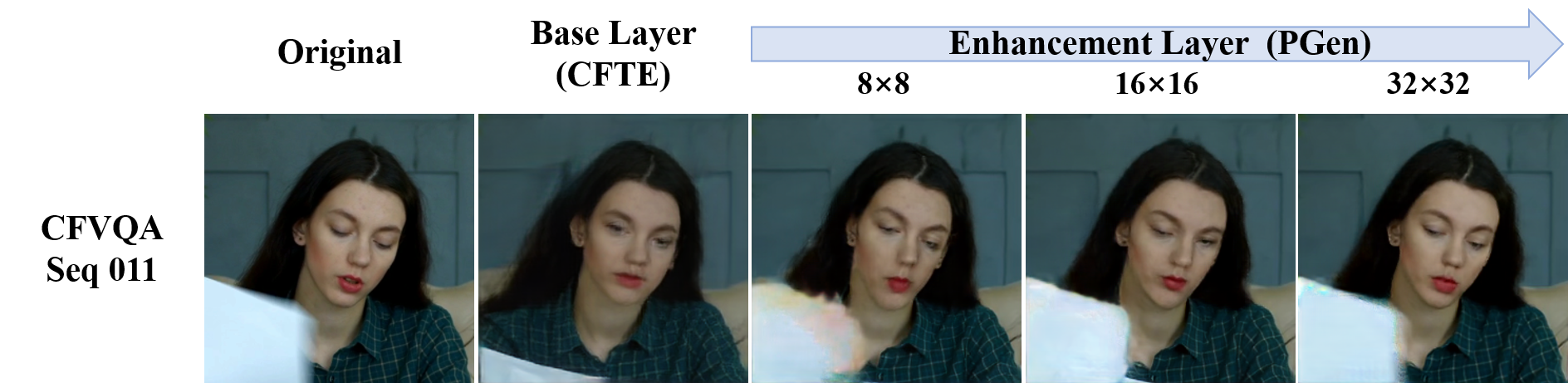}}\\
\subfloat[FOMM+PGen Example]{\includegraphics[width=0.48\textwidth,]{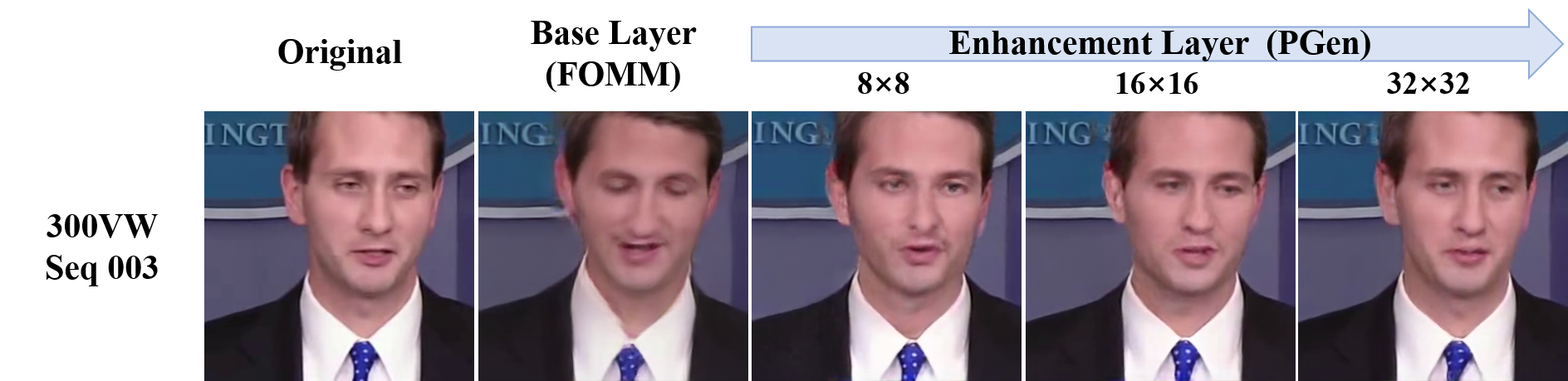}}\\
\subfloat[FV2V+PGen Example]{\includegraphics[width=0.48\textwidth]{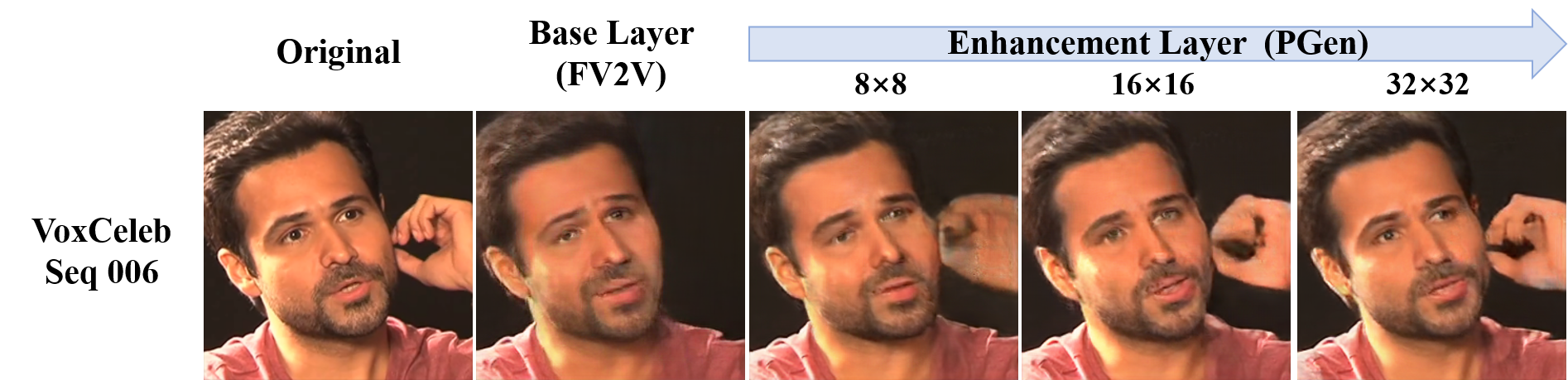}}
\caption{Subjective comparisons of reconstruction quality using GFVC anchors as base layer v.s. the proposed PGen with different granularities as enhancement layer. } 
\label{fig:demo_progressive} 
\end{figure}

\subsection{Performance Comparisons}

\subsubsection{RD Performance}
Fig. \ref{fig8} shows the RD performance of the VVC codec, three different GFVC methods and our proposed PGen-based GFVC scheme in terms of different quality measures (i.e., Rate-DISTS, Rate-LPIPS, Rate-FVD and Rate-MANIQA). It should be noted that Fig.  \ref{fig8} uses 1-DISTS, 1- LPIPS and 3000-FVD and MANIQA in order to present more coherent RD curves.
As illustrated in Fig. \ref{fig8}, the proposed PGen-based method can achieve better RD performance compared with VVC and also extend the bitrate range of GFVC algorithms from low bitrate range ($<$ 10 kbps) to relatively high bitrate ranges (10$\sim$50 kbps) in terms of Rate-DISTS and Rate-MANIQA. 
As for Rate-LPIPS, the proposed PGen-based algorithm also can achieve performance gains at middle bitrate ranges (10$\sim$20 kbps) but have a performance loss at higher bitrate ranges compared with VVC. 
As for the temporal-domain measurement Rate-FVD, the proposed PGen-based scheme can effectively expand the bitrate range of base-layer GFVC algorithms and achieve performance close to that of VVC, indicating that the proposed PGen can achieve good temporal consistency. Moreover, for base-layer GFVC algorithms with different facial features, the proposed PGen mechanism can show similar performance, illustrating that it has good plug-and-play property for all existing GFVC algorithms.

As shown in Table \ref{table_bd256}, in comparison with the latest VVC codec at high bitrate ranges, our proposed PGen scheme can achieve obvious bit-rate savings in terms of DISTS, FVD and MANIQA measures for different base-layer GFVC algorithms and different testing datasets. In particular, the proposed PGen-based algorithm can achieve 35.43\%, 32.19\% and 31.10\% average bit-rate savings in terms of Rate-DISTS for CFTE, FOMM and FV2V, respectively. In addition, there are 13.03\%, 11.39\% and 20.71\% average performance gains in terms of Rate-FVD for CFTE, FOMM and FV2V, respectively. As for Rate-MANIQA, the proposed PGen scheme can facilitate 18.72\%, 51.95\% and 42.12\% BDrate improvement for CFTE, FOMM and FV2V, respectively. As for Rate-LPIPS, there are no performance gains on average, which is consistent with the observations in Fig. \ref{fig8}.

\subsubsection{Subjective Comparisons}
As shown in Fig. \ref{fig:demo_subjective}, the proposed PGen scheme can facilitate all existing GFVC algorithms to better deliver faithful and high-fidelity face videos, and show advantageous reconstruction capabilities compared with the VVC codec at similar bitrate. In particular, it can be observed that face videos reconstructed from the VVC codec usually exhibit severe blocking/blurring artifacts at given bitrate (10$\sim$40 kbps), whilst our proposed scheme can reconstruct face videos with higher face fidelity, more accurate local facial motion and better background motion information.

Fig. \ref{fig:demo_progressive} provides visual examples to verify that the pro- posed PGen framework can reduce annoying visual artifacts and/or recover missing details caused by GFVC methods. For example, as shown in Fig. \ref{fig:demo_progressive} (a), the base-layer reconstruction results suffer from poor face fidelity and loss of details in textured background and the flipping page, while the enhancement-layer reconstruction results can progressively recover quality in these areas  with different-granularity features. Other visual examples are also provided, such as improving the face and expression fidelity in Fig. \ref{fig:demo_progressive} and restoring the missing hand in Fig. \ref{fig:demo_progressive} (c).

\begin{figure}[tb]
\centering
\subfloat[Rate-DISTS]{\includegraphics[width=0.24\textwidth]{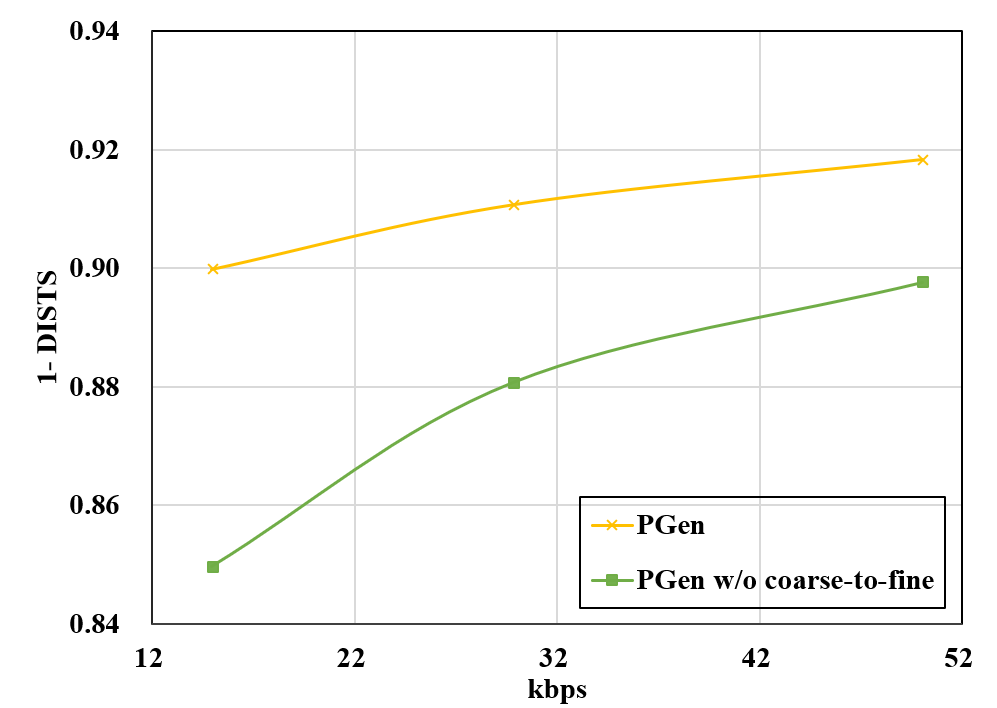}}
\subfloat[Rate-LPIPS]{\includegraphics[width=0.24\textwidth,]{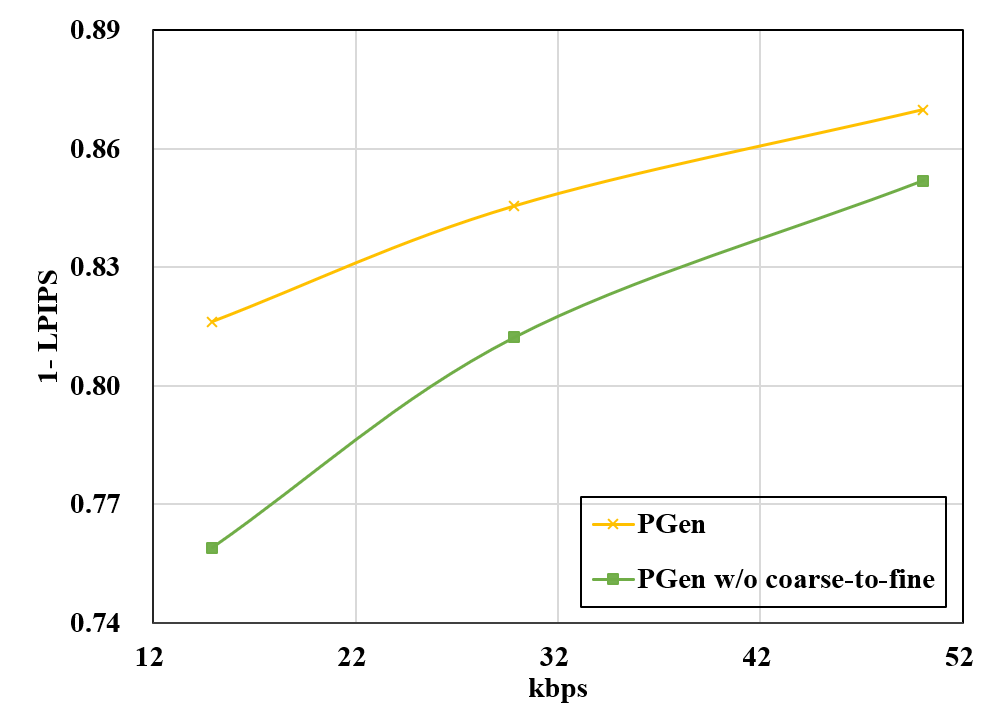}}\\
\subfloat[Rate-FVD]{\includegraphics[width=0.24\textwidth]{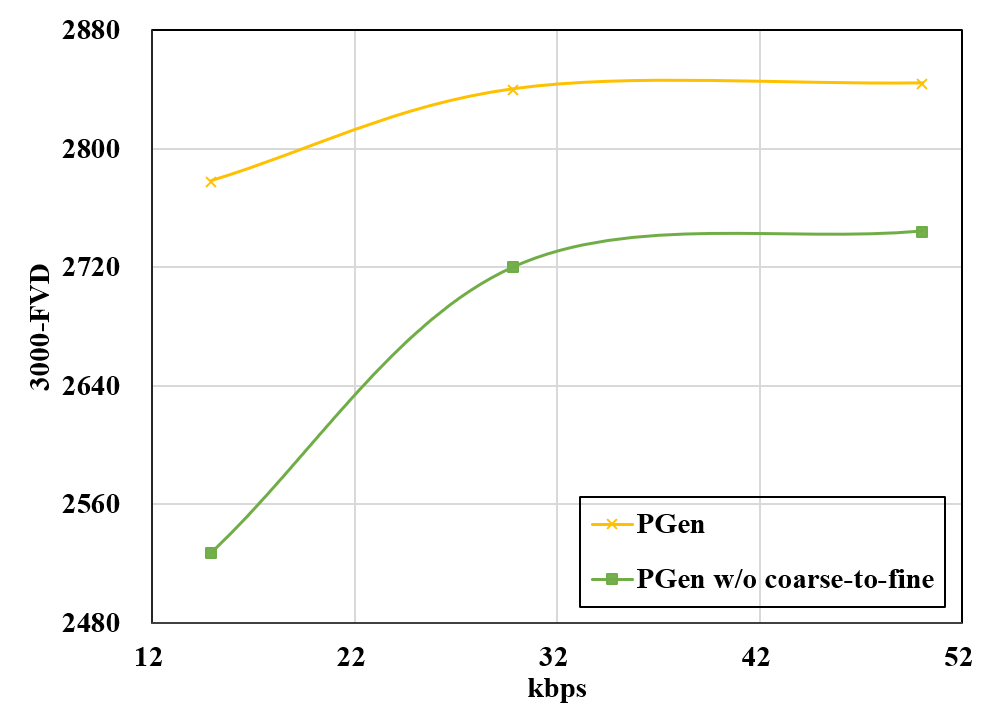}}
\subfloat[Rate-MANIQA]{\includegraphics[width=0.24\textwidth,]{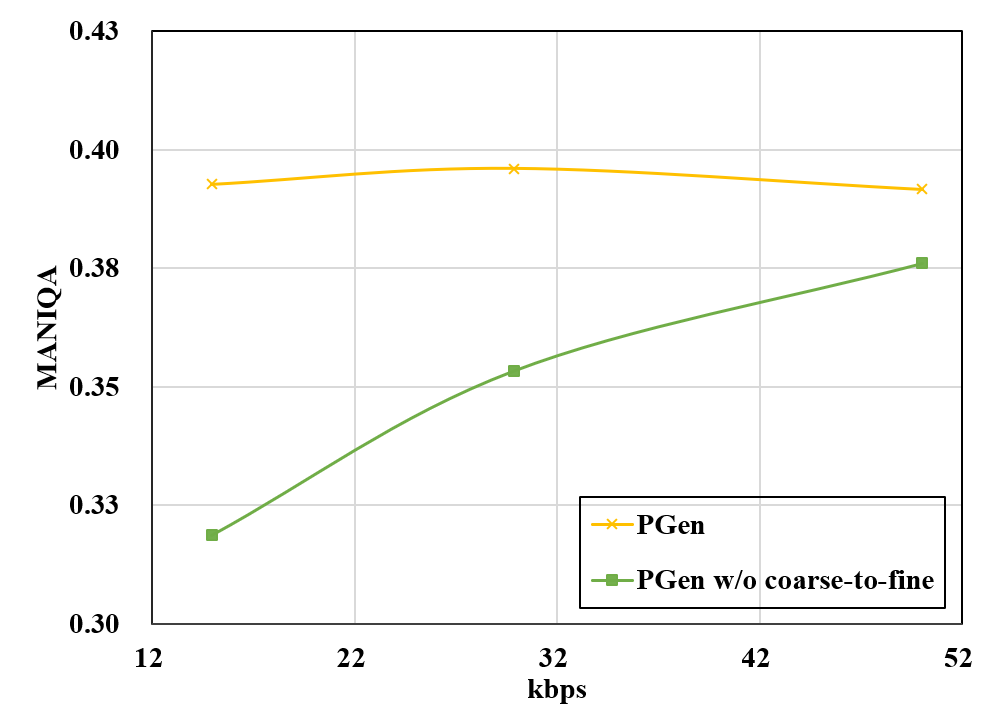}}\\
\subfloat[Rate-PSNR]{\includegraphics[width=0.24\textwidth]{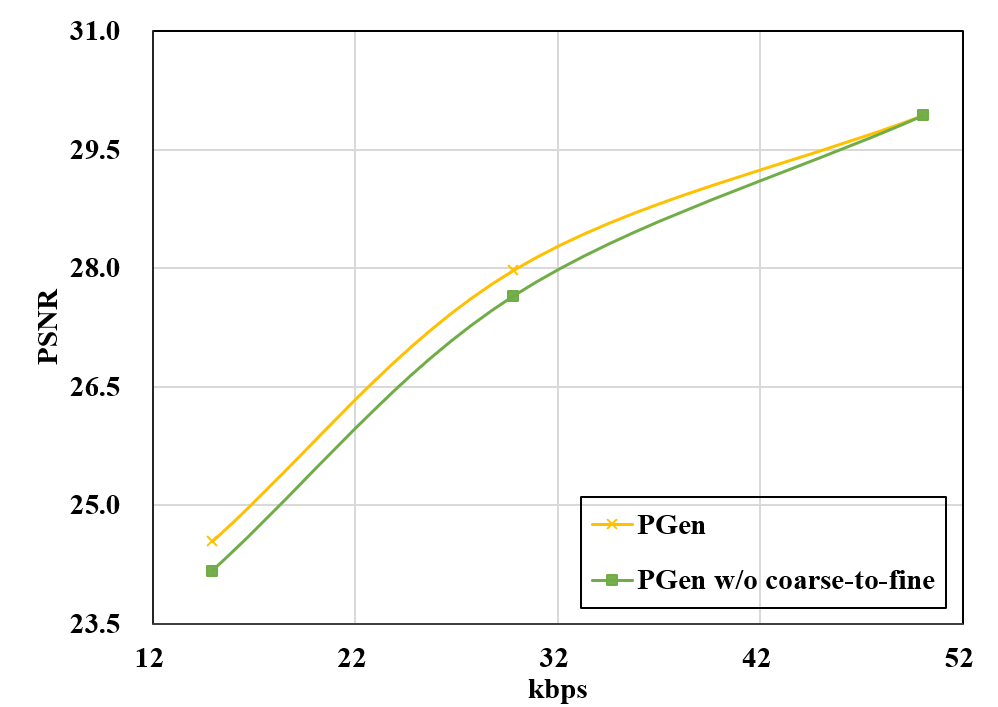}}
\subfloat[Rate-SSIM]{\includegraphics[width=0.24\textwidth,]{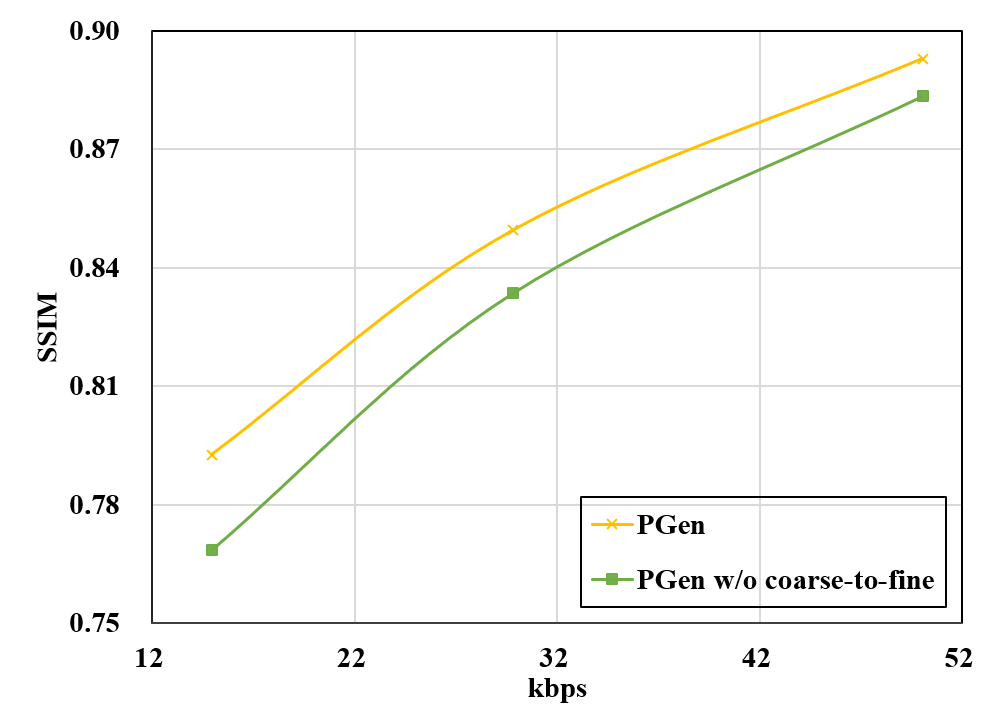}}
\caption{Ablation study of coarse-to-fine frame generation and attention-guided signal enhancement for the proposed PGen scheme. } %
\label{fig:ablation_study} 
\end{figure}

\begin{figure}[tb]
\centering
\subfloat[FOMM: Rate-PSNR]{\includegraphics[width=0.24\textwidth]{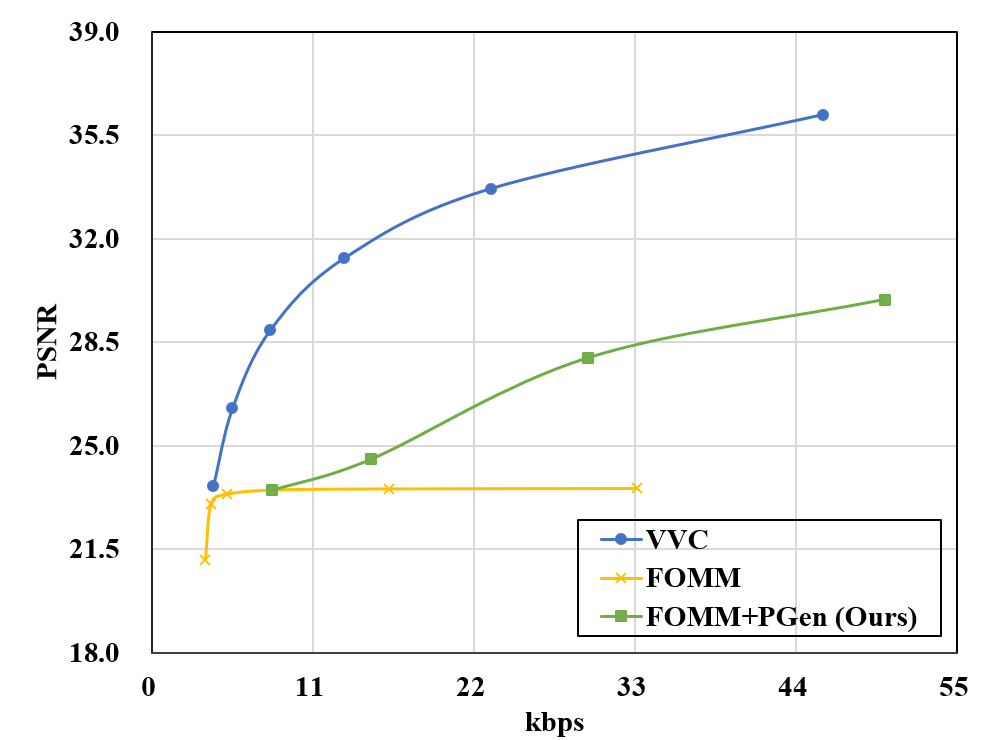}}
\subfloat[FOMM: Rate-SSIM]{\includegraphics[width=0.24\textwidth,]{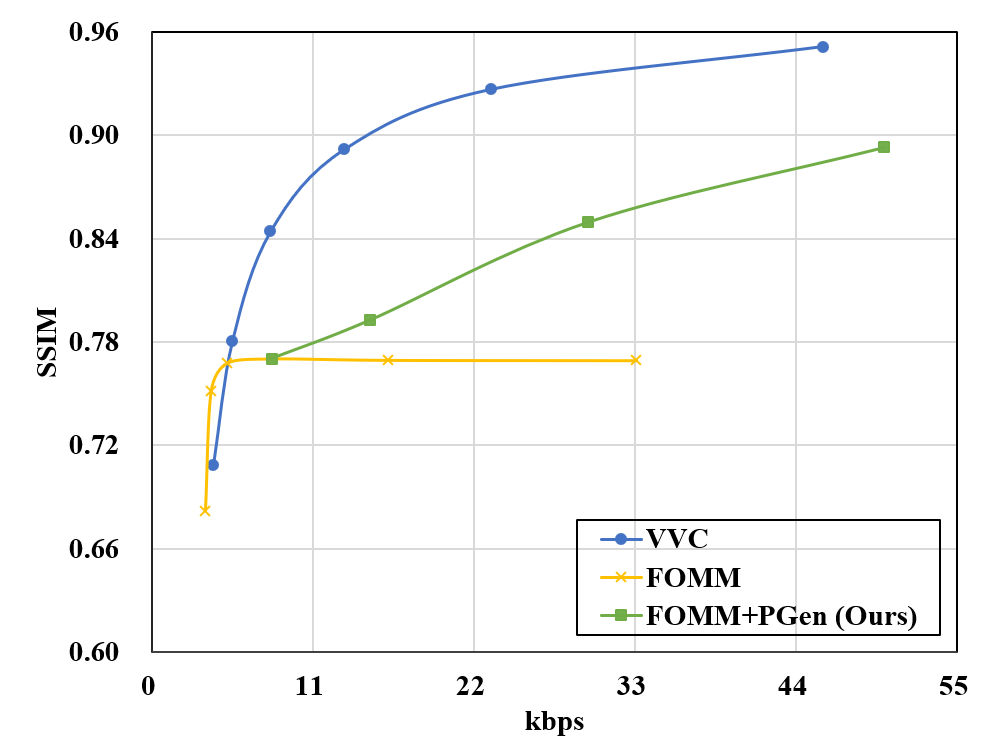}}\\
\subfloat[FV2V: Rate-PSNR]{\includegraphics[width=0.24\textwidth]{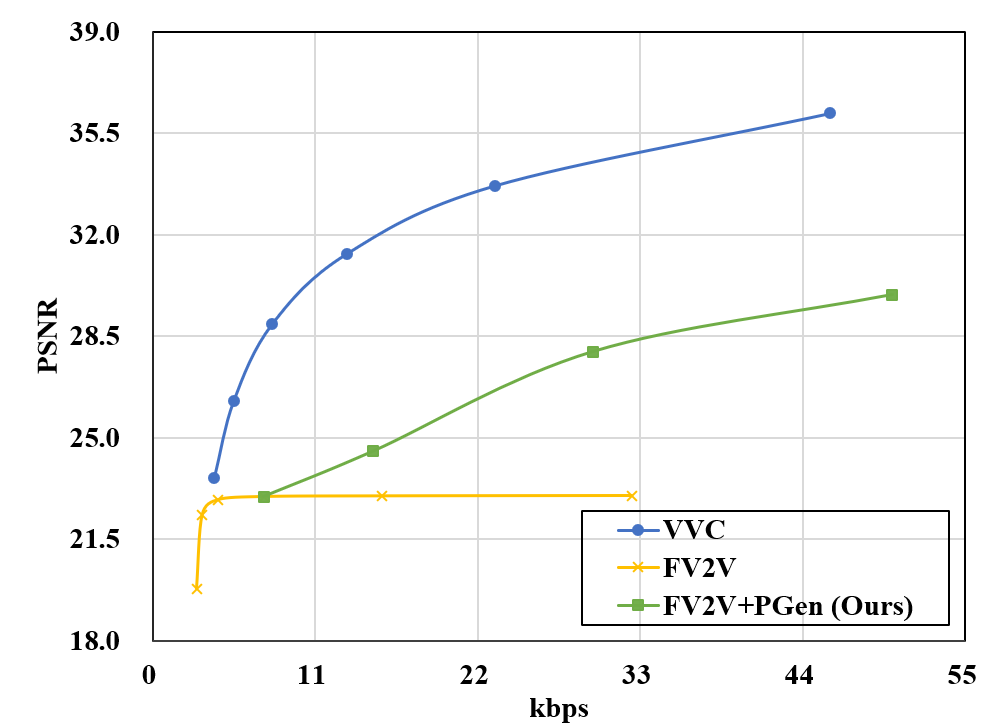}}
\subfloat[FV2V: Rate-SSIM]{\includegraphics[width=0.24\textwidth,]{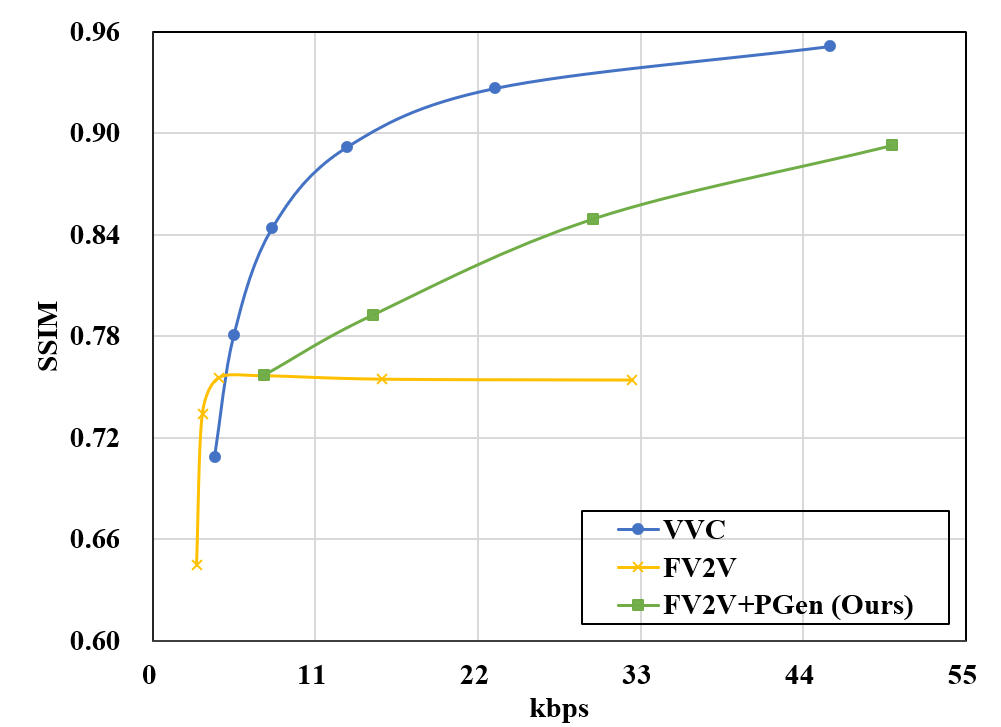}}\\
\subfloat[CFTE: Rate-PSNR]{\includegraphics[width=0.24\textwidth]{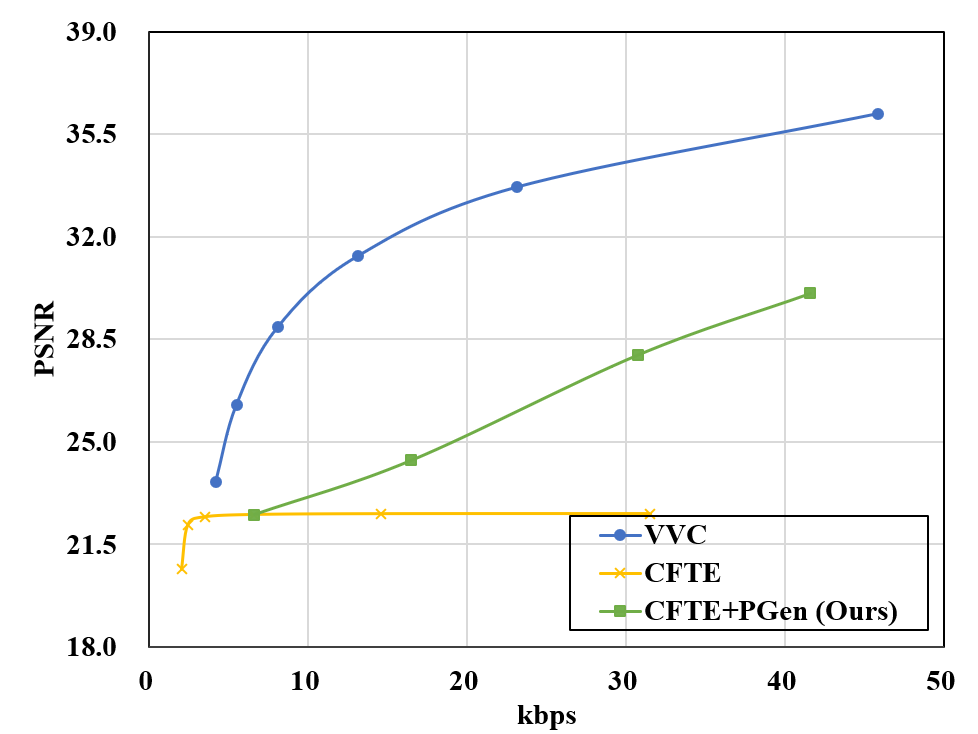}}
\subfloat[CFTE: Rate-SSIM]{\includegraphics[width=0.24\textwidth,]{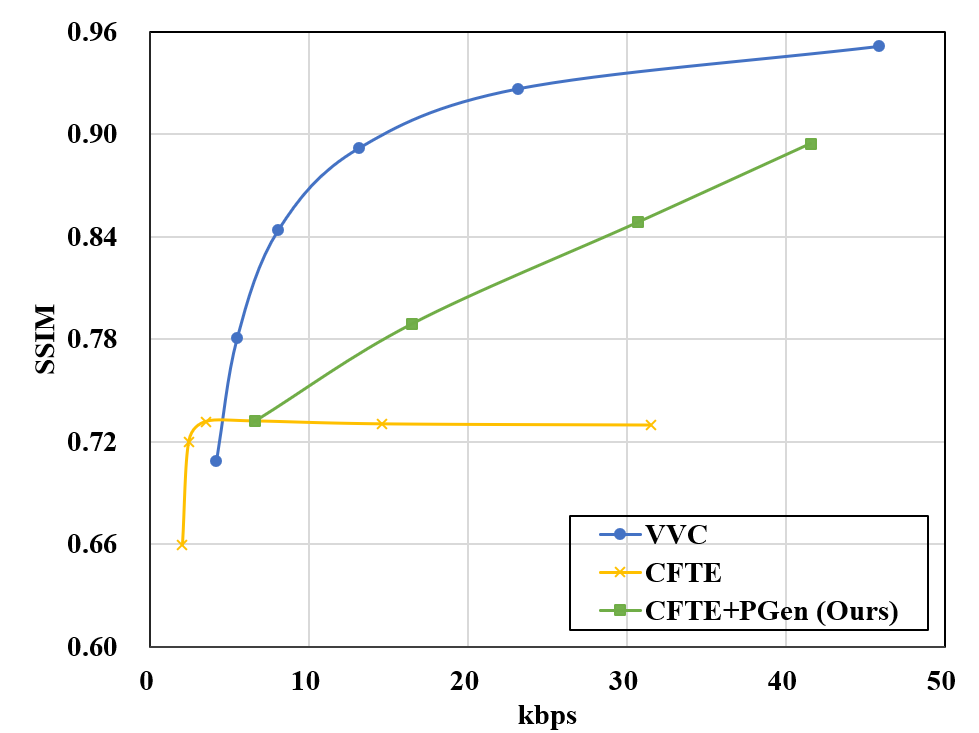}}
\caption{RD performance comparisons with VVC~\cite{bross2021overview}, FOMM~\cite{FOMM}, FV2V~\cite{wang2021Nvidia}, CFTE~\cite{CHEN2022DCC} and our proposed PGen scheme in terms of PSNR and SSIM. } %
\label{fig9} 
\end{figure}

\begin{figure}[tb]
\centering
\centerline{\includegraphics[width=0.5\textwidth]{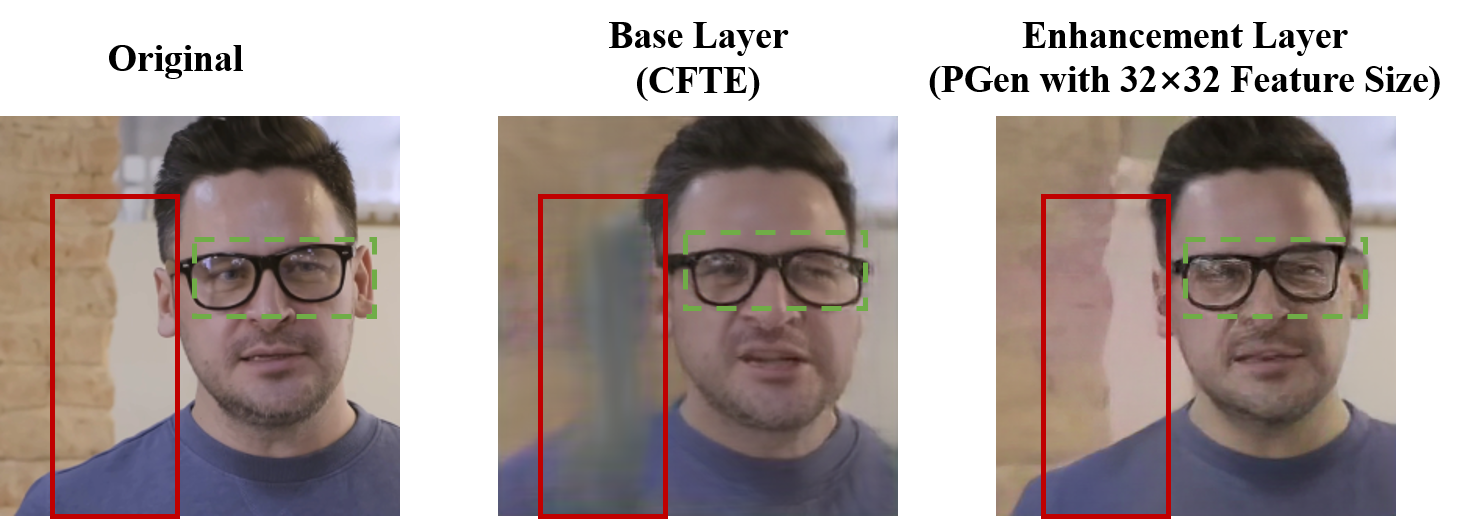}}
\caption{Poor-quality/unrealistic case of the proposed PGen scheme, where the solid red box reveals the unrealistic background synthesis and fading, and the green dashed box reveals the occluded distortions in the local facial details.}
\label{failure_case}
\vspace{-0.5em}
\end{figure} 

\subsubsection{Model Efficiency}

Table \ref{table_time} \& Table \ref{table_complexity} illustrate the coding speed of the proposed PGen algorithm in terms of inference time and model complexity in terms of number of model parameters (Params) and number of operations measured by multiply-adds (MAdd), respectively. The inference process is executed on Tesla-A100 with 15 core CPUs (Intel(R) Xeon(R) Platinum 8369B CPU @ 2.90GHz), and the
testing sequence has 250 frames at the resolution 256$\times$256. In addition, we average the actual inference time of VVC with 6 different testing QPs (QPs=27, 32, 37, 42, 47 and 52) and PGen with 3 different feature size (size= 8×8, 16×16 and 32×32), respectively. Herein, the VVC codec only use CPUs, while PGen relies on GPU for acceleration. Experimental results show that unlike the traditional VVC codec, the proposed PGen has more balanced inference time between the encoder and the decoder. This is in contrast to VVC whose encoder is significantly slower than its decoder. As for model complexity of the proposed PGen, the component of frame generator accounts for the vast majority of parameters and operation. 

\subsubsection{Ablation Study}
To verify the effectiveness of various components in the proposed PGen scheme, the ablation study in coarse-to-fine frame generation strategy is carried out. For coarse-to-fine frame generation, we remove the fine-generation part and only keep the coarse-generation part, \textit{i.e.,} PGen w/o coarse-to-fine. Fig. \ref{fig:ablation_study} illustrates the effectiveness of the coarse-to-fine mechanism in our proposed PGen framework in terms of different objective quality measures. In particular, when the coarse-to-fine generation strategy is abandoned, there is an obvious decline in objective quality for the reconstruction results. This is because the coarse generation in enhancement layer is always faced with double-accumulation errors from base layer reconstruction, resulting in misalignment of pixels and features. On the contrary, by using the coarse-to-fine generation results to predict optical flow and warp the original key-reference frame, the accumulation of generation errors can be effectively avoided.

\section{Discussion}
The PGen framework with scalable representation and layered reconstruction is proposed to extend the bitrate and quality range beyond what existing GFVC methods can cover, showing promising RD performance and good subjective reconstruction at relatively high bitrate ranges compared with the latest VVC codec. Although remarkable progress has been made, the current GFVC or even generative coding algorithms are still faced with some challenges, which we discuss as follows,

\section{Discussion}
The PGen framework with scalable representation and layered reconstruction is proposed to extend the bitrate and quality range beyond what existing GFVC methods can cover, showing promising RD performance and good subjective reconstruction at relatively high bitrate ranges compared with the latest VVC codec. Although remarkable progress has been made, the current GFVC or even generative coding algorithms are still faced with some challenges, which we discuss as follows,

\begin{itemize}
\item{\textbf{Poor pixel-level alignment}: Fig. \ref{fig9} shows the poor RD performance of the proposed PGen-based algorithms compared with VVC in terms of pixel-level measures PSNR and SSIM, although the proposed scheme can achieve obvious improvement compared with the base-layer GFVC algorithms. This further illustrates that even though the PGen scheme with different- granularity features can effectively compensate for the loss of information, it still cannot achieve pixel alignment since the algorithm itself is still a GAN- based reconstruction algorithm. How to make generative compression scheme achieve levels of pixel-level alignment  similar to traditional hybrid codecs is worth future study.} 

\item{\textbf{Model lightweight \& deployment}: As shown in Table \ref{table_time} \& Table \ref{table_complexity}, the model complexity of current GFVC algorithms is still very high. As a result, it could be difficult to implement terminal-side deployment and serve practical applications in the near future. In addition, although lightweight algorithms like distillation~\cite{gou2021knowledge}, quantization~\cite{polino2018model} and Pruning~\cite{han2015deep_compression} are now mature enough, the accuracy of the model and the final performance will inevitably decrease once lightweight has been implemented. As such, the GFVC standardization and deployment with hardware-software co-design should be further explored in the future, in pursuits to facilitate comprehensive GFVC investigations from JVET~\cite{m64987,JVET-AG0042}. } 

\item{\textbf{Model generalization \& robustness}: As illustrated in Fig. \ref{fig8} and Table \ref{table_bd256}, the proposed PGen framework can be well compatible with the existing GFVC algorithms for bitrate range extension and perceptual quality improvement. But due to the inherent  limitations of deep generative models, the PGen scheme sometimes can still reconstruct poor-quality results with obvious visual artifacts and unrealistic synthesis results as illustrated in Fig. \ref{failure_case}. Therefore, it is necessary to further optimize GFVC robustness to improve its generalization ability.}
\end{itemize}

\section{Conclusion}
In this paper, we present a universal enhanced coding framework with scalable representation and layered reconstruction, extending the GFVC codec’s versatility in terms of handling face video communication under a wide range of bitrate scenarios. Bitstreams of the proposed PGen framework can be layered with scalable representation from the prospective of multiple feature granularities and richness levels aimed at providing enriched signal to perceptually compensate for motion errors from overly compact feature representation. Experimental results demonstrate that our proposed PGen framework not only delivers reconstructed face video with more faithful motion representation and higher face fidelity, but also shows a superior RD performance in wide bitrate ranges compared with the latest VVC codec.

\bibliographystyle{IEEEtran}
\bibliography{main}

\begin{thebibliography}{100}
\providecommand{\url}[1]{#1}
\csname url@samestyle\endcsname
\providecommand{\newblock}{\relax}
\providecommand{\bibinfo}[2]{#2}
\providecommand{\BIBentrySTDinterwordspacing}{\spaceskip=0pt\relax}
\providecommand{\BIBentryALTinterwordstretchfactor}{4}
\providecommand{\BIBentryALTinterwordspacing}{\spaceskip=\fontdimen2\font plus
\BIBentryALTinterwordstretchfactor\fontdimen3\font minus \fontdimen4\font\relax}
\providecommand{\BIBforeignlanguage}[2]{{%
\expandafter\ifx\csname l@#1\endcsname\relax
\typeout{** WARNING: IEEEtran.bst: No hyphenation pattern has been}%
\typeout{** loaded for the language `#1'. Using the pattern for}%
\typeout{** the default language instead.}%
\else
\language=\csname l@#1\endcsname
\fi
#2}}
\providecommand{\BIBdecl}{\relax}
\BIBdecl

\bibitem{7268565}
W.~Schreiber, C.~Knapp, and N.~Kay, ``Synthetic highs—an experimental tv bandwidth reduction system,'' \emph{Journal of the SMPTE}, vol.~68, no.~8, pp. 525--537, 1959.

\bibitem{1457470}
D.~Pearson and J.~Robinson, ``Visual communication at very low data rates,'' \emph{Proceedings of the IEEE}, vol.~73, no.~4, pp. 795--812, 1985.

\bibitem{1989Object}
H.~G. Musmann, M.~Hötter, and J.~Ostermann, ``Object-oriented analysis-synthesis coding of moving images,'' \emph{Signal Processing: Image Communication}, vol.~1, no.~2, pp. 117--138, 1989.

\bibitem{lopez1995head}
R.~Lopez and T.~Huang, ``Head pose computation for very low bit-rate video coding,'' in \emph{International Conference on Computer Analysis of Images and Patterns}, 1995, pp. 440--447.

\bibitem{150969}
Y.~Nakaya, Y.~Chuah, and H.~Harashima, ``Model-based/waveform hybrid coding for videotelephone images,'' in \emph{International Conference on Acoustics, Speech, and Signal Processing}, 1991, pp. 2741--2744 vol.4.

\bibitem{364463}
K.~Aizawa and T.~Huang, ``Model-based image coding advanced video coding techniques for very low bit-rate applications,'' \emph{Proceedings of the IEEE}, vol.~83, no.~2, pp. 259--271, 1995.

\bibitem{9764682}
C.~Kong, B.~Chen, H.~Li, S.~Wang, A.~Rocha, and S.~Kwong, ``Detect and locate: Exposing face manipulation by semantic- and noise-level telltales,'' \emph{IEEE Transactions on Information Forensics and Security}, vol.~17, pp. 1741--1756, 2022.

\bibitem{9349088}
C.~Kong, B.~Chen, W.~Yang, H.~Li, P.~Chen, and S.~Wang, ``Appearance matters, so does audio: Revealing the hidden face via cross-modality transfer,'' \emph{IEEE Transactions on Circuits and Systems for Video Technology}, vol.~32, no.~1, pp. 423--436, 2022.

\bibitem{9868051}
C.~Kong, K.~Zheng, S.~Wang, A.~Rocha, and H.~Li, ``Beyond the pixel world: A novel acoustic-based face anti-spoofing system for smartphones,'' \emph{IEEE Transactions on Information Forensics and Security}, vol.~17, pp. 3238--3253, 2022.

\bibitem{10345691}
F.-T. Hong, L.~Shen, and D.~Xu, ``{DaGAN++}: Depth-aware generative adversarial network for talking head video generation,'' \emph{IEEE Transactions on Pattern Analysis and Machine Intelligence}, vol.~46, no.~5, pp. 2997--3012, 2024.

\bibitem{kong2025pixel}
C.~Kong, A.~Luo, S.~Wang, H.~Li, A.~Rocha, and A.~C. Kot, ``Pixel-inconsistency modeling for image manipulation localization,'' \emph{IEEE Transactions on Pattern Analysis and Machine Intelligence}, 2025.

\bibitem{wang2018recurrent}
W.~Wang, Y.~Yan, Z.~Cui, J.~Feng, S.~Yan, and N.~Sebe, ``Recurrent face aging with hierarchical autoregressive memory,'' \emph{IEEE transactions on pattern analysis and machine intelligence}, vol.~41, no.~3, pp. 654--668, 2018.

\bibitem{10381809}
T.~Liu, S.~Li, M.~Xu, L.~Yang, and X.~Wang, ``Assessing face image quality: A large-scale database and a transformer method,'' \emph{IEEE Transactions on Pattern Analysis and Machine Intelligence}, vol.~46, no.~5, pp. 3981--4000, 2024.

\bibitem{10547422}
A.~Tang, T.~He, X.~Tan, J.~Ling, R.~Li, S.~Zhao, J.~Bian, and L.~Song, ``Memories are one-to-many mapping alleviators in talking face generation,'' \emph{IEEE Transactions on Pattern Analysis and Machine Intelligence}, vol.~46, no.~12, pp. 8758--8770, 2024.

\bibitem{6920084}
H.~Han, C.~Otto, X.~Liu, and A.~K. Jain, ``Demographic estimation from face images: Human vs. machine performance,'' \emph{IEEE Transactions on Pattern Analysis and Machine Intelligence}, vol.~37, no.~6, pp. 1148--1161, 2015.

\bibitem{9099607}
S.~Zhang, C.~Chi, Z.~Lei, and S.~Z. Li, ``{RefineFace}: Refinement neural network for high performance face detection,'' \emph{IEEE Transactions on Pattern Analysis and Machine Intelligence}, vol.~43, no.~11, pp. 4008--4020, 2021.

\bibitem{VAE}
D.~P. Kingma and M.~Welling, ``Auto-encoding variational bayes,'' in \emph{International Conference on Learning Representations}, 2014, p.~14.

\bibitem{goodfellow2014generative}
I.~Goodfellow, J.~Pouget-Abadie, M.~Mirza, B.~Xu, D.~Warde-Farley, S.~Ozair, A.~Courville, and Y.~Bengio, ``Generative adversarial nets,'' \emph{Advances in Neural Information Processing Systems}, vol.~27, 2014.

\bibitem{NEURIPS2021_49ad23d1}
P.~Dhariwal and A.~Nichol, ``Diffusion models beat gans on image synthesis,'' in \emph{Advances in Neural Information Processing Systems}, vol.~34, 2021, pp. 8780--8794.

\bibitem{FOMM}
A.~Siarohin, S.~Lathuili{\`e}re, S.~Tulyakov, E.~Ricci, and N.~Sebe, ``First order motion model for image animation,'' \emph{Advances in Neural Information Processing Systems}, vol.~32, pp. 7137--7147, 2019.

\bibitem{8954170}
A.~Siarohin, S.~Lathuilière, S.~Tulyakov, E.~Ricci, and N.~Sebe, ``Animating arbitrary objects via deep motion transfer,'' in \emph{Proceedings of IEEE/CVF Conference on Computer Vision and Pattern Recognition}, 2019, pp. 2372--2381.

\bibitem{siarohin2021motion}
A.~Siarohin, O.~J. Woodford, J.~Ren, M.~Chai, and S.~Tulyakov, ``Motion representations for articulated animation,'' in \emph{Proceedings of the IEEE/CVF Conference on Computer Vision and Pattern Recognition}, 2021, pp. 13\,648--13\,657.

\bibitem{facebook2021}
M.~Oquab, P.~Stock, O.~Gafni, D.~Haziza, T.~Xu, P.~Zhang, O.~Celebi, Y.~Hasson, P.~Labatut, B.~Bose-Kolanu, T.~Peyronel, and C.~Couprie, ``Low bandwidth video-chat compression using deep generative models,'' in \emph{Proceedings of IEEE/CVF Conference on Computer Vision and Pattern Recognition Workshop}, 2021.

\bibitem{ultralow}
G.~Konuko, G.~Valenzise, and S.~Lathuili{\`{e}}re, ``Ultra-low bitrate video conferencing using deep image animation,'' in \emph{IEEE International Conference on Acoustics, Speech and Signal Processing}, 06 2021, pp. 4210--4214.

\bibitem{9859867}
A.~Tang, Y.~Huang, J.~Ling, Z.~Zhang, Y.~Zhang, R.~Xie, and L.~Song, ``Generative compression for face video: A hybrid scheme,'' in \emph{IEEE International Conference on Multimedia and Expo}, jul 2022, pp. 1--6.

\bibitem{CHEN2022DCC}
B.~Chen, Z.~Wang, B.~Li, R.~Lin, S.~Wang, and Y.~Ye, ``Beyond keypoint coding: Temporal evolution inference with compact feature representation for talking face video compression,'' in \emph{Data Compression Conference}, 2022, pp. 13--22.

\bibitem{icip2022zhao}
Z.~Wang, B.~Chen, Y.~Ye, and S.~Wang, ``Dynamic multi-reference generative prediction for face video compression,'' in \emph{IEEE International Conference on Image Processing}, 2022, pp. 896--900.

\bibitem{chen2023generative}
B.~Chen, J.~Chen, S.~Wang, and Y.~Ye, ``Generative face video coding techniques and standardization efforts: A review,'' in \emph{Data Compression Conference}, 2024, pp. 103--112.

\bibitem{10811831}
Z.~Wang, B.~Chen, S.~Wang, S.~Wang, Y.~Ye, and S.~Ma, ``Ultra-low bitrate face video compression based on conversions from 3d keypoints to 2d motion map,'' \emph{IEEE Transactions on Image Processing}, vol.~33, pp. 6850--6864, 2024.

\bibitem{10743340}
Z.~Zhang, B.~Chen, S.~Yin, S.~Wang, and Y.~Ye, ``Light-weighted temporal evolution inference for generative face video compression,'' in \emph{2024 IEEE 26th International Workshop on Multimedia Signal Processing (MMSP)}, 2024, pp. 1--6.

\bibitem{henaff2019perceptual}
O.~J. H{\'e}naff, R.~L. Goris, and E.~P. Simoncelli, ``Perceptual straightening of natural videos,'' \emph{Nature neuroscience}, vol.~22, no.~6, pp. 984--991, 2019.

\bibitem{chen2024standardizing}
B.~Chen, Y.~Ye, J.~Chen, R.-L. Liao, S.~Yin, S.~Wang, K.~Yang, Y.~Li, Y.~Xu, Y.-K. Wang \emph{et~al.}, ``Standardizing generative face video compression using supplemental enhancement information,'' \emph{arXiv preprint arXiv:2410.15105}, 2024.

\bibitem{JVET-AJ2035}
S.~McCarthy and B.~Chen, ``Test conditions and evaluation procedures for generative face video coding,'' \emph{{The Joint Video Experts Team of ITU-T SG 16 WP 3 and ISO/IEC JTC 1/SC 29, doc. no. JVET-AJ035}}, November 2024.

\bibitem{JVET-AG0042}
B.~Chen, J.~Chen, R.-L. Liao, Y.~Ye, and S.~Wang, ``{AHG16}: Proposed common software tools and testing conditions for generative face video compression,'' \emph{{The Joint Video Experts Team of ITU-T SG 16 WP 3 and ISO/IEC JTC 1/SC 29, doc. no. JVET-AG0042}}, January 2024.

\bibitem{JVET-AH0114}
B.~Chen, Y.~Ye, G.Konuko, G.~Valenzise, S.~Yin, and S.~Wang, ``{AHG} 16: Updated common software tools for generative face video compression,'' \emph{{The JVET of ITU-T SG 16 WP 3 and ISO/IEC JTC 1/SC 29, doc. no. JVET-AH0114}}, April 2024.

\bibitem{JVET-AJ2006}
J.~Boyce, J.~Chen, S.~Deshpande \emph{et~al.}, ``Additional {SEI} messages for {VSEI} version 4 ({Draft 4}),'' \emph{{The JVET of ITU-T SG 16 WP 3 and ISO/IEC JTC 1/SC 29, doc. no. JVET-AJ2006}}, November 2024.

\bibitem{jia2019layered}
C.~Jia, Z.~Liu, Y.~Wang, S.~Ma, and W.~Gao, ``Layered image compression using scalable auto-encoder,'' in \emph{2019 IEEE Conference on Multimedia Information Processing and Retrieval (MIPR)}.\hskip 1em plus 0.5em minus 0.4em\relax IEEE, 2019, pp. 431--436.

\bibitem{9547677}
Y.~Mei, L.~Li, Z.~Li, and F.~Li, ``Learning-based scalable image compression with latent-feature reuse and prediction,'' \emph{IEEE Transactions on Multimedia}, vol.~24, pp. 4143--4157, 2022.

\bibitem{man1982computational}
D.~Man and A.~Vision, ``A computational investigation into the human representation and processing of visual information,'' \emph{WH San Francisco: Freeman and Company, San Francisco}, vol.~1, 1982.

\bibitem{schwarz2007overview}
H.~Schwarz, D.~Marpe, and T.~Wiegand, ``Overview of the scalable video coding extension of the h. 264/avc standard,'' \emph{IEEE Transactions on Circuits and Systems for Video Technology}, vol.~17, no.~9, pp. 1103--1120, 2007.

\bibitem{wiegand2003overview}
T.~Wiegand, G.~J. Sullivan, G.~Bjontegaard, and A.~Luthra, ``Overview of the {H.264/AVC} video coding standard,'' \emph{IEEE Transactions on Circuits and Systems for Video Technology}, vol.~13, no.~7, pp. 560--576, 2003.

\bibitem{sullivan2012overview}
G.~Sullivan, J.~Ohm, W.~Han, and T.~Wiegand, ``Overview of the {H}igh {E}fficiency {V}ideo {C}oding ({HEVC}) standard,'' \emph{IEEE Transactions on Circuits and Systems for Video Technology}, vol.~22, no.~12, pp. 1649--1668, 2012.

\bibitem{bross2021overview}
B.~Bross, Y.~Wang, Y.~Ye, S.~Liu, J.~Chen, G.~Sullivan, and J.~Ohm, ``Overview of the {Versatile Video Coding (VVC)} standard and its applications,'' \emph{IEEE Transactions on Circuits and Systems for Video Technology}, vol.~31, no.~10, pp. 3736--3764, 2021.

\bibitem{ohm2005advances}
J.-R. Ohm, ``Advances in scalable video coding,'' \emph{Proceedings of the IEEE}, vol.~93, no.~1, pp. 42--56, 2005.

\bibitem{amon2007file}
P.~Amon, T.~Rathgen, and D.~Singer, ``File format for scalable video coding,'' \emph{IEEE Transactions on Circuits and Systems for Video Technology}, vol.~17, no.~9, pp. 1174--1185, 2007.

\bibitem{boyce2015overview}
J.~M. Boyce, Y.~Ye, J.~Chen, and A.~K. Ramasubramonian, ``Overview of shvc: Scalable extensions of the high efficiency video coding standard,'' \emph{IEEE Transactions on Circuits and Systems for Video Technology}, vol.~26, no.~1, pp. 20--34, 2015.

\bibitem{ye2014scalable}
Y.~Ye and P.~Andrivon, ``The scalable extensions of hevc for ultra-high-definition video delivery,'' \emph{IEEE MultiMedia}, vol.~21, no.~3, pp. 58--64, 2014.

\bibitem{zhu2019generative}
L.~Zhu, S.~Kwong, Y.~Zhang, S.~Wang, and X.~Wang, ``Generative adversarial network-based intra prediction for video coding,'' \emph{IEEE Transactions on Multimedia}, vol.~22, no.~1, pp. 45--58, 2019.

\bibitem{zhao2018enhanced}
Z.~Zhao, S.~Wang, S.~Wang, X.~Zhang, S.~Ma, and J.~Yang, ``Enhanced bi-prediction with convolutional neural network for high-efficiency video coding,'' \emph{IEEE Transactions on Circuits and Systems for Video Technology}, vol.~29, no.~11, pp. 3291--3301, 2018.

\bibitem{ma2019convolutional}
C.~Ma, D.~Liu, X.~Peng, L.~Li, and F.~Wu, ``Convolutional neural network-based arithmetic coding for hevc intra-predicted residues,'' \emph{IEEE Transactions on Circuits and Systems for Video Technology}, vol.~30, no.~7, pp. 1901--1916, 2019.

\bibitem{liu2019one}
J.~Liu, S.~Xia, W.~Yang, M.~Li, and D.~Liu, ``One-for-all: Grouped variation network-based fractional interpolation in video coding,'' \emph{IEEE Transactions on Image Processing}, vol.~28, no.~5, pp. 2140--2151, 2019.

\bibitem{jia2019content}
C.~Jia, S.~Wang, X.~Zhang, S.~Wang, J.~Liu, S.~Pu, and S.~Ma, ``Content-aware convolutional neural network for in-loop filtering in high efficiency video coding,'' \emph{IEEE Transactions on Image Processing}, vol.~28, no.~7, pp. 3343--3356, 2019.

\bibitem{MengJZWM22a}
X.~Meng, C.~Jia, X.~Zhang, S.~Wang, and S.~Ma, ``Deformable wiener filter for future video coding,'' \emph{IEEE Transactions on Image Processing}, vol.~31, pp. 7222--7236, 2022.

\bibitem{balle2018variational}
J.~Ball{\'e}, D.~Minnen, S.~Singh, S.~J. Hwang, and N.~Johnston, ``Variational image compression with a scale hyperprior,'' in \emph{International Conference on Learning Representations}, 2018.

\bibitem{lu2019dvc}
G.~Lu, W.~Ouyang, D.~Xu, X.~Zhang, C.~Cai, and Z.~Gao, ``{DVC}: An end-to-end deep video compression framework,'' in \emph{Proceedings of the IEEE/CVF Conference on Computer Vision and Pattern Recognition}, 2019.

\bibitem{yang2020learningRLVC}
R.~Yang, F.~Mentzer, L.~Van~Gool, and R.~Timofte, ``Learning for video compression with recurrent auto-encoder and recurrent probability model,'' \emph{IEEE Journal of Selected Topics in Signal Processing}, vol.~15, no.~2, pp. 388--401, 2020.

\bibitem{liu9578733}
B.~Liu, Y.~Chen, S.~Liu, and H.-S. Kim, ``Deep learning in latent space for video prediction and compression,'' in \emph{Proceedings of the IEEE/CVF Conference on Computer Vision and Pattern Recognition}, 2021, pp. 701--710.

\bibitem{lin2023deepsvc}
H.~Lin, B.~Chen, Z.~Zhang, J.~Lin, X.~Wang, and T.~Zhao, ``Deepsvc: Deep scalable video coding for both machine and human vision,'' in \emph{Proceedings of the 31st ACM International Conference on Multimedia}, 2023, pp. 9205--9214.

\bibitem{zhang2023elfic}
Z.~Zhang, B.~Chen, H.~Lin, J.~Lin, X.~Wang, and T.~Zhao, ``Elfic: A learning-based flexible image codec with rate-distortion-complexity optimization,'' in \emph{Proceedings of the 31st ACM International Conference on Multimedia}, 2023, pp. 9252--9261.

\bibitem{li2023neural}
J.~Li, B.~Li, and Y.~Lu, ``Neural video compression with diverse contexts,'' in \emph{Proceedings of the IEEE/CVF Conference on Computer Vision and Pattern Recognition}, 2023.

\bibitem{wang2023EVC}
G.-H. Wang, J.~Li, B.~Li, and Y.~Lu, ``Evc: Towards real-time neural image compression with mask decay,'' in \emph{International Conference on Learning Representations}, 2023.

\bibitem{li2024neural}
J.~Li, B.~Li, and Y.~Lu, ``Neural video compression with feature modulation,'' in \emph{Proceedings of the IEEE/CVF Conference on Computer Vision and Pattern Recognition}, 2024.

\bibitem{wiles2018x2face}
O.~Wiles, A.~Koepke, and A.~Zisserman, ``X2face: A network for controlling face generation using images, audio, and pose codes,'' in \emph{European Conference on Computer Vision}, 2018, pp. 670--686.

\bibitem{zakharov2019few}
E.~Zakharov, A.~Shysheya, E.~Burkov, and V.~Lempitsky, ``Few-shot adversarial learning of realistic neural talking head models,'' in \emph{Proceedings of the IEEE/CVF International Conference on Computer Vision}, 2019, pp. 9459--9468.

\bibitem{wang2019few}
T.-C. Wang, M.-Y. Liu, A.~Tao, G.~Liu, B.~Catanzaro, and J.~Kautz, ``Few-shot video-to-video synthesis.'' in \emph{Advances in Neural Information Processing Systems}, 2019.

\bibitem{hong2022depth}
F.-T. Hong, L.~Zhang, L.~Shen, and D.~Xu, ``Depth-aware generative adversarial network for talking head video generation,'' in \emph{Proceedings of the IEEE/CVF Conference on Computer Vision and Pattern Recognition}, 2022.

\bibitem{9455985}
D.~Feng, Y.~Huang, Y.~Zhang, J.~Ling, A.~Tang, and L.~Song, ``A generative compression framework for low bandwidth video conference,'' in \emph{IEEE International Conference on Multimedia and Expo Workshop}, 2021, pp. 1--6.

\bibitem{chen2023interactive}
B.~Chen, Z.~Wang, B.~Li, S.~Wang, S.~Wang, and Y.~Ye, ``Interactive face video coding: A generative compression framework,'' \emph{arXiv preprint arXiv:2302.09919}, 2023.

\bibitem{compressing2022bmvc}
M.~Agarwal, A.~Gupta, R.~Mukhopadhyay, V.~P. Namboodiri, and C.~Jawahar, ``Compressing video calls using synthetic talking heads,'' in \emph{British Machine Vision Conference}, 2023.

\bibitem{konuko2023predictive}
G.~Konuko, S.~Lathuili{\`e}re, and G.~Valenzise, ``Predictive coding for animation-based video compression,'' in \emph{IEEE International Conference on Image Processing}, 2023.

\bibitem{volokitin2022neural}
A.~Volokitin, S.~Brugger, A.~Benlalah, S.~Martin, B.~Amberg, and M.~Tschannen, ``Neural face video compression using multiple views,'' in \emph{Proceedings of the IEEE/CVF Conference on Computer Vision and Pattern Recognition}, 2022, pp. 1738--1742.

\bibitem{chen2023csvt}
B.~Chen, Z.~Wang, B.~Li, S.~Wang, and Y.~Ye, ``Compact temporal trajectory representation for talking face video compression,'' \emph{IEEE Transactions on Circuits and Systems for Video Technology}, vol.~33, no.~11, pp. 7009--7023, 2023.

\bibitem{CHEN2025DCC}
B.~Chen, S.~Yin, Z.~Zhang, J.~Chen, R.-L. Liao, L.~Zhu, S.~Wang, and Y.~Ye, ``Beyond {GFVC}: A progressive face video compression framework with adaptive visual tokens,'' in \emph{Data Compression Conference}, 2025.

\bibitem{10743918}
R.~Zhou, R.-L. Liao, B.~Chen, Y.~Ye, and J.~Chen, ``Enhanced multi-resolution generative face video compression,'' in \emph{2024 IEEE 26th International Workshop on Multimedia Signal Processing (MMSP)}, 2024, pp. 1--4.

\bibitem{yin2024parametertranslator}
S.~Yin, B.~Chen, S.~Wang, and Y.~Ye, ``Enabling translatability of generative face video coding: A unified face feature transcoding framework,'' in \emph{Data Compression Conference}, 2024, pp. 113--122.

\bibitem{JVET-AD0051}
B.~Chen, J.~Chen, Y.~Ye, and S.~Wang, ``{AHG9}: Common {SEI} message of generative face video,'' \emph{{The JVET of ITU-T SG 16 WP 3 and ISO/IEC JTC 1/SC 29, doc. no. JVET-AD0051}}, April 2023.

\bibitem{JVET-AI0191}
Y.-K. Wang, K.~Yang, Y.~Li, and Y.~Xu, ``{AHG9}: On {GFV} facial parameters signalling,'' \emph{{The JVET of ITU-T SG 16 WP 3 and ISO/IEC JTC 1/SC 29, doc. no. JVET-AI0191}}, July 2024.

\bibitem{JVET-AE0280}
B.~Chen, J.~Chen, Y.~Ye, S.~Wang, S.~McCarthy, P.~Yin, G.-M. Su, A.~K. Choudhury, and W.~Husak, ``{AHG9}: Common text for proposed generative face video {SEI} message,'' \emph{{The JVET of ITU-T SG 16 WP 3 and ISO/IEC JTC 1/SC 29, doc. no. JVET-AE0280}}, July 2023.

\bibitem{JVET-AI0189}
Y.-K. Wang, K.~Yang, Y.~Li, and Y.~Xu, ``{AHG9}: On signalling of and/or specifying {GFV} translator, generator, and enhancer {NNs},'' \emph{{The JVET of ITU-T SG 16 WP 3 and ISO/IEC JTC 1/SC 29, doc. no. JVET-AI0189}}, July 2024.

\bibitem{JVET-AG0048}
S.~Yin, B.~Chen, J.~Chen, R.-L. Liao, Y.~Ye, and S.~Wang, ``{AHG16}: Interoperability study on parameter translator of generative face video coding,'' \emph{{The JVET of ITU-T SG 16 WP 3 and ISO/IEC JTC 1/SC 29, doc. no. JVET-AG0048}}, January 2024.

\bibitem{JVET-AI0194}
S.~Gehlot, G.-M. Su, P.~Yin, S.~McCarthy, and G.~J. Sullivan, ``{AHG9/AHG16}: On chroma key fusion for the generative face video {SEI} message,'' \emph{{The JVET of ITU-T SG 16 WP 3 and ISO/IEC JTC 1/SC 29, doc. no. JVET-AI0194 }}, July 2024.

\bibitem{JVET-AH0127}
J.~Chen, B.~Chen, Y.~Ye, R.-L. Liao, S.~Yin, and S.~Wang, ``{AHG9/AHG16}: The {SEI} message design for scalable representation and layered reconstruction for generative face video compression,'' \emph{{The JVET of ITU-T SG 16 WP 3 and ISO/IEC JTC 1/SC 29, doc. no. JVET-AH0127}}, April 2024.

\bibitem{JVET-AH0110}
B.~Chen, Y.~Ye, J.~Chen, R.-L. Liao, S.~Yin, and S.~Wang, ``{AHG16}: Scalable representation and layered reconstruction for generative face video compression,'' \emph{{The Joint Video Experts Team of ITU-T SG 16 WP 3 and ISO/IEC JTC 1/SC 29, doc. no. JVET-AH0110}}, April 2024.

\bibitem{JVET-AG0139}
R.~Zou, R.-L. Liao, B.~Chen, Y.~Ye, and J.~Chen, ``{AHG16}: Depthwise separable convolution for generative face video compression,'' \emph{{The JVET of ITU-T SG 16 WP 3 and ISO/IEC JTC 1/SC 29, doc. no. JVET-AG0139}}, January 2024.

\bibitem{dists}
K.~Ding, K.~Ma, S.~Wang, and E.~Simoncelli, ``Image quality assessment: Unifying structure and texture similarity,'' \emph{IEEE Transactions on Pattern Analysis and Machine Intelligence}, 2020.

\bibitem{lpips}
R.~Zhang, P.~Isola, A.~Efros, E.~Shechtman, and O.~Wang, ``The unreasonable effectiveness of deep features as a perceptual metric,'' in \emph{Proceedings of the IEEE/CVF Conference on Computer Vision and Pattern Recognition}, 2018.

\bibitem{2009Mean}
W.~Zhou and A.~C. Bovik, ``Mean squared error: Love it or leave it? a new look at signal fidelity measures,'' \emph{IEEE Signal Processing Magazine}, vol.~26, no.~1, pp. 98--117, 2009.

\bibitem{wang2004image}
Z.~Wang, A.~C. Bovik, H.~R. Sheikh, and E.~P. Simoncelli, ``Image quality assessment: from error visibility to structural similarity,'' \emph{IEEE transactions on image processing}, vol.~13, no.~4, pp. 600--612, 2004.

\bibitem{wang2021Nvidia}
T.~Wang, A.~Mallya, and M.~Liu, ``One-shot free-view neural talking-head synthesis for video conferencing,'' in \emph{Proceedings of the IEEE/CVF Conference on Computer Vision and Pattern Recognition}, 2021, pp. 10\,039--10\,049.

\bibitem{RFB15a}
O.~Ronneberger, P.Fischer, and T.~Brox, ``{U-Net}: Convolutional networks for biomedical image segmentation,'' in \emph{Medical Image Computing and Computer-Assisted Intervention}, 2015, pp. 234--241.

\bibitem{Hourglass2016}
A.~Newell, K.~Yang, and J.~Deng, ``Stacked hourglass networks for human pose estimation,'' in \emph{European Conference on Computer Vision}, 2016, pp. 483--499.

\bibitem{J2015Density}
J.~Ball{\'e}, V.~Laparra, and E.~Simoncelli, ``Density modeling of images using a generalized normalization transformation,'' in \emph{International Conference on Learning Representations}, 2016.

\bibitem{NEURIPS2018_53edebc5}
D.~Minnen, J.~Ball\'{e}, and G.~D. Toderici, ``Joint autoregressive and hierarchical priors for learned image compression,'' in \emph{Advances in Neural Information Processing Systems}, vol.~31, 2018.

\bibitem{park2019SPADE}
T.~Park, M.-Y. Liu, T.-C. Wang, and J.-Y. Zhu, ``Semantic image synthesis with spatially-adaptive normalization,'' in \emph{Proceedings of the IEEE Conference on Computer Vision and Pattern Recognition}, 2019.

\bibitem{Nagrani17}
A.~Nagrani, J.~S. Chung, and A.~Zisserman, ``Voxceleb: a large-scale speaker identification dataset,'' in \emph{INTERSPEECH}, 2017.

\bibitem{300VW}
G.~G. Chrysos, E.~Antonakos, S.~Zafeiriou, and P.~Snape, ``Offline deformable face tracking in arbitrary videos,'' in \emph{Proceedings of the IEEE/CVF International Conference on Computer Vision Workshop}, 2015.

\bibitem{li2023perceptual}
Y.~Li, B.~Chen, B.~Chen, M.~Wang, S.~Wang, and W.~Lin, ``Perceptual quality assessment of face video compression: A benchmark and an effective method,'' \emph{IEEE Transactions on Multimedia}, pp. 1--13, 2024.

\bibitem{zhu2022celebvhq}
H.~Zhu, W.~Wu, W.~Zhu, L.~Jiang, S.~Tang, L.~Zhang, Z.~Liu, and C.~C. Loy, ``{CelebV-HQ}: A large-scale video facial attributes dataset,'' in \emph{European Conference on Computer Vision}, 2022.

\bibitem{compressionmeasure}
Y.~Li, S.~Wang, X.~Zhang, S.~Wang, S.~Ma, and Y.~Wang, ``Quality assessment of end-to-end learned image compression: The benchmark an objective measure,'' in \emph{Proceedings of the ACM International Conference on Multimedia}, 2021.

\bibitem{Unterthiner2019FVDAN}
T.~Unterthiner, S.~van Steenkiste, K.~Kurach, R.~Marinier, M.~Michalski, and S.~Gelly, ``Fvd: A new metric for video generation,'' in \emph{7th International Conference on Learning Representations workshop}, 2019.

\bibitem{heusel2017gans}
M.~Heusel, H.~Ramsauer, T.~Unterthiner, B.~Nessler, and S.~Hochreiter, ``Gans trained by a two time-scale update rule converge to a local nash equilibrium,'' \emph{Advances in neural information processing systems}, vol.~30, 2017.

\bibitem{yang2022maniqa}
S.~Yang, T.~Wu, S.~Shi, S.~Lao, Y.~Gong, M.~Cao, J.~Wang, and Y.~Yang, ``Maniqa: Multi-dimension attention network for no-reference image quality assessment,'' in \emph{Proceedings of the IEEE/CVF Conference on Computer Vision and Pattern Recognition}, 2022, pp. 1191--1200.

\bibitem{Bjntegaard2001CalculationOA}
G.~Bjontegaard, ``Calculation of average psnr differences between rd-curves,'' \emph{ITU SG16 Doc. VCEG-M33}, 2001.

\bibitem{gou2021knowledge}
J.~Gou, B.~Yu, S.~J. Maybank, and D.~Tao, ``Knowledge distillation: A survey,'' \emph{International Journal of Computer Vision}, vol. 129, no.~6, pp. 1789--1819, 2021.

\bibitem{polino2018model}
A.~Polino, R.~Pascanu, and D.~Alistarh, ``Model compression via distillation and quantization,'' \emph{arXiv preprint arXiv:1802.05668}, 2018.

\bibitem{han2015deep_compression}
S.~Han, H.~Mao, and W.~J. Dally, ``Deep compression: Compressing deep neural networks with pruning, trained quantization and huffman coding,'' \emph{International Conference on Learning Representations}, 2016.

\bibitem{m64987}
Y.~Ye, S.~McCarthy, H.~B.-T., Z.~Lv, S.~Wang, K.~Zhang, M.~Karczewicz, and I.~Moccagatta, ``On {VVC}-assisted ultra-low rate generative face video coding,'' \emph{{MPEG ISO/IEC JTC 1/SC 29/WG 2 doc. no. m64987}}, October 2023.

\end{thebibliography}
\end{document}